\documentclass[journal]{new-aiaa}
\usepackage[utf8]{inputenc}
\usepackage{textcomp}

\usepackage{graphicx}
\usepackage{subfigure}
\usepackage{amsmath}
\usepackage[version=4]{mhchem}
\usepackage{siunitx}
\usepackage{longtable,tabularx}
\usepackage{mathtools}

\usepackage{xtab,booktabs}
\usepackage{indentfirst} % For every paragraph with 4 spaces
\usepackage{multirow} % Required for multi-rows
\usepackage{newtxmath}
\usepackage{algorithm}
\usepackage{algorithmic}

\usepackage{siunitx}

\newtheorem{assumption}{Assumption}
\newtheorem{definition}{Definition}
\newtheorem{remark}{Remark}

\usepackage{xcolor}
\newcommand{\col}{\color{black}}
\newcommand{\colm}{\color{black}}

\newcommand{\colb}{\color{black}}

\title{Distributed Spatial-Temporal Trajectory Optimization for Unmanned-Aerial-Vehicle Swarm}

\author{Xiaobo Zheng \footnote{PhD Student, School of Aerospace Engineering, 5th Zhongguancun South Street}, Defu Lin  \footnote{Professor, School of Aerospace Engineering, 5th Zhongguancun South Street}}
\affil{Beijing Institute of Technology, Beijing 100081, China}
\author{Pan Tang \footnote{Assidant Engineer, No.1 Beijing Dahongmen South Road}}
\affil{Beijing Institute of Aerospace Systems Engineering, Beijing 100076, China}
\author{Shaoming He \footnote{Professor, School of Aerospace Engineering, 5th Zhongguancun South Street, Member AIAA, Email: \texttt{shaoming.he@bit.edu.cn}}}
\affil{Beijing Institute of Technology, Beijing 100081, China}

\begin{document}

\maketitle

%%%%%%%%%%%%%%%%%%%%%%%%%%%%%%%

\begin{abstract}
Swarm trajectory optimization problems are a well-recognized class of multi-agent optimal control problems with strong nonlinearity. However, the heuristic nature of needing to set the final time for agents beforehand and the time-consuming limitation of the significant number of iterations prohibit the application of existing methods to large-scale swarm of Unmanned Aerial Vehicles (UAVs) in practice. {\colb In this paper, we propose a spatial-temporal trajectory optimization framework that accomplishes multi-UAV consensus based on the Alternating Direction Multiplier Method (ADMM) and uses Differential Dynamic Programming (DDP) for fast local planning of individual UAVs.} The introduced framework is a two-level architecture that employs Parameterized DDP (PDDP) as the trajectory optimizer for each UAV, and ADMM to satisfy the local constraints and accomplish the spatial-temporal parameter consensus among all UAVs. This results in a fully distributed algorithm called Distributed Parameterized DDP (D-PDDP). In addition, an adaptive tuning criterion based on the spectral gradient method for the penalty parameter is proposed to reduce the number of algorithmic iterations. Several simulation examples are presented to verify the effectiveness of the proposed algorithm.
\end{abstract}

%%%%%%%%%%%%%%%%%%%%%%%%%%%%%%%

\section{Introduction}
Autonomous swarm of UAVs have received increasing attention in recent years due to its wide range of applications in both military and civilian missions, e.g., wild fire monitoring \cite{julian2019distributed}, search and rescue \cite{9256816}, situational awareness \cite{Paul2018Scharre}, cooperative interception \cite{balhance2017cooperative}. {\colb It is well-recognized that trajectory optimization plays an essential part in enhancing the swarm's autonomy} and the main objective is to accomplish mission with optimality while ensuring safety of the swarm. Mathematically, this problem can be formulated as a Multi-Agent Optimal Control (MAOC) problem, i.e., finding the optimal control history of each agent to minimize the global common performance metrics {\colb (e.g., control effort consumption, mission completion time) while complying with multiple nonlinear constraints (e.g., dynamic constraint, control constraint, obstacle avoidance, and inter-UAV collision avoidance).}

%In recent years, the field of UAV swarms has received more attention \cite{doi:10.1080/00963402.2018.1533209}, and the cooperation of swarm UAVs significantly enhances their capability in typical missions \cite{9256816,9657810,9438461,10158519,8428707} such as enemy reconnaissance, situational awareness, destructive strikes, and swarm confrontation. 

%Swarm technology covers a wide range of topics, including coordinated control, multi-target tracking, decision-making, and distributed trajectory optimization, and so on \cite{10180039,9998989,doi:10.2514/1.G002937,doi:10.2514/1.G006206}. Between these techniques, swarm trajectory optimization is a key technology to improve the autonomy level of UAV swarms. The number of swarms varies from tens to hundreds, and each individual UAV in the resulting team needs to cooperate to achieve a shared objective \cite{10230306,9484757}. The main goal of swarm trajectory optimization is to accomplish the swarm mission with optimal effectiveness while ensuring the UAV swarm flies safely in the environment. This problem can be modeled as a Multi-Agent Optimal Control (MAOC) problem to find the optimal control history of each agent to minimize the global common performance metrics (e.g., energy consumption, flight endurance, flight duration) while satisfying multiple nonlinear constraints (e.g., dynamics, control constraints, no-fly zones, and inter-aircraft collision avoidance).

{\colb Trajectory optimization for a single agent has been widely studied and different types of numerical methods are used to solve this nonlinear problem, such as} pseudo-spectral methods \cite{7875167,doi:10.2514/1.G005038}, sequential quadratic programming \cite{doi:10.2514/1.G004874,8884201}, sequential convex programming \cite{doi:10.2514/1.G000218,8447526}, and DDP \cite{doi:10.2514/1.G003516}. The authors in \cite{chai2019review} provided a well-organized description of the understanding of different numerical optimization algorithms from the perspective highlighting their advantages and mathematical nature. {\colb Due to its rapid convergence features and well-developed problem formulation framework, DDP has emerged as a popular solution for resolving nonlinear optimal control problems \cite{icinco04,aziz2019hybrid,ozaki2018stochastic,he2019computational,zhang2022aerocapture}.} 
%Among these methods, DDP has become a widespread technique for solving nonlinear optimal control problems due to its well-developed problem formulation framework and fast convergence properties \cite{icinco04,aziz2019hybrid,ozaki2018stochastic,he2019computational,zhang2022aerocapture}. 
DDP is derived from classical Dynamic Programming (DP) that is based on Bellman's optimality principle and leverages second-order Taylor expansion to approximate the value function. {\colb Through this method, the original problem is decomposed into smaller dimensional subproblems related to the number of discrete points, and then the optimal variation of the control effort is determined using the first-order optimality condition to continually create new trajectories until convergence. With second-order Taylor approximation in the domain of nominal trajectories, DDP overcomes the curse of dimensionality of traditional Dynamic Programming (DP) methods \cite{doi:10.1080/00207176608921369,pavlov2021interior} and has faster second-order convergence rates \cite{manchester2016derivative}.}
%the first-order optimality condition is used to find the optimal increment of the control sequence to generate new trajectories continuously until convergence. 
%This means that the DDP offers application possibilities for high-speed vehicles (e.g., missiles, unmanned aerial vehicles) that require real-time trajectory planning. In addition, the DDP provides locally optimal feedback controllers based on the algorithm's value function after generating the optimal control history, which may be helpful when the actual flight trajectory is perturbed away from the nominal trajectory.

Unlike the single agent scenario, solving the MAOC problem requires a suitable architecture to share the information among agents. Compared to the centralized architecture, the distributed architecture provides improved robustness against single agent failure and scalability with respect to the number of agents in the swarm \cite{shorinwa2023distributeda}. For this reason, the distributed trajectory optimization shows great potentials in application to UAV swarms since the onboard computational resource is usually very limited \cite{shorinwa2023distributeda}. Three typical types of distributed trajectory optimization methods to solve MAOC reported in the literature are distributed gradient descent \cite{6930814}, distributed sequential convex programming \cite{7590162,7398129}, and ADMM \cite{MAL-016,6731604}. Among them, the ADMM has attracted significant attention due to its ability to generate distributed structures, solve problems involving numerous optimization variables, and handle equality constraints \cite{Saravanos2021DistributedCS,9147590}. A detailed review of ADMM is given by Boyd and shows how this method can be applied to large-scale distributed optimization problems \cite{MAL-016}. ADMM decomposes the original large-scale optimization problem into a number of simpler sub-problems, each of which optimizes an independent objective function and corresponding decision variables. These sub-problems are solved by alternately updating the primal and dual variables to minimize the augmented Lagrangian function separately by their respective optimization solvers, and then the global solution of the original problem is obtained by gathering the solutions of the sub-problems. Notice that the original ADMM method requires a central node to collect the information from all nodes for the dual variable updating. This assumption was eliminated in \cite{6731604} by introducing the auxiliary primal variables in place of the communication protocol constraints in the original ADMM, resulting in a fully distributed consensus alternating directional multiplier method (C-ADMM).

%For solving a combined optimal control problem with multiple agents, it is necessary to use a suitable architecture to solve this multi-agent optimal control problem.
%A straightforward approach is to use a centralized architecture to solve this problem, which means that a single computationally powerful computing unit is assumed to perform all computations and communicate with each agent \cite{7795736}. However, the centralized architecture approach's computational demand grows exponentially with the number of agents, resulting in poor scalability and making it difficult to apply in practical large-scale systems. Conversely, 
%Distributed architectures have been extensively studied in multi-agent optimal control problems due to their good scalability for large-scale agent systems and their ability to improve computational efficiency by exploiting distributed features \cite{halstedSurveyDistributedOptimization2021,shorinwaDistributedOptimizationMethods2023a}. 

%The properties of ADMM give the method the possibility of generating a distributed nature.  
Unfortunately, the strongly non-convex nature of the UAV swarm trajectory optimization problem, introduced by dynamic constraints, collision constraints, communication distance maintenance, makes it difficult to apply the ADMM directly to practical missions. One possible way is to combine the local single agent numerical solvers with ADMM to deal with the dynamic constraints, coordination constraints, and consensus constraints in MAOC. The authors in \cite{9485100,9589675} proposed a multi-agent DDP approach based on ADMM, but the dual variable update step of the algorithm is still computed centrally and hence their practical scalability is limited. {\colb One notable work in \cite{10288223} proposed a framework in which the distributed computation based on ADMM algorithm and the local planning based on DDP is performed.} The fully distributed nature of the framework shows superior scalability against sequential convex programming and quadratic programming methods. However, the algorithm needs to set the terminal time for each agent beforehand and two additional issues are till need to be considered: time optimization and convergence acceleration.

%by Saravanos et al. proposed a framework in which both primal and dual variables can be computed in a distributed manner, and the fully distributed nature of the framework allows the algorithm to scale to thousands of unmanned vehicle systems, obtaining superior scalability against sequential convex programming and quadratic programming methods \cite{10288223}.

%It must be combined with local trajectory solvers such as DDP to handle the non-convex dynamics and then perform the algorithms. Researchers have combined the computational efficiency and scalability of DDP with the distributed nature of ADMM for dealing with the dynamics constraints, coordination constraints, and consensus constraints that are presented in multi-unmanned system problems. 

In some practical UAV swarm application, pre-setting the terminal time is intractable in nature for some missions and time optimization is a critical metric from the mission perspective. For example, the mission duration should be minimized in some missions, e.g., search and rescue, courier delivery \cite{horyna2023decentralized,li2022traffic}. The authors in \cite{niRobustMultiRobotTrajectory2022b} developed a fast time-optimal trajectory planning method for UAVs using the ADMM algorithm, where the ADMM is utilized as a trajectory optimization algorithm rather than a distributed architectural approach for multi-agent systems, i.e., the communication process and distributed computing is not considered during the algorithm design. A swarm UAV trajectory planner that considers spatial-temporal optimization to shape the trajectory and modify the time allocation simultaneously was developed in \cite{zhouSwarmMicroFlying2022} for autonomous flight in complicated jungle environments. However, this work relies more on individual sensors and onboard planners of each UAV, with less information sharing between the UAVs, therefore showing limited mission-specific robustness, and challenges adapting to sudden mission changes. {\colb By dynamically optimizing the total time of the trajectory as well as the control effort, we reviewed the original DDP in our earlier work \cite{9582791,doi:10.2514/1.G007113} and developed flexible final time DDP (FFT-DDP).}
%In our previous work \cite{9582791,doi:10.2514/1.G007113}, we revisited the original DDP by dynamically optimizing the final time in conjunction with the control input and proposed flexible final time DDP (FFT-DDP). 
A method, i.e., PDDP, to simultaneously optimize the parameters and control inputs is proposed in \cite{oshin2022parameterized} with rigorous mathematical proof showing that the algorithm enjoys the property of fast and stable convergence to the minimum cost. By setting the flight time as an unknown parameter, PDDP can be utilized to solve the optimal flight time problem. However, the extension of these algorithms to the multi-agent framework is not straightforward and requires a substantial adjustment. Also, in the mission of simultaneous attack or sequential attack of a high-value target, the terminal time interval between two vehicles should be specified. This will introduces the terminal time sequence constraints and has not been resolved before.

Another critical issue in applying ADMM is that the convergence performance as well as the reliability largely depends on the selection of its penalty parameters. Residual balancing (RB) is a well-established and available adaptive method for solving general problems. This approach is based on the idea of balancing the primal and dual residuals to be approximately equal while improves the convergence by increasing or decreasing the penalty parameter with iterations \cite{MAL-016}. However, the performance of the RB method varies greatly with the size of the problem. The authors in \cite{doi:10.1137/120896219} utilized the Nesterov acceleration method to speed up the convergence of the algorithm, but the method does not work well for general problems. The Barzilai-Borwein `spectral' method was leveraged in \cite{pmlr-v54-xu17a, Xu_2017_CVPR} to estimate the local curvature of the objective function to establish a penalty parameter adaptive approach with theoretical convergence proof. The extension to distributed optimization, however, needs further adjustment.

Motivated by the above observations, this paper proposes a distributed spatial-temporal optimization method based on PDDP and ADMM. The proposed algorithm is built on a two-level architecture basis, in which the PDDP layer is employed to solve local trajectory optimization problem with control limits, while ADMM layer is utilized to satisfy the local path constraints as well as the spatial-temporal coordination among all UAVs. {\col The first major contribution of this work lies in the development of a method that simultaneously optimizes both the trajectories and flight times of a UAV swarm by resolving the issue setting the final time for agents beforehand. This dual optimization of spatial and temporal aspects in a single framework is a significant step forward in multi-agent swarm control. To the best of the authors' knowledge, the distributed free-time DDP is the first fully decentralized joint spatial-temporal optimization method and is capable of being applied to large-scale UAV swarms. The second key contribution is the introduction of a parameter adaptive approach designed to accelerate the convergence of the ADMM layer. This approach dynamically adjusts the penalty parameter during the optimization process, significantly reducing the number of iterations required, which in turn leads to computational time savings.} A variety of numerical simulations with flight time optimization and time coordination constraints are leveraged to evaluate the effectiveness of the proposed algorithm. Up to the best of our knowledge, no similar works have been reported in the literature. The results reveal that the proposed algorithm is capable of handling different mission scenarios, e.g., free terminal time, simultaneous arrival and specified time intervals, and is capable of being applied to large-scale UAV swarms.

The remainder of the paper is organized as follows. Section \ref{sec: preliminaries} introduces the general form of the multi-agent optimal control problem and briefly reviews the PDDP and ADMM methods. The proposed D-PDDP algorithm is described in details in Section \ref{sec: innovation}, followed by the adaptive acceleration in Section \ref{sec: ACD-PDDP}. A variety of numerical evaluations of the algorithm's performance are presented in Section \ref{sec: application} and some conclusions are offered in Section \ref{sec: conclusion}.

% contains the conclusion of the paper and a discussion of future research directions.

%In this paper, we denote scalar by $a$, vectors by $\boldsymbol{a}$ and matrices by $\mathbf A$. $\boldsymbol{0}_n$ denotes $n$ dimensional zero vectors and ${\mathbf I}_{n}$ is the $n$-by-$n$ identity matrix. Discrete integers $\{ a,a+1,a+2,\ldots,b \}$ in the interval $\left[ a,b \right]$ can be denoted as $\llbracket a,b \rrbracket$. The stack of vectors $\{ \boldsymbol{a}_1,\boldsymbol{a}_2,\ldots,\boldsymbol{a}_N \}$ can be denoted by $\boldsymbol{A}$. The number of elements of a set $\mathcal{C}$ can be denoted as $\vert \mathcal{C} \vert$. The indicator function $\mathcal{I}_{\mathcal{C}} (\boldsymbol{x})$ is defined to denote whether the vector $\boldsymbol{x}$ is on the set $\mathcal{C}$. If the vector $\boldsymbol{x}$ is on the set $\mathcal{C}$, then $\mathcal{I}_{\mathcal{C}} (\boldsymbol{x})=0$; conversely $\boldsymbol{x}$ is not on $\mathcal{C}$, then $\mathcal{I}_{\mathcal{C}} (\boldsymbol{x})=+\infty$.
%The operator ${\max}(\boldsymbol{a}, \boldsymbol{b})$ (respectively, ${\min}(\boldsymbol{a},\boldsymbol{b})$) finds the maximum value (respectively the minimum value) from $\boldsymbol{a}$ and $\boldsymbol{b}$.

%%%%%%%%%%%%%%%%%%%%%%%%%%%%%%

\section{Backgrounds and Preliminaries} \label{sec: preliminaries}
In this section, we first introduce the general MAOC problem and then {\col provide} some preliminaries of the {\col PDDP} and ADMM for the completeness of the paper.

\subsection{Multi-Agent Optimal Control Problem} \label{ssec:formulation}
For a typical MAOC problem, {\colb there is a group of $M$ agents represented as a set $\mathcal{M} = \{1,... ,M\}$. Before describing the mathematical problems, we first define the concepts related to neighbors \cite{10288223}. 

\begin{definition}
	(Neighbor set): The `neighbor' set is a set that contains all neighbor-agents $j\in \mathcal{M}$ (including agent $i$) of agent $i\in \mathcal{M}$, which can be denoted as $\mathcal{N}_i\in \mathcal{M}$.
	%(Neighborhood Set): The `neighborhood' set $\mathcal{N}_i \subseteq \mathcal{M}$ of an agent $i \in \mathcal{M}$ is the set that contains all agents $j \in \mathcal{M}$ that are neighbors of $i$ (including $i$).
\end{definition}
\begin{definition}
	(Deemed Neighbor Set): A "deemed" neighbor set, represented as $\mathcal{P}_i = \{j \in \mathcal{M}: i \in \mathcal{N}_j\}$, is a collection of agents $j\in \mathcal{M}$ that regard agent $i\in \mathcal{M}$ as their neighbor.
	%a `deemed' neighbor set is a set of agents $j\in \mathcal{M}$ that consider agent $i\in \mathcal{M}$ as their neighbor, which can be denoted as $\mathcal{P}_i = \{j \in \mathcal{M}: i \in \mathcal{N}_j\}$.
	%(Neighbor-of Set): The `neighbor-of' set $\mathcal{P}_i \subseteq \mathcal{M}$ of an agent $i \in \mathcal{M}$ is defined as $\mathcal{P}_i = \{j \in \mathcal{M}: i \in \mathcal{N}_j\}$. In other words, it is the set that contains all agents $j \in \mathcal{M}$ that consider $i$ as a neighbor.
\end{definition}
}

In a UAV swarm, we let $j \in \mathcal{N}_i$ if UAV $j$ is within the effective communication range of $i$. Note that there is no requirement for $i$ and $j$ to be neighbors of each other, i.e., there is no requirement for $\mathcal{N}_i = \mathcal{P}_i$. Furthermore, we require that the agents communicate with each other only locally by the following assumptions.
\begin{assumption}
	{\colb Each agent $i \in \mathcal{M}$ is limited to exchanging information with agents $j \in \mathcal{M}$ that are both $i$'s neighbors and regard $i$ to be a neighbor; that is, $i$ can only interact with $j \in \mathcal{N}_i \cup \mathcal{P}_i$.
	%Each agent $i \in \mathcal{M}$ can only communicate with agents $j \in \mathcal{M}$ that are neighbors of $i$ and also regard $i$ as a neighbor, i.e., $i$ can only exchange information with $j \in \mathcal{N}_i \cup \mathcal{P}_i$.
	}
	%Each agent $i \in \mathcal{M}$ is able to exchange information only with agents $j \in \mathcal{N}_i \cup \mathcal{P}_i$.
\end{assumption}

We consider that each agent $i$ of the swarm is governed by a discrete-time nonlinear dynamics as
%
%has a general nonlinear differential equation that describes the dynamics of an autonomous vehicle as
%\begin{equation}
%	\dot{\boldsymbol{x}}_i=F_{i}(\boldsymbol{x}_i, \boldsymbol{u}_i), \quad {\boldsymbol{x}_i}\left( {t_{i,0}} \right) = {\boldsymbol{x}}_{i,0}
%	\label{eq:consys}
%\end{equation}
%where ${\boldsymbol{x}_{i}\in \mathbb{R}^{p_i}}$ and ${\boldsymbol{u}_{i}\in \mathbb{R}^{q_i}}$ are the state and control vectors of the $i$-th agent, respectively. $F_{i}$ is a nonlinear and continuously differentiable function of $\boldsymbol{x}_{i}$ and $\boldsymbol{u}_{i}$. The state of the considered dynamic system at the initial time instant $t_{0}$ is denoted as ${\boldsymbol{x}}_{i,0}$.
%
%The trajectory optimizer used in this paper is formulated in a discrete-time manner and therefore a discretization of the continuous time system is required to allow the application of the trajectory optimizer. By using a simple Eulerian discretization or the Runge-Kutta method, the continuous dynamics \eqref{eq:consys} of the $i$th agent can be revised as the following discrete-time nonlinear equations
\begin{equation}
\boldsymbol{x}_{i,k+1} = \boldsymbol{f}_i \left( \boldsymbol{x}_{i,k}, \boldsymbol{u}_{i,k} \right)
\label{eq:k1}
\end{equation}
where ${\boldsymbol{x}_{i,k}\in \mathbb{R}^{p_i}}$ and ${\boldsymbol{u}_{i,k}\in \mathbb{R}^{q_i}}$ are the state and control vectors of the $i$-th agent, respectively. {\colb The subscript $k = \left\{ {0,1, \ldots ,N} \right\}$ represent the time instant in the discrete framework, and the value of a variable at time instant $t_k$ is indicated by the variable with subscript $k$. $\boldsymbol{f}_i$ represents discrete system dynamics with nonlinear properties.}
%variables with subscript $k$ represent the value of the variable at time instant $t_k$.
%The index $k = \left\{ {0,1, \ldots ,N} \right\}$ stands for the discrete time instant. $\boldsymbol{f}_i$ is a nonlinear function that describes the system dynamics, and the variable with subscript $k$ represents the corresponding value at time instant $t_k$.

{\colb The problem's global cost function, which is attempted to minimize by each agent, can be given by}
%The global cost function of the problem that all agents in the team aiming at minimizing is
\begin{equation}
%	J = \sum_{i=1}^{M}  J_i(\boldsymbol{x}_i,\boldsymbol{u}_i)
	J = \sum_{i=1}^{M}  J_i\left(\boldsymbol{X}_i, \boldsymbol{U}_i\right)
	\label{eq: all costf}
\end{equation}
in which $\boldsymbol{X}_{i}$ and $\boldsymbol{U}_{i}$ represent the stack of state vectors and control inputs from all time instants of agent $i$. Each agent has their own local objective function that contains terminal constraint and running cost as
\begin{equation}
	J_i\left(\boldsymbol{X}_i, \boldsymbol{U}_i\right) = \sum_{k=0}^{N-1}  l_{i}(\boldsymbol{x}_{i,k}, \boldsymbol{u}_{i,k}) + \phi_{i}(\boldsymbol{x}_{i,N}, t_{i,N})
	\label{eq: discrete cost func}
\end{equation}
{\colb in which the running cost is denoted by scalar function $l_{i}(\boldsymbol{x}_{i,k}, \boldsymbol{u}_{i,k})$ while the terminal cost is represented by $\phi_{i}(\boldsymbol{x}_{i,N}, t_{i,N})$.}
%scalars function $\phi_{i}(\boldsymbol{x}_{i,N}, t_{i,N})$ stands for the terminal constraint and $l_{i}(\boldsymbol{x}_{i,k}, \boldsymbol{u}_{i,k})$ refers to the running cost.

Due to the physical constraints, e.g., actuation limitations, {\colb the control effort of each agent is commonly constrained}, i.e.,
\begin{equation}
	\boldsymbol{b}_{i,k}\left(\boldsymbol{u}_{i, k}\right) \leq 0
	\label{eq: control constraint}
\end{equation}
where $\boldsymbol{b}_{i,k}\left(\boldsymbol{u}_{i, k}\right)$ denotes the control effort constraint function.

{\colb In addition to the control effort constraints, each vehicle's trajectory will also be limited by nonlinear path constraints, such as obstacle avoidance:
%Except for the control input constraint, the trajectory of each vehicle is also subject to nonlinear path constraints, e.g., no-fly zones, as
\begin{equation}
	\boldsymbol{a}_{i,k} \left( {\boldsymbol{x}_{i,k}} \right) \leq 0
	\label{eq: nonlinear constraint}
\end{equation}
where $\boldsymbol{a}_{i,k}$ is a nonlinear function for describing path constraints.}

In particular, in a spatial-temporal optimization problem, it is free for each vehicle w.r.t. its final time. Also, the terminal time should be within the boundaries of the vehicle's achievable range. Thus, we formulate the terminal time constraint as
\begin{equation}
	c_{i, N}\left(t_{i, N}\right) \leq 0
\end{equation}
where $c_{i, N}\left(t_{i, N}\right)$ is a general nonlinear function.

With the definition of neighbors, the inter-agent state constraint between $i$ and its neighbors can be expressed as
\begin{equation}
	\boldsymbol{d}_{i j, k}\left(\boldsymbol{x}_{i, k}, \boldsymbol{x}_{j, k}\right) \leq 0
	\label{eq:intercons1}
\end{equation}
and the inter-agent terminal time constraint between $i$ and its neighbors can be formulated as:
\begin{equation}
	\boldsymbol{e}_{i j, N}\left(t_{i, N}, t_{j, N}\right) = 0
	\label{eq:intercons2}
\end{equation}
{\colb in which the above two inter-agent constraints do not contain agent $i$ itself, i.e. $j\neq i$. And} the expression of $\boldsymbol{d}_{i j, k}\left(\boldsymbol{x}_{i, k}, \boldsymbol{x}_{j, k}\right)$ represents the constraint of collision avoidance between UAV $i$ and $j$, or the constraint that $i$ and $j$ maintain their connection within the communication distance. And the function $\boldsymbol{e}_{i j, N}\left(t_{i, N}, t_{j, N}\right)$ stands for the constraint that the all agents have the same final time or complete a mission in a certain time sequence.

In summary, the multi-agent optimal control problem, denoted as \textit{MAOC}$_1$, considered in this paper has the following form.\\
\textit{MAOC}$_1$: Find
\begin{equation}
	\begin{aligned}
		\{ \boldsymbol{U}_{i}^*, t_{i,N}^* \}&=\min {\sum_{i=1}^{M}  J_i\left(\boldsymbol{X}_i, \boldsymbol{U}_i\right)}\\
		s.t. \qquad &\boldsymbol{x}_{i,k+1} = \boldsymbol{f}_{i}\left( \boldsymbol{x}_{i,k}, \boldsymbol{u}_{i,k} \right)\\
		&\boldsymbol{a}_{i,k}\left( {\boldsymbol{x}_{i,k}} \right) \le 0 \\
		&\boldsymbol{b}_{i,k}\left(\boldsymbol{u}_{i, k}\right) \leq 0\\
		&c_{i, N}\left(t_{i, N}\right) \leq 0\\
		&\boldsymbol{d}_{i j, k}\left(\boldsymbol{x}_{i, k}, \boldsymbol{x}_{j, k}\right) \leq 0, \quad j \neq i\\
		&\boldsymbol{e}_{i j, N}\left(t_{i, N}, t_{j, N}\right) = 0, \quad j \neq i
	\end{aligned}
	\label{eq: MAOC1}
\end{equation}

{\colb Problem \textit{MAOC}$_1$ is assumed to have a solution. Because of the inherent nonlinearity, analytical problem solving is usually unfeasible. As a result, numerical procedures are typically employed to tackle this problem.}
%It is assumed that problem \textit{MAOC}$_1$ has a solution. However, finding the solution analytically is in general intractable due to the inherent nonlinearity and hence numerical algorithms are usually leveraged to solve the problem.

\subsection{{\col Parameterized Differential Dynamic Programming}}
The PDDP algorithm is a variant of DDP to optimize the control sequence and unknown parameter simultaneously. By abstracting the terminal time as an unknown parameter, PDDP is able to solve the problem of optimizing spatial-temporal trajectory without pre-setting terminal time for a single agent. For the completeness, we briefly review the key points of PDDP in this subsection and ignore the agent index for simplicity. PDDP is designed to solve the optimization problem as
\begin{equation}
	\begin{aligned}
		\underset{\boldsymbol{U}; \boldsymbol{\theta}}{\operatorname{\min}} J(\boldsymbol{U}; \boldsymbol{\theta}) &= \sum\limits_{k = 1}^{N - 1} {l \left( \boldsymbol{x}_k,\boldsymbol{u}_k;\boldsymbol{\theta}\right)} + {\phi}\left( \boldsymbol{x}_N; \boldsymbol{\theta} \right) \\
		s.t. \quad \boldsymbol{x}_{k+1} &= \boldsymbol{f} \left( \boldsymbol{x}_k, \boldsymbol{u}_k; \boldsymbol{\theta}\right) \\
	\end{aligned}
	\label{eq:prob1}
\end{equation}
where the semicolon separating the parameter $\boldsymbol{\theta}$ from the state $\boldsymbol{x}$ and the control $\boldsymbol{u}$ is to show that $\boldsymbol{\theta}$ is a time-invariant parameter, independent of the time instant $k$.
%Consider the discretized dynamics model of a nonlinear system
%\begin{equation}
%	\boldsymbol{x}_{k+1} = \boldsymbol{f} \left( \boldsymbol{x}_k, \boldsymbol{u}_k \right)
%	\label{eq:dissys1}
%\end{equation}
%
%Let $\boldsymbol{U}_k = \left\{ \boldsymbol{u}_k,\boldsymbol{u}_{k+1}, \ldots ,\boldsymbol{u}_{N - 1}  \right\}$ be the partial control sequence. 

%The cost-to-go $J_k$ at time instant $t_k$ is defined as
%\begin{equation}
%J_k= \sum\limits_{\kappa = k}^{N - 1} {l \left( \boldsymbol{x}_\kappa,\boldsymbol{u}_\kappa \right)} + {\phi }\left( \boldsymbol{x}_N, t_N \right)
%\label{eq:k4}
%\end{equation}

%control sequence $\boldsymbol{U}_0$ control input $\boldsymbol{u}_k$
{\colb The basic idea of the original PDDP is to transform the problem of optimizing the entire trajectory into the backward sequential optimization problem for each time instant using Bellman's optimality principle.
%The key idea of the original PDDP is to convert the problem of optimizing a trajectory over an entire time history into the problem of optimizing a single time instant backward sequentially using the Bellman optimality principle.
%The parameterized value function, also defined as the parameterized optimal cost-to-go, is given by
The parameterized optimal cost-to-go, which can be also defined as the parameterized value function, is provided as}
\begin{equation}
		V\left(\boldsymbol{x}_{k}; \boldsymbol{\theta}\right) =  \min [\underbrace{l\left(\boldsymbol{x}_{k}, \boldsymbol{u}_{k}; \boldsymbol{\theta}\right)+V\left(\boldsymbol{x}_{k+1};\boldsymbol{\theta}\right)}_{Q\left(\boldsymbol{x}_{k}, \boldsymbol{u}_{k}; \boldsymbol{\theta}\right)}]
\label{eq: bell}
\end{equation}
{\colb and the action-value function is specified as $Q\left(\boldsymbol{x}_{k}, \boldsymbol{u}_{k}; \boldsymbol{\theta}\right)$.}
%where $Q\left(\boldsymbol{x}_{k}, \boldsymbol{u}_{k}; \boldsymbol{\theta}\right)$ is defined as action-value function.
%The above are the update rules in the control history but not the terminal instant $t_N$.

Unlike the standard DDP that only considers quadratic approximations with respect to the state $\boldsymbol{x}_k$ and the control $\boldsymbol{u}_k$ along a nominal trajectory $\left({\boldsymbol{x}}_{k}, {\boldsymbol{u}}_{k}\right)$ at time $t_{k}$, the parameterized DDP also needs to consider quadratic approximations of the value function with respect to parameter $\boldsymbol{\theta}$, which gives
\begin{equation}
	\begin{aligned}
		V\left(\boldsymbol{x}_k + \delta \boldsymbol{x}_k ; \boldsymbol{\theta} + \delta \boldsymbol{\theta}\right) \approx & V\left(\boldsymbol{x}_k; \boldsymbol{\theta}\right)+V_{\boldsymbol{x},k}^{\mathrm{T}} \delta \boldsymbol{x}_k +V_{\theta,k}^{\mathrm{T}} \delta \boldsymbol{\theta} \\
		& +\frac{1}{2}\left[\begin{array}{c}
								\delta \boldsymbol{x}_k \\
								\delta \boldsymbol{\theta}
							\end{array}\right]^{\mathrm{T}}\left
							[\begin{array}{ll}
								V_{\boldsymbol{x} \boldsymbol{x},k} & V_{\boldsymbol{x} \boldsymbol{\theta},k} \\
								V_{\boldsymbol{\theta} \boldsymbol{x},k} & V_{\boldsymbol{\theta} \boldsymbol{\theta},k}
							\end{array}\right]
							\left[\begin{array}{c}
								\delta \boldsymbol{x}_k \\
								\delta \boldsymbol{\theta}
							\end{array}\right]
	\end{aligned}
	\label{eq:vtn1}
\end{equation}
where {\colb the state, control and parameter deviations at $t_k$ from the nominal trajectory is denoted as $\delta\boldsymbol{x}_{k}$, $\delta\boldsymbol{u}_{k}$ and $\delta\boldsymbol{\theta}$. $V_{\boldsymbol{x},k}$, $V_{\theta,k}$ are the gradient vectors and $V_{\boldsymbol{x} \boldsymbol{x},k}$, $V_{\boldsymbol{x} \boldsymbol{\theta},k}$, $V_{\boldsymbol{\theta} \boldsymbol{x},k}$, $V_{\boldsymbol{\theta} \boldsymbol{\theta},k}$ are the Hessian at time instant $k$ of the value function $V\left(\boldsymbol{x}_k; \boldsymbol{\theta}\right)$.}

{\colb Likewise, $Q\left(\boldsymbol{x}_{k}, \boldsymbol{u}_{k}; \boldsymbol{\theta}\right)$ defined in Eq.~\eqref{eq: bell} has a quadratic approximation that is provided by}
\begin{equation}
	\begin{aligned}
		&Q({\boldsymbol{x}}_{k} + \delta \boldsymbol{x}_{k}, {\boldsymbol{u}}_{k} + \delta \boldsymbol{u}_{k}; {\boldsymbol{\theta}} + \delta \boldsymbol{\theta}) \approx Q({\boldsymbol{x}}_{k}, {\boldsymbol{u}}_{k}; {\boldsymbol{\theta}}) + Q_{\boldsymbol{x},k}^{\mathrm{T}} \delta \boldsymbol{x}_{k} + Q_{\boldsymbol{u},k}^{\mathrm{T}} \delta \boldsymbol{u}_{k} + Q_{\boldsymbol{\theta},k}^{\mathrm{T}} \delta \boldsymbol{\theta}\\
		&\quad\quad\quad \quad\quad\quad\quad\quad \quad\quad+\frac{1}{2}\left[\begin{array}{l} \delta \boldsymbol{x}_{k} \\ \delta \boldsymbol{u}_{k} \\ \delta \boldsymbol{\theta} \end{array}\right]^{\mathrm{T}}
		\left[\begin{array}{lll} Q_{\boldsymbol{x} \boldsymbol{x},k} & Q_{\boldsymbol{x} \boldsymbol{u},k} & Q_{\boldsymbol{x} \boldsymbol{\theta},k} \\ Q_{\boldsymbol{u} \boldsymbol{x},k} & Q_{\boldsymbol{u} \boldsymbol{u},k} & Q_{\boldsymbol{u} \boldsymbol{\theta},k} \\ Q_{\boldsymbol{\theta} \boldsymbol{x},k} & Q_{\boldsymbol{\theta} \boldsymbol{u},k} & Q_{\boldsymbol{\theta} \boldsymbol{\theta},k} \end{array}\right]
		\left[\begin{array}{l} \delta \boldsymbol{x}_{k} \\ \delta \boldsymbol{u}_{k} \\ \delta \boldsymbol{\theta} \end{array}\right] 
	\end{aligned}
	\label{eq:qapp}
\end{equation}

{\colb Organizing the same order terms of $\delta \boldsymbol{x}$, $\delta \boldsymbol{u}$ and $\delta \boldsymbol{\theta}$ will provide the derivatives of ${Q}\left( \boldsymbol{x}_k,\boldsymbol{u}_k; \boldsymbol{\theta}\right)$ as}
%The gradient vectors and Hessian matrices of the action value function ${Q}\left( \boldsymbol{x}_k,\boldsymbol{u}_k; \boldsymbol{\theta}\right) $ can be readily obtained by the definition and grouping the corresponding terms with the same order of $\delta \boldsymbol{x}$, $\delta \boldsymbol{u}$ and $\delta \boldsymbol{\theta}$ yields
\begin{equation}
\begin{split}
& Q_{\boldsymbol{x}, k}= l_{\boldsymbol{x}, k} + \boldsymbol{f}_{\boldsymbol{x}, k}^{\mathrm{T}} V_{\boldsymbol{x},k+1}\\ 
& Q_{\boldsymbol{u}, k}= l_{\boldsymbol{u}, k} + \boldsymbol{f}_{\boldsymbol{u}, k}^{\mathrm{T}} V_{\boldsymbol{x},k+1}\\ 
& Q_{\boldsymbol{\theta}, k}= l_{\boldsymbol{\theta}, k} + V_{\boldsymbol{\theta}, k+1} + \boldsymbol{f}_{\boldsymbol{\theta}, k}^{\mathrm{T}} V_{\boldsymbol{x},k+1}\\ 
& Q_{\boldsymbol{x} \boldsymbol{x}, k} = l_{\boldsymbol{x} \boldsymbol{x}, k} + \boldsymbol{f}_{\boldsymbol{x}, k}^{\mathrm{T}} V_{\boldsymbol{x} \boldsymbol{x},k+1} \boldsymbol{f}_{\boldsymbol{x}, k}\\
%+V_{\boldsymbol{x}}\left(\boldsymbol{x}_{k+1}\right) \boldsymbol{f}_{\boldsymbol{x} \boldsymbol{x}} \left(\boldsymbol{x}_{k}, \boldsymbol{u}_{k}\right)\\ 
& Q_{\boldsymbol{u} \boldsymbol{u}, k} =l_{\boldsymbol{u} \boldsymbol{u}, k} +\boldsymbol{f}_{\boldsymbol{u}, k}^{\mathrm{T}} V_{\boldsymbol{x} \boldsymbol{x},k+1} \boldsymbol{f}_{\boldsymbol{u}, k}\\
%+V_{\boldsymbol{x}}\left(\boldsymbol{x}_{k+1}\right)\boldsymbol{f}_{\boldsymbol{u} \boldsymbol{u}}\left(\boldsymbol{x}_{k}, \boldsymbol{u}_{k}\right)\\
& Q_{\boldsymbol{\theta} \boldsymbol{\theta}, k} =l_{\boldsymbol{\theta} \boldsymbol{\theta}, k}  + V_{\boldsymbol{\theta} \boldsymbol{\theta}, k+1} +\boldsymbol{f}_{\boldsymbol{\theta}, k}^{\mathrm{T}} V_{\boldsymbol{x} \boldsymbol{x},k+1} \boldsymbol{f}_{\boldsymbol{\theta}, k} +\boldsymbol{f}_{\boldsymbol{\theta}, k}^{\mathrm{T}} V_{\boldsymbol{x} \boldsymbol{\theta},k+1} + V_{\boldsymbol{\theta} \boldsymbol{x},k+1} \boldsymbol{f}_{\boldsymbol{\theta}, k}\\
&Q_{\boldsymbol{x} \boldsymbol{u}, k} = l_{\boldsymbol{x} \boldsymbol{u}, k} + \boldsymbol{f}_{\boldsymbol{x}, k}^{\rm{T}} V_{\boldsymbol{x} \boldsymbol{x},k+1} \boldsymbol{f}_{\boldsymbol{u}, k} = Q_{\boldsymbol{u} \boldsymbol{x},k}^{\mathrm{T}}\\
%+V_{\boldsymbol{x}}\left(\boldsymbol{x}_{k}, \boldsymbol{u}_{k}\right)\boldsymbol{f}_{\boldsymbol{u} \boldsymbol{x}}\left(\boldsymbol{x}_{k}, \boldsymbol{u}_{k}\right) \\ 
&Q_{\boldsymbol{x} \boldsymbol{\theta}, k} = l_{\boldsymbol{x} \boldsymbol{\theta}, k} +\boldsymbol{f}_{\boldsymbol{x}, k}^{\rm{T}} V_{\boldsymbol{x} \boldsymbol{x},k+1} \boldsymbol{f}_{\boldsymbol{\theta}, k} +\boldsymbol{f}_{\boldsymbol{x}, k}^{\rm{T}} V_{\boldsymbol{x} \boldsymbol{\theta},k+1} = Q_{\boldsymbol{\theta} \boldsymbol{x},k}^{\mathrm{T}}\\
&Q_{\boldsymbol{u} \boldsymbol{\theta}, k} = l_{\boldsymbol{u} \boldsymbol{\theta}, k} + \boldsymbol{f}_{\boldsymbol{u}, k}^{\rm{T}} V_{\boldsymbol{x} \boldsymbol{x},k+1} \boldsymbol{f}_{\boldsymbol{\theta}, k} +\boldsymbol{f}_{\boldsymbol{u}, k}^{\rm{T}} V_{\boldsymbol{u} \boldsymbol{\theta},k+1} = Q_{\boldsymbol{\theta} \boldsymbol{u},k}^{\mathrm{T}}\\
\end{split}
\label{eq:k9}
\end{equation}
%Notice that the calculation of the second-order derivatives of the dynamics, i.e., $\boldsymbol{f}_{\boldsymbol{x} \boldsymbol{x}}$, $\boldsymbol{f}_{\boldsymbol{x} \boldsymbol{u}}$, $\boldsymbol{f}_{\boldsymbol{u} \boldsymbol{u}}$, is computationally expensive and therefore DDP ignores these terms. identical as Eq. \eqref{eq:k9}. 

The second order expansion of $Q(\boldsymbol{x}_k, \boldsymbol{u}_k; \boldsymbol{\theta})$ is {\colb a quadratic function w.r.t. control deviation $\delta \boldsymbol{u}_k$, then necessary optimality condition can be utilized for deriving $\delta \boldsymbol{u}_k^*$ as}
\begin{align}
	\delta \boldsymbol{u}_{k}^{*}=
	\boldsymbol{k}_k + \boldsymbol{K}_k \delta \boldsymbol{x}_{k} + \boldsymbol{M}_k \delta \boldsymbol{\theta}_k
\label{eq:sol1}
\end{align}
where
\begin{eqnarray}
	\begin{aligned}
		\boldsymbol{k}_k &= -Q_{\boldsymbol{u} \boldsymbol{u},k}^{-1} Q_{\boldsymbol{u},k}\\
		\boldsymbol{K}_k &= -Q_{\boldsymbol{u} \boldsymbol{u},k}^{-1} Q_{\boldsymbol{u} \boldsymbol{x},k}\\
		\boldsymbol{M}_k &= -Q_{\boldsymbol{u} \boldsymbol{u},k}^{-1} Q_{\boldsymbol{u} \boldsymbol{\theta},k}
	\end{aligned}
	\label{eq:terms}
\end{eqnarray}
where the feedforward term $\boldsymbol{k}_k$ and the feedback term $\boldsymbol{K}_k$ are the same as in the original DDP method, with the addition of an extra feedback term $\boldsymbol{M}_k$ for the optimization of $\boldsymbol{\theta}$ in PDDP.

Since parameter $\boldsymbol{\theta}$ is time-independent, i.e., it does not vary with time, it only needs to be updated at the time instant $k=1$. In other words, the expression for optimal parameter increment $\delta {\boldsymbol{\theta}}^*$ can be obtained with the condition that the deviation of $\delta \boldsymbol{x}_1 = 0$ for Eq. \eqref{eq:vtn1} as follows
\begin{align}
	\delta \boldsymbol{\theta}^{*}= -V_{\boldsymbol{\theta} \boldsymbol{\theta},1}^{-1} V_{\boldsymbol{\theta},1}
\label{eq:sol_theta}
\end{align}

To ensure the convergence of the PDDP algorithm, the feedforward terms $\boldsymbol{k}_k$ of $\delta \boldsymbol{u}_{k}^{*}$ and $\boldsymbol{M}_k$ of $\delta \boldsymbol{\theta}^{*}$ need to be scaled by the damping coefficients $\alpha_l$, where the $\alpha_l$ can be determined by the line search method \cite{oshin2022parameterized}, and there is
\begin{align}
	\delta \boldsymbol{u}_{k}^{*} &= \alpha_l \boldsymbol{k}_k + \boldsymbol{K}_k \delta \boldsymbol{x}_{k} + \boldsymbol{M}_k \delta \boldsymbol{\theta}_k\\
	\delta \boldsymbol{\theta}^{*} &= - \alpha_l V_{\boldsymbol{\theta} \boldsymbol{\theta},1}^{-1} V_{\boldsymbol{\theta},1}
\label{eq:sol}
\end{align}

{\colb Substituting optimal control variation \eqref{eq:sol1} into the Taylor expansion of $Q(\boldsymbol{x}_k, \boldsymbol{u}_k; \boldsymbol{\theta})$ and grouping the same order terms of $\delta \boldsymbol{x}_k$ and $\delta \boldsymbol{\theta}$ yields}
\begin{equation}
	\begin{aligned}
		V\left(\boldsymbol{x}_{k};\boldsymbol{\theta}\right) &=Q\left({\boldsymbol{x}}_{k}, {\boldsymbol{u}}_{k}; \boldsymbol{\theta}\right) -(\frac{1}{2}\alpha_l^2-\alpha_l) Q_{\boldsymbol{u},k}^{\mathrm{T}} Q_{\boldsymbol{u} \boldsymbol{u},k}^{-1} Q_{\boldsymbol{u},k}\\
		V_{\boldsymbol{x},k} &=Q_{\boldsymbol{x},k}-Q_{\boldsymbol{x} \boldsymbol{u},k} Q_{\boldsymbol{u} \boldsymbol{u},k}^{-1} Q_{\boldsymbol{u},k} \\
		V_{\boldsymbol{\theta},k} &=Q_{\boldsymbol{\theta},k}-Q_{\boldsymbol{\theta} \boldsymbol{u},k} Q_{\boldsymbol{u} \boldsymbol{u},k}^{-1} Q_{\boldsymbol{u},k} \\
		V_{\boldsymbol{x} \boldsymbol{x},k} &=Q_{\boldsymbol{x} \boldsymbol{x},k}-Q_{\boldsymbol{x} \boldsymbol{u},k} Q_{\boldsymbol{u} \boldsymbol{u},k}^{-1} Q_{\boldsymbol{u} \boldsymbol{x},k}\\
		V_{\boldsymbol{x} \boldsymbol{\theta},k} &=Q_{\boldsymbol{x} \boldsymbol{\theta},k}-Q_{\boldsymbol{x} \boldsymbol{u},k} Q_{\boldsymbol{u} \boldsymbol{u},k}^{-1} Q_{\boldsymbol{u} \boldsymbol{\theta},k}
		= V_{\boldsymbol{\theta} \boldsymbol{x},k}^{\mathrm{T}}\\
		V_{\boldsymbol{\theta} \boldsymbol{\theta},k} &=Q_{\boldsymbol{\theta} \boldsymbol{\theta},k}-Q_{\boldsymbol{\theta} \boldsymbol{u},k} Q_{\boldsymbol{u} \boldsymbol{u},k}^{-1} Q_{\boldsymbol{u} \boldsymbol{\theta},k}
	\end{aligned}
	\label{eq:sol2}
\end{equation}

{\colb Before running the PDDP algorithm, control effort initial guess should be given for generating the nominal trajectory. Afterwards, the terminal value function is initialized using the terminal objective function as $V\left(\boldsymbol{x}_{N};\boldsymbol{\theta}\right) = {\phi}(\boldsymbol{x}_N;\boldsymbol{\theta})$, $V_{\boldsymbol{x},N} = {\phi}_{\boldsymbol{x},N}$, $V_{\boldsymbol{x} \boldsymbol{x},N} = {\phi}_{\boldsymbol{x} \boldsymbol{x},N}$, $V_{\boldsymbol{\theta},N} = {\phi}_{\boldsymbol{\theta},N}$, $V_{\boldsymbol{\theta} \boldsymbol{\theta},N} = {\phi}_{\boldsymbol{\theta} \boldsymbol{\theta},N}$, $V_{\boldsymbol{x} \boldsymbol{\theta},N} = {\phi}_{\boldsymbol{x} \boldsymbol{\theta},N}=V_{\boldsymbol{\theta} \boldsymbol{x},N}^{\mathrm{T}}$. Then the optimal control correction $\delta \boldsymbol{u}_k$ is computed using Eq. \eqref{eq:sol1} in backward order while the optimal parameter change $\delta \boldsymbol{\theta}$ is computed using Eq. \eqref{eq:sol_theta} at time instant $k=1$.
%Given an initial solution guess of the control sequence, PDDP initializes the terminal value function using $V\left(\boldsymbol{x}_{N};\boldsymbol{\theta}\right) = {\phi}(\boldsymbol{x}_N;\boldsymbol{\theta})$, $V_{\boldsymbol{x},N} = {\phi}_{\boldsymbol{x},N}$, $V_{\boldsymbol{x} \boldsymbol{x},N} = {\phi}_{\boldsymbol{x} \boldsymbol{x},N}$, $V_{\boldsymbol{\theta},N} = {\phi}_{\boldsymbol{\theta},N}$, $V_{\boldsymbol{\theta} \boldsymbol{\theta},N} = {\phi}_{\boldsymbol{\theta} \boldsymbol{\theta},N}$, $V_{\boldsymbol{x} \boldsymbol{\theta},N} = {\phi}_{\boldsymbol{x} \boldsymbol{\theta},N}=V_{\boldsymbol{\theta} \boldsymbol{x},N}^{\mathrm{T}}$, and runs a backward process from $k = N$ to $k=1$ to sequentially find the optimal control correction $\delta \boldsymbol{u}_k$ using Eq. \eqref{eq:sol1} and to find the optimal parameter variation $\delta \boldsymbol{\theta}$ using Eq. \eqref{eq:sol_theta} at $k=1$. 
%Then ${\boldsymbol{u}}_{k}=\boldsymbol{u}_{k}+\delta \boldsymbol{u}_{k}^{*}$, ${\boldsymbol{\theta}}=\boldsymbol{\theta}+\delta \boldsymbol{\theta}^{*}$ and the parameterized discrete dynamic \eqref{eq:dissys1} can be utilized in the forward pass to generate new trajectory.
In brief, the one-step control update and one-iteration parameter update for generating a new trajectory are given by}
\begin{equation}
	\begin{aligned}
		{\boldsymbol{u}}_{k} &=\boldsymbol{u}_{k}+\delta \boldsymbol{u}_{k}^{*} =\boldsymbol{u}_{k} + \alpha_l \boldsymbol{k}_k + \boldsymbol{K}_k \delta \boldsymbol{x}_{k} + \boldsymbol{M}_k \delta \boldsymbol{\theta}_k\\
		{\boldsymbol{\theta}}&=\boldsymbol{\theta}+\delta \boldsymbol{\theta}^{*} = \boldsymbol{\theta} - \alpha_l V_{\boldsymbol{\theta} \boldsymbol{\theta},1}^{-1} V_{\boldsymbol{\theta},1}
	\end{aligned}
	\label{eq:uup}
\end{equation}

%After the optimal control updates and optimal parameter updates are computed in the backward process, the forward process is triggered to generate a new nominal trajectory.
{\colb The forward process is initiated to produce another nominal trajectory once the backward process has computed the optimal control updates and optimal parameter updates. The PDDP method performs the backward and forward processes iteratively until the cost function's error between two iterations is smaller than a preset threshold, that is,}
%the error of the cost function between two iterations is less than a predefined threshold, i.e.,
%Once the backward pass is completed, a forward pass is triggered to find a new nominal state trajectory by substituting the updated control sequence and parameter into parameterized system dynamics. The PDDP method iteratively runs the backward pass and forward pass until the convergence condition is triggered. A commonly used termination criterion is the difference of cost function between two iterations is smaller than a lower threshold, i.e., 
\begin{equation}
\left|J^{(n)} - J^{(n-1)}\right|<\epsilon
\label{eq:ter}
\end{equation}
{\colb in which the constant $\epsilon>0$ has a comparatively low value and $J^{(n)}$ indicates the $n$-th iteration's objective function.} Once the termination condition is satisfied, the iteration process of the algorithm is terminated and the optimal state and controls are given.
%where $J^{(n)}$ denotes the cost function for the $n$-th iteration and $\epsilon>0$ is a constant with a very small value. 
%Notice that tuning the termination threshold factor $\epsilon$ is a tradeoff between the optimization performance and computational efficiency: a smaller value ensures accurate convergence performance but a higher computational burden.
The pseudo-code of PDDP is presented in Algorithm \ref{algo: PDDP} as follows:
\begin{algorithm}[h!]
%\textsl{}\setstretch{1.8}
\renewcommand{\algorithmicrequire}{\textbf{Input:}}
\renewcommand{\algorithmicensure}{\textbf{Output:}}
\caption{PDDP algorithm workflow}
\label{algo: PDDP}
\begin{algorithmic}[1]
	\STATE Initialize $\boldsymbol{U}_0=\left\{\boldsymbol{u}_0, \ldots, \boldsymbol{u}_{k}, \ldots,\boldsymbol{u}_{N-1}\right\}, \boldsymbol{\theta}_0$
	\FOR{$k = 0, ..., N-1$} 
		\STATE ${\boldsymbol{x}}_{k+1} = f( {\boldsymbol{x}}_{k}, {\boldsymbol{u}}_{k}; \boldsymbol{\theta})$
	\ENDFOR
	
	\STATE $J_0$ $\gets$ Eq.\eqref{eq:prob1}
	\WHILE{$\left|J_{i} - J_{i-1}\right|\ge\epsilon$}
		\STATE {$V\left(\boldsymbol{x}_{N};\boldsymbol{\theta}\right) = {\phi}(\boldsymbol{x}_N;\boldsymbol{\theta})$, $V_{\boldsymbol{x},N} = {\phi}_{\boldsymbol{x},N}$, $V_{\boldsymbol{x} \boldsymbol{x},N} = {\phi}_{\boldsymbol{x} \boldsymbol{x},N}$
		\STATE $V_{\boldsymbol{\theta},N} = {\phi}_{\boldsymbol{\theta},N}$, $V_{\boldsymbol{\theta} \boldsymbol{\theta},N} = {\phi}_{\boldsymbol{\theta} \boldsymbol{\theta},N}$, $V_{\boldsymbol{x} \boldsymbol{\theta},N} = {\phi}_{\boldsymbol{x} \boldsymbol{\theta},N}=V_{\boldsymbol{\theta} \boldsymbol{x},N}^{\mathrm{T}}$}
		
		\FOR{$k = N-1, ..., 0$}
			\STATE Using Eq.\eqref{eq:k9} to calculate the Jacobian and Hessian of $Q\left({\boldsymbol{x}}_{k}, {\boldsymbol{u}}_{k}; \boldsymbol{\theta}\right)$

			\STATE ${\boldsymbol{k}}_k, {\boldsymbol{K}}_k,{\boldsymbol{M}}_k$ $\gets$ Eq.\eqref{eq:terms}
			\STATE Using Eq.\eqref{eq:sol2} to calculate the derivatives of ${V}\left({\boldsymbol{x}}_{k}, {\boldsymbol{u}}_{k}; \boldsymbol{\theta}\right)$
			
			\IF{$k = 1$ }
				\STATE $\delta \boldsymbol{\theta}^{*}= -V_{\boldsymbol{\theta \theta}}^{-1} ( V_{\boldsymbol{\theta}} + V_{\boldsymbol{\theta} \boldsymbol{x}}^{\mathrm{T}} \delta \boldsymbol{x})$
			\ENDIF
		\ENDFOR

		\STATE $\boldsymbol{\theta}$ $\gets$ Eq.\eqref{eq:uup}

		\FOR{$k = 0, ..., N-1$} 
			\STATE ${\delta \boldsymbol{x}}_{k} = \boldsymbol{x}_{k}^{\mathrm{new}} - \boldsymbol{x}_{k}$
			%\STATE ${\boldsymbol{u}}_{k}$ $\gets$ Eq.\eqref{eq:usol}
			\STATE ${\boldsymbol{u}}_{k} =\min(\max( {\boldsymbol{u}}_{k} + {\boldsymbol{k}}_k + {\boldsymbol{K}}_k \delta \boldsymbol{x}_{k} + {\boldsymbol{M}}_k \delta \boldsymbol{\theta}, \boldsymbol{u}_{\min}), \boldsymbol{u}_{\max})$
			\STATE $ {\boldsymbol{x}}_{k+1}^{\mathrm{new}} = f( {\boldsymbol{x}}_{k}, {\boldsymbol{u}}_{k}; \boldsymbol{\theta} ) $
		\ENDFOR
		
		\STATE ${\boldsymbol{X}} = {\boldsymbol{X}}^{\mathrm{new}}$
	\ENDWHILE 
\end{algorithmic}  
\end{algorithm}

\begin{remark}
Consider the terminal time $t_{N}$ as an unknown parameter $\boldsymbol{\theta}$, the PDDP is able to find the optimal terminal time and the control sequence. However, it should be noted the forward pass of PDDP requires to integrate the system dynamics from the initial time instant until $t_{N}$. For reason, we artificially change the argument from $t$ to a $\iota_{k} =\frac{t_{k} - t_{i,1}}{t_{N}}\in [0,1]$. With this modification, the optimization problem can then be reformulated as
	\begin{equation}
		\begin{aligned}
			\underset{\boldsymbol{U}; t_{N}}{\operatorname{\min}} \ J(\boldsymbol{U}; t_{N}) &= \sum\limits_{k = 1}^{N - 1} {t_{i, N} l \left( \boldsymbol{x}_k(\iota_{k}),\boldsymbol{u}_k(\iota_{k})\right)} + {\phi}\left( \boldsymbol{x}_N(\iota_{N}) \right) \\
			s.t. \quad \boldsymbol{x}_{k+1} &= t_{N} \boldsymbol{f} \left( \boldsymbol{x}_k(\iota_{k}), \boldsymbol{u}_k(\iota_{k})\right) \\
		\end{aligned}
		\label{eq:probscal}
	\end{equation}
\end{remark}

\subsection{Alternating Direction Method of Multiplier}
%Subsequently, t
This subsection provides a brief description of the ADMM algorithm \cite{MAL-016}. ADMM has integrated the decomposability of the dual ascent method and the excellent convergence properties of multiplier method. {\colm The basic idea of ADMM is to decompose the optimization of the objective function into a number of sub-problems optimized on different variables or substructures, and then through a coordination mechanism (through the introduction of Lagrangian multipliers and dual variables) to ensure that the solutions of each subproblem converge to the global optimum.} The resulting method allows parallel optimization under more general conditions and the standard form is as
\begin{equation}
	\begin{aligned}
		& \min _{\boldsymbol{\varphi}, \boldsymbol{\psi}} H(\boldsymbol{\varphi})+G(\boldsymbol{\psi}) \\
		& \text { s.t. } \boldsymbol{\alpha} \boldsymbol{\varphi}+\boldsymbol{\beta} \boldsymbol{\psi}=\boldsymbol{\delta}
	\end{aligned}
	\label{adprob}
\end{equation}
{\colb In the standard ADMM form, the primal variables are $\boldsymbol{\varphi} \in \mathbb{R}^{n}$, $\boldsymbol{\psi} \in \mathbb{R}^{m}$, and the functions of $\boldsymbol{\varphi}$ and $\boldsymbol{\psi}$ are defined as $H:\mathbb{R}^{n} \to \mathbb{R}, G : \mathbb{R}^{m} \to \mathbb{R}$, and the coefficient matrix of $\boldsymbol{\varphi}$ and $\boldsymbol{\psi}$ in the constraint is $A \in \mathbb{R}^{r \times n}, B \in \mathbb{R}^{r\times m}$, and $\boldsymbol{\delta} \in \mathbb{R}^{r}$ is a constant vector.}
%where $\boldsymbol{\varphi} \in \mathbb{R}^{n}$, $\boldsymbol{\psi} \in \mathbb{R}^{m}$ are the primal variables, $H:\mathbb{R}^{n} \to \mathbb{R}, G : \mathbb{R}^{m} \to \mathbb{R}, A \in \mathbb{R}^{r \times n}, B \in \mathbb{R}^{r\times m}$, and $\boldsymbol{\delta} \in \mathbb{R}^{r}$.

The augmented Lagrange multipliers formation of problem \eqref{adprob} can be provided as
\begin{equation}
	\begin{aligned}
		\mathcal{L}(\boldsymbol{\varphi}, \boldsymbol{\psi}, \boldsymbol{\lambda})= & H(\boldsymbol{\varphi})+G(\boldsymbol{\psi})+\boldsymbol{\lambda}^{\mathrm{T}}(\boldsymbol{\alpha} \boldsymbol{\varphi}+\boldsymbol{\beta} \boldsymbol{\psi}-\boldsymbol{\delta}) \\
		& +\frac{\vartheta}{2}\|\boldsymbol{\alpha} \boldsymbol{\varphi}+\boldsymbol{\beta} \boldsymbol{\psi}-\boldsymbol{\delta}\|_2^2
	\end{aligned}
\end{equation}
{\colb in which the dual variable of the constraint is $\boldsymbol{\lambda} \in \mathbb{R}^{r}$ and the penalty parameter of it is $\vartheta>0$.}
%where $\boldsymbol{\lambda} \in \mathbb{R}^{r}$ is the dual variable and $\vartheta>0$ is the penalty parameter.
%for the constraint $\boldsymbol{\alpha}\boldsymbol{\varphi} + \boldsymbol{\beta}\boldsymbol{\psi} - \boldsymbol{\delta} =\boldsymbol{0}$

The update step for iterative optimization of the algorithm is given by:
\begin{equation}
	\begin{aligned}
		\boldsymbol{\varphi}^{n+1}&=\underset{\boldsymbol{\varphi}}{\arg \min } \mathcal{L}\left(\boldsymbol{\varphi}, \boldsymbol{\psi}^n, \boldsymbol{\lambda}^n\right)\\
		\boldsymbol{\psi}^{n+1}&=\underset{\boldsymbol{\psi}}{\arg \min } \mathcal{L}\left(\boldsymbol{\varphi}^{n+1}, \boldsymbol{\psi}, \boldsymbol{\lambda}^n\right)\\
		\boldsymbol{\lambda}^{n+1}&=\boldsymbol{\lambda}^{n}+\vartheta(\boldsymbol{\alpha}\boldsymbol{\varphi}^{n+1}+\boldsymbol{\beta}\boldsymbol{\psi}^{n+1}-\boldsymbol{\delta})
	\end{aligned}
\end{equation}
%denotes the algorithm iteration.
{\colb where $n$ represents the number of iterations of the algorithm. The standard form of ADMM described above contains only two variables $\boldsymbol{\varphi}$ and $\boldsymbol{\psi}$, but a multiblock form containing more than three variables can be developed via the ADMM method \cite{8814732}.}
%The ADMM method can be extended to a multiblock version with variables $x_{1},\ldots,x_{M}$, $M \geq 3$ that are updated in a sequential manner \cite{8814732}.

%%%%%%%%%%%%%%%%%%%%%%%%%%%%
\section{Distributed Parameterized Differential Dynamic Programming} \label{sec: innovation}
%In this section, the details of the proposed D-PDDP algorithm is presented. 
{\colb This section provides a comprehensive explanation of D-PDDP algorithm.} We first reformulate the problem to make it suitable in using distributed optimization and then derive the algorithm using the concept of ADMM. In the framework of free-final-time spatial-temporal trajectory optimization, the final time $t_{i,N}$ of each agent is chosen as the unknown parameter $\boldsymbol{\theta}$ in the PDDP. Therefore, $t_{i,N}$ is used as the symbolic representation of the spatial-temporal trajectory optimization in the following derivation and as the symbolic representation of unknown parameter in the PDDP algorithm in this section.

\subsection{Problem Reformulation} \label{ssec:D-PDDP}
The problem \eqref{eq: MAOC1} is a general form of the multi-agent system problem, {\colb which is necessary to be transformed to optimize it using the distributed optimization method \cite{MAL-016}.}
%and it is necessary to convert it into a form that is suitable for distributed optimization \cite{MAL-016}. 
Since the local spatial-temporal constraints need to be solved by the ADMM method in the two-layer framework, {\colb the `safe' copy variable $\tilde{\boldsymbol{x}}_{i,k} \in \mathbb{R}^{p_i}$, $\tilde{\boldsymbol{u}}_{i,k} \in \mathbb{R}^{q_i}$, $\tilde{t}_{i,N} \in \mathbb{R}$ of ${\boldsymbol{x}}_{i,k}$, ${\boldsymbol{u}}_{i,k}$ and ${t}_{i,N}$ are introduced first,} which should consensus with the actual trajectory computed by the {\col PDDP} algorithm, as
\begin{equation}
	\tilde{\boldsymbol{x}}_{i, k}=\boldsymbol{x}_{i, k}, \tilde{\boldsymbol{u}}_{i, k}=\boldsymbol{u}_{i, k}, \tilde{t}_{i,N}=t_{i,N}
\end{equation}
and they should also satisfy the {\colb local constraints of a single agent, i.e., $\boldsymbol{a}_{i,k} (\tilde{x}_{i,k}) \leq 0$, $\boldsymbol{b}_{i,k}\left(\tilde{\boldsymbol{u}}_{i, k}\right) \leq 0$ and $c_{i, N}\left(\tilde{t}_{i, N}\right) \leq 0$.}

Since the inter-agent constraints \eqref{eq:intercons1} and \eqref{eq:intercons2} introduce the inter-couplings among the agents, indicating that problem \eqref{eq: MAOC1} is difficult to solve in a decentralized manner. For this reason, each agent within the proposed architecture is required to keep a `safe' copy of its variables and those of its neighbors to achieve consensus. Specifically, the state variable and final time of $j$ that under the viewpoint of $i$ is defined as: $\tilde{\boldsymbol{x}}^{i}_{j,k} \in \mathbb{R}^{p_{j}}$, $\tilde{t}^{i}_{j,N} \in \mathbb{R}$, $j \in \mathcal{N}_i$, $i \in \mathcal{M}$, i.e., $\tilde{\boldsymbol{x}}^{i}_{j,k}$ is a variable that is relatively safe and $\tilde{t}^{i}_{j,N}$ is a proper final time of agent $j$ from agent $i$'s perspective. Thus, augmented state and time variable, which are stacked from all neighbors of agent $i$, can be defined as
\begin{equation}
	\tilde{\boldsymbol{x}}_{i, k}^{\mathrm{a}}=\left[\left\{\tilde{\boldsymbol{x}}_{j, k}^i\right\}_{j \in \mathcal{N}_i}\right] \in \mathbb{R}^{\tilde{p}_i}, \tilde{p}_i=\sum_{j \in \mathcal{N}_i} p_j
\end{equation}
\begin{equation}
	\tilde{\boldsymbol{t}}_{i, N}^{\mathrm{a}}=\left[\left\{\tilde{t}_{j, N}^i\right\}_{j \in \mathcal{N}_i}\right] \in \mathbb{R}^{|\mathcal{N}_i|}
\end{equation}
%where the definition of state history $\tilde{\boldsymbol{x}}^{i}_{j}$, $\tilde{\boldsymbol{x}}_{i}^{\mathrm{a}}$ are the same as $\boldsymbol{x}_{i}$ in \eqref{eq:consys} while the $\tilde{t}_{j}^i$, $\tilde{t}^{a}_i$ are the same as $t_{i}$. 

%By introducing the copy variables, the inter-agent constraints \eqref{eq:intercons1} and \eqref{eq:intercons2} can be rewritten from the perspective of each agent $i$ as 
{\colb Constraints \eqref{eq:intercons1} and \eqref{eq:intercons2} should be reviewed from the viewpoint of every agent $i$ with the incorporation of copy variables. The inter-agent state constant can be recast as $\boldsymbol{d}_{ij,k}(\tilde{\boldsymbol{x}}_{i,k}, \tilde{\boldsymbol{x}}^i_{j,k}) \leq 0$ and the inter-agent time constraint also could be rewritten as $\boldsymbol{e}_{ij, N}(\tilde{t}_{i, N}, \tilde{t}^i_{j, N}) \leq 0$, or represented using a compact form with}
%{\colb With the introduction of copy variables, constraints \eqref{eq:intercons1} and \eqref{eq:intercons2} can be revisited from the perspective of each agent $i$, which can be rewritten as $\boldsymbol{d}_{ij,k}(\tilde{\boldsymbol{x}}_{i,k}, \tilde{\boldsymbol{x}}^i_{j,k}) \leq 0$ and $\boldsymbol{e}_{ij, N}(\tilde{t}_{i, N}, \tilde{t}^i_{j, N}) \leq 0$, or denoted as}
\begin{equation}
	\begin{aligned}
		\boldsymbol{d}_{i, k}^{\mathrm{a}}\left(\tilde{\boldsymbol{x}}_{i, k}^{\mathrm{a}}\right) &\leq 0\\
		\boldsymbol{e}_{i, N}^{\mathrm{a}}\left(\tilde{\boldsymbol{t}}_{i, N}^{\mathrm{a}}\right) &\leq 0
	\end{aligned}
\end{equation}
in which $\boldsymbol{d}_{i, k}^{\mathrm{a}}\left(\tilde{\boldsymbol{x}}_{i, k}^{\mathrm{a}}\right)=\left[\left\{\boldsymbol{d}_{i j, k}\left(\tilde{\boldsymbol{x}}_{i, k}, \tilde{\boldsymbol{x}}_{j, k}^i\right)\right\}_{j \in \mathcal{N}_i \backslash\{i\}}\right]$, $\boldsymbol{e}_{i, N}^{\mathrm{a}}\left(\tilde{t}_{i, k}^{\mathrm{a}}\right)=\left[\left\{\boldsymbol{e}_{i j, N}\left(\tilde{t}_{i, N}, \tilde{t}_{j, N}^i\right)\right\}_{j \in \mathcal{N}_i \backslash\{i\}}\right]$.

{\colb Furthermore, since multiple neighbors of agent $i$ may hold different perspectives on $i$'s variables, a consensus between these perspectives needs to be achieved to avoid conflict. Thus, the state and time variables for consensus} are defined as $\boldsymbol{z}_{i, k} \in \mathbb{R}^{p}$ and $s_{i,N} \in \mathbb{R}$, where $i\in \mathcal{M}$. Then, the consensus constraints can be enforced as
\begin{equation}
	\tilde{\boldsymbol{x}}_{j,k}^i=\boldsymbol{z}_{j,k}, \quad t_{j,N}^i=s_{j,N}, \quad j \in \mathcal{N}_i
	\label{eq:consconstr}
\end{equation}
and the compact form considering all the neighbors of Eq. \eqref{eq:consconstr} is given by 
\begin{equation}
	\tilde{\boldsymbol{x}}_{i,k}^{\mathrm{a}}=\boldsymbol{z}_{i,k}^{\mathrm{a}}, \quad \tilde{\boldsymbol{t}}_{i,N}^{\mathrm{a}}=\boldsymbol{s}_{i,N}^{\mathrm{a}}
\end{equation}
where $\boldsymbol{z}_{i, k}^{\mathrm{a}}=\left[\left\{\boldsymbol{z}_{j, k}\right\}_{j \in \mathcal{N}_i}\right] \in \mathbb{R}^{\tilde{p}_i}$, $\boldsymbol{s}_{i,N}^{\mathrm{a}}=\left[\left\{\boldsymbol{s}_{j,N}\right\}_{j \in \mathcal{N}_i}\right] \in \mathbb{R}^{M}$. 

In summary, we can transform the original \textit{MAOC}$_1$ problem into the following equivalent optimization problem.\\
\textit{MAOC}$_2$: Find
\begin{equation}
	\begin{aligned}
		\{ \boldsymbol{U}_{i}^*, t_{i,N}^* \}&=\min {\sum_{i=1}^{M}  J_i\left(\boldsymbol{X}_i, \boldsymbol{U}_i\right)}\\
		s.t. \qquad &\boldsymbol{x}_{i,k+1} = \boldsymbol{f}_{i}\left( \boldsymbol{x}_{i,k}, \boldsymbol{u}_{i,k} \right)\\
		&\boldsymbol{a}_{i,k}\left( {\tilde{\boldsymbol{x}}_{i,k}} \right) \le 0 \\
		&\boldsymbol{b}_{i,k}\left(\tilde{\boldsymbol{u}}_{i, k}\right) \leq 0\\
		&c_{i, N}\left(\tilde{t}_{i, N}\right) \leq 0\\
		&\boldsymbol{d}_{i,k}^{\mathrm{a}}\left(\tilde{\boldsymbol{x}}_{i,k}^{\mathrm{a}}\right) \leq 0\\
		&\boldsymbol{e}_{i,N}^{\mathrm{a}}\left(\tilde{\boldsymbol{t}}_{i,N}^{\mathrm{a}}\right) = 0\\
		\boldsymbol{u}_{i,k}=\tilde{\boldsymbol{u}}_{i,k}, \ \boldsymbol{x}_{i,k}=\tilde{\boldsymbol{x}}_{i,k}, &\ t_{i,N}=\tilde{t}_{i,N}, \ \tilde{\boldsymbol{x}}_{i,k}^{\mathrm{a}}=\boldsymbol{z}_{i,k}^{\mathrm{a}},\ \tilde{\boldsymbol{t}}_{i,N}^{\mathrm{a}}=\boldsymbol{s}_{i,N}^{\mathrm{a}}
	\end{aligned}
	\label{eq: MAOC2}
\end{equation}

\begin{figure*}[htb]
	\centering
	\includegraphics[width=0.95\textwidth]{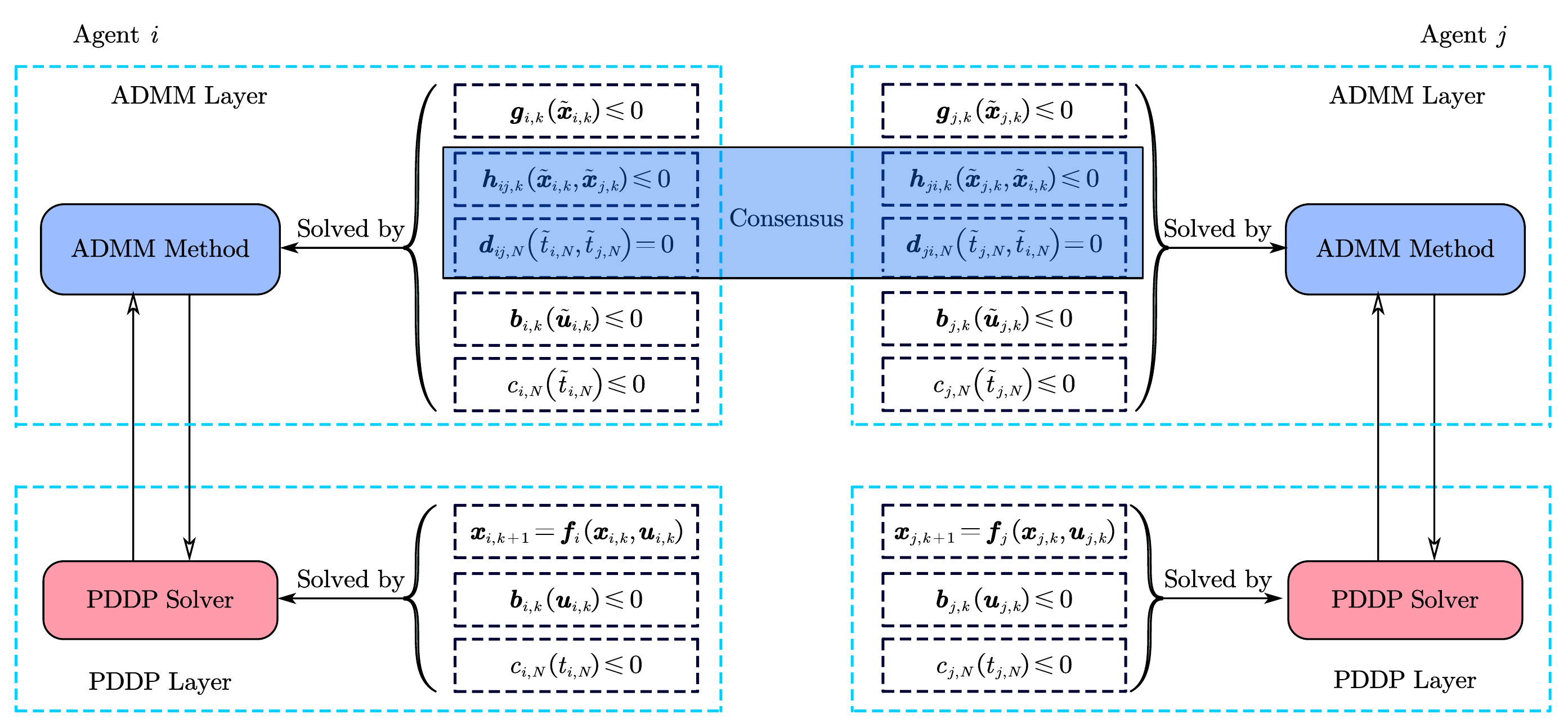}
	\caption{\label{fig:twolayer}
	Schematic diagram of a two-layer structure for distributed spatial-temporal optimization.}
\end{figure*}

The algorithm proposed in this subsection to solve \textit{MAOC}$_2$ has a two-layer structure, where `inter-agent' spatial-temporal constraints are handled by ADMM, while flexible-final-time trajectory optimization with dynamics and control limits for a single agent is solved using {\col PDDP}. The general idea of the proposed algorithm can be expressed in the Figure \ref{fig:twolayer}.

\subsection{Problem Solution}
To Problem \eqref{eq: MAOC2} using ADMM, {\colb we first construct the standard ADMM form by fusing the constraints into the objective function using the corresponding indicator function}
%we first replace the constraints with their indicator functions for constructing the standard ADMM form as
\begin{equation}
	\begin{aligned}
		\min & \sum_{i=1}^{M} \sum_{k=0}^{N} J_i\left(\boldsymbol{x}_{i,k}, \boldsymbol{u}_{i,k}\right)+\mathcal{I}_{\boldsymbol{f}_i}\left(\boldsymbol{x}_{i,k}, \boldsymbol{u}_{i,k}\right)+\mathcal{I}_{\boldsymbol{b}_{i,k}}\left(\tilde{\boldsymbol{u}}_{i,k}\right)+\mathcal{I}_{c_i}\left(\tilde{t}_{i,N}\right)\\
		& +\mathcal{I}_{\boldsymbol{a}_{i,k}}\left(\tilde{\boldsymbol{x}}_{i,k},\tilde{\boldsymbol{u}}_{i,k}\right)+\mathcal{I}_{\boldsymbol{d}_{i,k}^{\mathrm{a}}}\left(\tilde{\boldsymbol{x}}_{i,k}^{\mathrm{a}}\right)+\mathcal{I}_{\boldsymbol{e}_{i,N}^{\mathrm{a}}}\left(\tilde{\boldsymbol{t}}_{i,N}^{\mathrm{a}}\right) \\
		\text { s.t. } & \boldsymbol{u}_{i,k}=\tilde{\boldsymbol{u}}_{i,k},\ \boldsymbol{x}_{i,k}=\tilde{\boldsymbol{x}}_{i,k},\ t_{i,N}=\tilde{t}_{i,N},\ \tilde{\boldsymbol{x}}_{i,k}^{\mathrm{a}}=\boldsymbol{z}_{i,k}^{\mathrm{a}},\ \tilde{\boldsymbol{t}}_{i,N}^{\mathrm{a}}=\boldsymbol{s}_{i,N}^{\mathrm{a}}
	\end{aligned}
	\label{eq:prob2}
\end{equation}
where $J_i\left(\boldsymbol{x}_{i,k}, \boldsymbol{u}_{i,k}\right) = l_{i}(\boldsymbol{x}_{i,k}, \boldsymbol{u}_{i,k}) $, if $k <N$ and $J_i\left(\boldsymbol{x}_{i,k}, \boldsymbol{u}_{i,k}\right) =\phi_{i}(\boldsymbol{x}_{i,N}, t_{i,N})$ if $k=N$. The indicator function $\mathcal{I}_{\mathcal{C}} (\boldsymbol{x})$ is defined to denote whether the vector $\boldsymbol{x}$ is on the set $\mathcal{C}$. If the vector $\boldsymbol{x}$ is on the set $\mathcal{C}$, then $\mathcal{I}_{\mathcal{C}} (\boldsymbol{x})=0$; conversely $\boldsymbol{x}$ is not on $\mathcal{C}$, then $\mathcal{I}_{\mathcal{C}} (\boldsymbol{x})=+\infty$.

As the derivation process of standard ADMM, the augmented Lagrange multipliers formation of problem \eqref{eq:prob2} {\colb is given by}
\begin{equation}
	\begin{aligned}
		\mathcal{L}= & \sum_{i=1}^M \sum_{k=0}^{N} J_i\left(\boldsymbol{x}_{i,k}, \boldsymbol{u}_{i,k}\right)+\mathcal{I}_{\boldsymbol{f}_i}\left(\boldsymbol{x}_{i,k}, \boldsymbol{u}_{i,k}\right)+\mathcal{I}_{\boldsymbol{b}_{i,k}}\left(\tilde{\boldsymbol{u}}_{i,k}\right)+\mathcal{I}_{c_i}\left(\tilde{t}_{i,N}\right)\\
		& +\mathcal{I}_{\boldsymbol{a}_{i,k}}\left(\tilde{\boldsymbol{x}}_{i,k}\right) +\mathcal{I}_{\boldsymbol{d}_{i,k}^{\mathrm{a}}}\left(\tilde{\boldsymbol{x}}_{i,k}^{\mathrm{a}}\right) +\mathcal{I}_{\boldsymbol{e}_{i,k}^{\mathrm{a}}}\left(\tilde{\boldsymbol{t}}_{i,N}^{\mathrm{a}}\right)\\
		& +\boldsymbol{\zeta}_{i,k}^{\mathrm{T}}\left(\boldsymbol{u}_{i,k}-\tilde{\boldsymbol{u}}_{i,k}\right) +\boldsymbol{\lambda}_{i,k}^{\mathrm{T}}\left(\boldsymbol{x}_{i,k}-\tilde{\boldsymbol{x}}_{i,k}\right) +{\col \nu_{i,N}}^{\mathrm{T}}\left(t_{i,N}-\tilde{t}_{i,N}\right)\\
		& +\boldsymbol{y}_{i,k}^{\mathrm{T}}\left(\tilde{\boldsymbol{x}}_{i,k}^{\mathrm{a}}-\boldsymbol{z}_{i,k}^{\mathrm{a}}\right) +\boldsymbol{\eta}_{i,N}^{\mathrm{T}}\left(\tilde{\boldsymbol{t}}_{i,N}^{\mathrm{a}}-\boldsymbol{s}_{i,N}^{\mathrm{a}}\right) \\
		& +\frac{\tau}{2}\left\|\boldsymbol{u}_{i,k}-\tilde{\boldsymbol{u}}_{i,k}\right\|_2^2 +\frac{\rho}{2}\left\|\boldsymbol{x}_{i,k}-\tilde{\boldsymbol{x}}_{i,k}\right\|_2^2 +\frac{\sigma}{2}\left\|t_{i,N}-\tilde{t}_{i,N}\right\|_2^2\\
		& +\frac{\mu}{2}\left\|\tilde{\boldsymbol{x}}_{i,k}^{\mathrm{a}}-\boldsymbol{z}_{i,k}^{\mathrm{a}}\right\|_2^2 +\frac{\gamma}{2}\left\|\tilde{\boldsymbol{t}}_{i,k}^{\mathrm{a}}-\boldsymbol{s}_{i,N}^{\mathrm{a}}\right\|_2^2
		\end{aligned}
\end{equation}
where {\colb the dual variables of constraints are represented by $\boldsymbol{\zeta}_{i,k}$, $\boldsymbol{\lambda}_{i,k}$, ${\col \nu_{i,N}}$, $\boldsymbol{y}_{i,k}$, $\boldsymbol{\eta}_{i,N}$, and the penalty parameters of constraints are denoted using $\tau,\rho,\sigma,\mu,\gamma>0$. The D-PDDP is divided into three optimization(computation) steps, which are derived as follows.}
%$\boldsymbol{\zeta}_{i,k}$, $\boldsymbol{\lambda}_{i,k}$, ${\col \nu_{i,N}}$, $\boldsymbol{y}_{i,k}$, $\boldsymbol{\eta}_{i,N}$ are the corresponding dual variables of the constraints and $\tau,\rho,\sigma,\mu,\gamma>0$ are the penalty parameters. Next, the update rule of D-PDDP is derived step by step, in which the number of iterations is denoted with $n$.

{\it Optimization Step 1. Single-Agent PDDP Update}: In this update step, PDDP algorithm is employed to minimize the augmented Lagrangian with respect to the local variables $\boldsymbol{x}_{i,k}$, $\boldsymbol{u}_{i,k}$, $t_{i,N}$ as a spatial-temporal trajectory optimizer for a single agent, i.e.,
\begin{equation}
	\left\{\boldsymbol{x}_{i,k}, \boldsymbol{u}_{i,k}, t_{i,N}\right\}^{n+1}=\arg \min \mathcal{L}\left(\boldsymbol{x}_{i,k}, \boldsymbol{u}_{i,k}, t_{i,N}, \tilde{\boldsymbol{x}}_{i,k}^{\mathrm{a}, n}, \tilde{\boldsymbol{u}}_{i,k}^n, \tilde{\boldsymbol{t}}_{i,N}^{\mathrm{a}, n}, \boldsymbol{z}_{i,k}^{\mathrm{a},n}, s_{i,N}^{\mathrm{a},n}, \boldsymbol{\zeta}_{i,k}^n, \boldsymbol{\lambda}_{i,k}^n, \nu_{i,N}^n, \boldsymbol{y}_{i,k}^n, \boldsymbol{\eta}_{i,N}^n\right)
	\label{eq:stp1prob}
\end{equation}
{\colb where the number of iterations is denoted by $n$. This optimization step can be decomposed into subproblems that can be computed in a parallel way for each agent $i \in \mathcal{M}$ in the following form}
%, and this optimization step is computed for each agent $i \in \mathcal{M}$ at each time instant $k$. This results in the following $M$ subproblems that can be solved in parallel for all agents
\begin{equation}
	\begin{gathered}
		\left\{\boldsymbol{x}_{i,k}, \boldsymbol{u}_{i,k}; t_{i,N}\right\}^{n+1}=\arg \min \sum_{k=1}^{N-1}\left[\hat{l}_i\left(\boldsymbol{x}_{i, k}, \boldsymbol{u}_{i, k}; t_{i,N}\right)\right]+\hat{\phi}_i\left(\boldsymbol{x}_{i, N}; t_{i,N}\right) \\
		\text { s.t. } \boldsymbol{x}_{i, k+1}=\boldsymbol{f}_i\left(\boldsymbol{x}_{i, k}, \boldsymbol{u}_{i, k}; t_{i,N}\right)
	\end{gathered}
	\label{eq:step1}
\end{equation}
{\colb where the running cost $\hat{l}_i\left(\boldsymbol{x}_{i, k}, \boldsymbol{u}_{i, k}; t_{i,N}\right)$ and the terminal cost $\hat{\phi}_i\left(\boldsymbol{x}_{i, N}; t_{i,N}\right)$ are revised as}
\begin{equation}
	\begin{gathered}
		\hat{l}_{i}\left(\boldsymbol{x}_{i, k}, \boldsymbol{u}_{i, k}; t_{i,N}\right)=l_{i}\left(\boldsymbol{x}_{i, k}, \boldsymbol{u}_{i, k}; t_{i,N}\right)+\frac{\rho}{2}\left\|\boldsymbol{x}_{i, k}-\tilde{\boldsymbol{x}}_{i, k}+\frac{\boldsymbol{\lambda}_{i, k}}{\rho}\right\|_2^2 +\frac{\tau}{2}\left\|\boldsymbol{u}_{i, k}-\tilde{\boldsymbol{u}}_{i, k}+\frac{\boldsymbol{\zeta}_{i, k}}{\tau}\right\|_2^2 \\
		\hat{\phi}_i\left(\boldsymbol{x}_{i,N}; t_{i,N}\right) = \phi_i\left(\boldsymbol{x}_{i,N};t_{i, N}\right)  +\frac{\rho}{2}\left\|\boldsymbol{x}_{i,N}-\tilde{\boldsymbol{x}}_{i,N}+\frac{\boldsymbol{\lambda}_{i,N}}{\rho}\right\|_2^2
		+\frac{\sigma}{2}\left\|t_{i,N}-\tilde{t}_{i,N}+\frac{\nu_{i,N}}{\sigma}\right\|_2^2
		\end{gathered}
\end{equation}
{\colb Every agent $i\in \mathcal{M}$ computes a single subproblem above in parallel using the PDDP method. The derivatives of the action value function ${Q}\left( \boldsymbol{x}_{i,k},\boldsymbol{u}_{i,k}; \boldsymbol{\theta}_i\right)$ are revised as}
%Each of these subproblems is solved in parallel by every agent $i\in \mathcal{M}$ with PDDP. The backward pass of the PDDP at the terminal time instant ${t}_{i,N}$ is revised as the following form
\begin{equation}
\begin{split}
& Q_{\boldsymbol{x}_i, k}= l_{\boldsymbol{x}_i, k} + \boldsymbol{f}_{\boldsymbol{x}_i, k}^{\mathrm{T}} V_{\boldsymbol{x}_i,k+1} +\rho\left(\boldsymbol{x}_{i, k}-\tilde{\boldsymbol{x}}_{i, k}\right)+\boldsymbol{\lambda}_{i, k}\\ 
& Q_{\boldsymbol{u}_i, k}= l_{\boldsymbol{u}_i, k} + \boldsymbol{f}_{\boldsymbol{u}_i, k}^{\mathrm{T}} V_{\boldsymbol{x}_i,k+1} +\tau\left(\boldsymbol{u}_{i, k}-\tilde{\boldsymbol{u}}_{i, k}\right)+\boldsymbol{\zeta}_{i, k}\\ 
& Q_{\boldsymbol{\theta}_i, k}= l_{\boldsymbol{\theta}_i, k} + V_{\boldsymbol{\theta}_i, k+1} + \boldsymbol{f}_{\boldsymbol{\theta}_i, k}^{\mathrm{T}} V_{\boldsymbol{x}_i,k+1}\\ 
& Q_{\boldsymbol{x}_i \boldsymbol{x}_i, k} = l_{\boldsymbol{x}_i \boldsymbol{x}_i, k} + \boldsymbol{f}_{\boldsymbol{x}_i, k}^{\mathrm{T}} V_{\boldsymbol{x}_i \boldsymbol{x}_i,k+1} \boldsymbol{f}_{\boldsymbol{x}_i, k} +\rho \boldsymbol{I}\\
%+V_{\boldsymbol{x}}\left(\boldsymbol{x}_{k+1}\right) \boldsymbol{f}_{\boldsymbol{x} \boldsymbol{x}} \left(\boldsymbol{x}_{k}, \boldsymbol{u}_{k}\right)\\ 
& Q_{\boldsymbol{u}_i \boldsymbol{u}_i, k} =l_{\boldsymbol{u}_i \boldsymbol{u}_i, k} +\boldsymbol{f}_{\boldsymbol{u}_i, k}^{\mathrm{T}} V_{\boldsymbol{x}_i \boldsymbol{x}_i,k+1} \boldsymbol{f}_{\boldsymbol{u}_i, k} +\tau \boldsymbol{I}\\
%+V_{\boldsymbol{x}}\left(\boldsymbol{x}_{k+1}\right)\boldsymbol{f}_{\boldsymbol{u} \boldsymbol{u}}\left(\boldsymbol{x}_{k}, \boldsymbol{u}_{k}\right)\\
& Q_{\boldsymbol{\theta}_i \boldsymbol{\theta}_i, k} =l_{\boldsymbol{\theta}_i \boldsymbol{\theta}_i, k}  + V_{\boldsymbol{\theta}_i \boldsymbol{\theta}_i, k+1} +\boldsymbol{f}_{\boldsymbol{\theta}_i, k}^{\mathrm{T}} V_{\boldsymbol{x}_i \boldsymbol{x}_i,k+1} \boldsymbol{f}_{\boldsymbol{\theta}_i, k} +\boldsymbol{f}_{\boldsymbol{\theta}_i, k}^{\mathrm{T}} V_{\boldsymbol{x}_i \boldsymbol{\theta}_i,k+1} + V_{\boldsymbol{\theta}_i \boldsymbol{x}_i,k+1} \boldsymbol{f}_{\boldsymbol{\theta}_i, k}\\
&Q_{\boldsymbol{x}_i \boldsymbol{u}_i, k} = l_{\boldsymbol{x}_i \boldsymbol{u}_i, k} + \boldsymbol{f}_{\boldsymbol{x}_i, k}^{\rm{T}} V_{\boldsymbol{x}_i \boldsymbol{x}_i,k+1} \boldsymbol{f}_{\boldsymbol{u}_i, k} = Q_{\boldsymbol{u}_i \boldsymbol{x}_i,k}^{\mathrm{T}}\\
%+V_{\boldsymbol{x}}\left(\boldsymbol{x}_{k}, \boldsymbol{u}_{k}\right)\boldsymbol{f}_{\boldsymbol{u} \boldsymbol{x}}\left(\boldsymbol{x}_{k}, \boldsymbol{u}_{k}\right) \\ 
&Q_{\boldsymbol{x}_i \boldsymbol{\theta}_i, k} = l_{\boldsymbol{x}_i \boldsymbol{\theta}_i, k} +\boldsymbol{f}_{\boldsymbol{x}_i, k}^{\rm{T}} V_{\boldsymbol{x}_i \boldsymbol{x}_i,k+1} \boldsymbol{f}_{\boldsymbol{\theta}_i, k} +\boldsymbol{f}_{\boldsymbol{x}_i, k}^{\rm{T}} V_{\boldsymbol{x}_i \boldsymbol{\theta}_i,k+1} = Q_{\boldsymbol{\theta}_i \boldsymbol{x}_i,k}^{\mathrm{T}}\\
&Q_{\boldsymbol{u}_i \boldsymbol{\theta}_i, k} = l_{\boldsymbol{u}_i \boldsymbol{\theta}_i, k} + \boldsymbol{f}_{\boldsymbol{u}_i, k}^{\rm{T}} V_{\boldsymbol{x}_i \boldsymbol{x}_i,k+1} \boldsymbol{f}_{\boldsymbol{\theta}_i, k} +\boldsymbol{f}_{\boldsymbol{u}_i, k}^{\rm{T}} V_{\boldsymbol{u}_i \boldsymbol{\theta}_i,k+1} = Q_{\boldsymbol{\theta}_i \boldsymbol{u}_i,k}^{\mathrm{T}}\\
\end{split}
\label{eq:quus}
\end{equation}

%The terminal conditions for the value functions $V(\boldsymbol{x}_{i,N}, t_{i,N})$ and its derivatives will be
{\colb For the value functions $V(\boldsymbol{x}_{i,N}, t_{i,N})$ and its derivatives, the terminal conditions will be given by}
\begin{equation}
	\begin{aligned}
		V\left(\boldsymbol{x}_{i,N};t_{i,N}\right) &=\phi_i\left(\boldsymbol{x}_{i,N};t_{i, N}\right)  +\frac{\rho}{2}\left\|\boldsymbol{x}_{i,N}-\tilde{\boldsymbol{x}}_{i,N}+\frac{\boldsymbol{\lambda}_{i,N}}{\rho}\right\|_2^2 +\frac{\sigma}{2}\left\|t_{i,N}-\tilde{t}_{i,N}+\frac{\nu_{i,N}}{\sigma}\right\|_2^2\\
		V_{\boldsymbol{x}_i,N} &= {\phi}_{\boldsymbol{x}_i,N} + \rho\left(\boldsymbol{x}_{i, N-1}-\tilde{\boldsymbol{x}}_{i, N-1}\right)+\boldsymbol{\lambda}_{i, N-1}\\
		V_{\boldsymbol{\theta}_i,N} &={\phi}_{\boldsymbol{\theta}_i,N} + \sigma\left(t_{i, N-1}-\tilde{t}_{i, N-1}\right)+\nu_{i, N-1}\\
		V_{\boldsymbol{x}_i \boldsymbol{x}_i,N} &=\phi_{\boldsymbol{x}_i \boldsymbol{x}_i, N} +\rho \boldsymbol{I}_{p_i \times p_i}\\
		V_{\boldsymbol{x}_i \boldsymbol{\theta}_i,N} &=\phi_{\boldsymbol{x}_i \boldsymbol{\theta}_i, N}
		= V_{\boldsymbol{\theta}_i \boldsymbol{x}_i,N}^{\mathrm{T}}\\
		V_{\boldsymbol{\theta}_i \boldsymbol{\theta}_i,N} &=\phi_{\boldsymbol{\theta}_i \boldsymbol{\theta}_i, N} +\sigma \boldsymbol{I}_{1 \times 1}
	\end{aligned}
	\label{eq:vtts}
\end{equation}

%With a poor initial guess of the control sequence and the final time, PDDP might diverge and cannot converge to the local optimum. 
With the above derivation, the dynamic constraint represented in Fig.~\ref{fig:twolayer} is handled. Moreover, the PDDP solver also needs to deal with the constraints $\boldsymbol{b}_{i,k}\left({\boldsymbol{u}}_{i, k}\right) \leq 0$ shown in Fig.~\ref{fig:twolayer} to prevent the algorithm from diverging due to poor initial guesses. {\colb A new control effort is limited within a bounded region using the element-wise clamping, or projection operator, to guarantee numerical stability as}
%To ensure numerical stability, the update of the control update is limited within a bounded region using the element-wise clamping, or projection operator, as
\begin{equation}
	 {\boldsymbol{u}}_{i,k}=\min(\max(\boldsymbol{u}_{i,k} + \delta \boldsymbol{u}_{i,k}^{*}, \boldsymbol{u}_{\min}), \boldsymbol{u}_{\max})
\end{equation}
{\colb in which the lower and higher boundaries of the control effort are denoted using $\boldsymbol{u}_{\min}$ and $\boldsymbol{u}_{\max}$, respectively.}
%where $\boldsymbol{u}_{\min}$ and $\boldsymbol{u}_{\max}$ represent the lower and the upper bounds of the control effort, respectively. 

Likewise, the final time update is also limited to ensure the constraint $c_{i, N}\left(\tilde{t}_{i, N}\right) \leq 0$ in Fig.~\ref{fig:twolayer} as
\begin{equation}
	 {t}_{i, N}=\min(\max( {t}_{i, N} + \delta t_{i, N}^{*}, t_{\min}), t_{\max})
\end{equation}
{\colb in which the final time's boundaries are denoted using $t_{\min}$ and $t_{\max}$.}
%where $t_{\min}$ and $t_{\max}$ represent {\colb the boundaries of the final time.}
%the lower and the upper bounds of the final time, respectively.

\begin{remark}
{\colb The new local problems \eqref{eq:stp1prob} include a significant modification compared to the unconstrained single-agent problem, i.e., the new objective function added terms that enforce consensus to be reached between local and the copy variables, which will constrain the local constraints of agents to reach consensus between PDDP and ADMM.}
%the new costs include extra terms that encourage a consensus between local and copy variables, which lead to a consensus of the local constraints between PDDP and ADMM.
\end{remark}

{\it Optimization Step 2. Single-Agent Safe Update}: In this update step, ADMM method is utilized to handle the rest local constraints as the following update rule
\begin{equation}
	\left\{\tilde{\boldsymbol{x}}_{i,k}^{\mathrm{a}}, \tilde{\boldsymbol{u}}_{i,k}, \tilde{\boldsymbol{t}}_{i,N}^{\mathrm{a}}\right\}^{n+1}=\arg \min \mathcal{L}\left(\boldsymbol{x}_{i,k}^{n+1}, \boldsymbol{u}_{i,k}^{n+1}, t_{i,N}^{n+1}, \tilde{\boldsymbol{x}}_{i,k}^{\mathrm{a}}, \tilde{\boldsymbol{u}}_{i,k}, \tilde{\boldsymbol{t}}_{i,N}^{\mathrm{a}}, \boldsymbol{z}_{i,k}^{\mathrm{a},n}, s_{i,N}^{\mathrm{a},n}, \boldsymbol{\zeta}_{i,k}^n, \boldsymbol{\lambda}_{i,k}^n, {\col \nu_{i,N}^n}, \boldsymbol{y}_{i,k}^n, \boldsymbol{\eta}_{i,N}^n\right)
	\label{eq:step2prb}
\end{equation}

Since the state and control variables of each agent are decoupled, problem \eqref{eq:step2prb} can be decomposed to the following $M\times (N+1)$ state subproblems for every agent $i$ at each time instant $k$, which can be solved in parallel as
\begin{equation}
	\begin{aligned}
		\tilde{\boldsymbol{x}}_{i,k}^{\mathrm{a}, n+1}&=\arg \min \frac{\rho}{2}\left\|\boldsymbol{x}_{i,k}-\tilde{\boldsymbol{x}}_{i,k}+\frac{\boldsymbol{\lambda}_{i,k}}{\rho}\right\|_2^2+\frac{\mu}{2}\left\|\tilde{\boldsymbol{x}}_{i,k}^{\mathrm{a}}-\boldsymbol{z}_{i,k}^{\mathrm{a}}+\frac{\boldsymbol{y}_{i,k}}{\mu}\right\|_2^2 \\
		&\text { s.t. } \boldsymbol{a}_{i, k}\left(\tilde{\boldsymbol{x}}_{i, k}\right) \leq 0,\ \boldsymbol{d}_{i, k}^{\mathrm{a}}\left(\tilde{\boldsymbol{x}}_{i, k}^{\mathrm{a}}\right) \leq 0
	\end{aligned}
	\label{eq:stsub}
\end{equation}
%and $M\times N$ control subproblems
{\colb Similarly, the control effort can be decomposed into $M\times N$ smaller problems for solving}
\begin{equation}
	\begin{aligned}
		& \tilde{\boldsymbol{u}}_{i,k}^{n+1}=\arg \min \frac{\tau}{2}\left\|\boldsymbol{u}_{i,k}-\tilde{\boldsymbol{u}}_{i,k}+\frac{\boldsymbol{\zeta}_{i,k}}{\tau}\right\|_2^2 \\
		& \text { s.t. } \quad \boldsymbol{b}_{i, k}\left(\tilde{\boldsymbol{u}}_{i, k}\right) \leq 0 
	\end{aligned}
	\label{eq:ctrsub}
\end{equation}
and $M$ terminal time optimization subproblems
\begin{equation}
	\begin{aligned}
		\tilde{\boldsymbol{t}}_{i,N}^{\mathrm{a}, n+1}=\arg \min &\frac{\sigma}{2}\left\|t_{i,N}-\tilde{t}_{i,N}+\frac{\nu_{i,N}}{\sigma}\right\|_2^2+\frac{\gamma}{2}\left\|\tilde{\boldsymbol{t}}_{i,N}^{\mathrm{a}}-\boldsymbol{s}_{i,N}^{\mathrm{a}}+\frac{\boldsymbol{\eta}_{i,N}}{\gamma}\right\|_2^2 \\
		&\text { s.t. } c_{i, N}\left(\tilde{t}_{i, N}\right) \leq 0,\ \boldsymbol{e}_{i, N}^{\mathrm{a}}\left(\tilde{\boldsymbol{t}}_{i, N}^{\mathrm{a}}\right) = 0
	\end{aligned}
	\label{eq:tsub}
\end{equation}
{\col where the control input constraints $\boldsymbol{b}_{i, k}\left(\tilde{\boldsymbol{u}}_{i, k}\right) \leq 0$ can be handled in a similar way as step 1 and the terminal time boundary constraints $c_{i, N}\left(\tilde{t}_{i, N}\right) \leq 0$ will be solved with the constraint $\boldsymbol{e}_{i, N}^{\mathrm{a}}\left(\tilde{\boldsymbol{t}}_{i, N}^{\mathrm{a}}\right) = 0$ by the standard solver.}
%What can be found from the subproblems above is that each agent's state and control sub-problem can be further decomposed into $N+1$ smaller sub-problems (i.e., for each time instant $k = 0, \ldots, N$), i.e:
%\begin{equation}
%	\begin{aligned}
%		\tilde{\boldsymbol{x}}_{i, k}^{\mathrm{a}, n+1}= & \arg \min \frac{\rho}{2}\left\|\boldsymbol{x}_{i, k}-\tilde{\boldsymbol{x}}_{i, k}+\frac{\lambda_{i, k}}{\rho}\right\|_2^2 \\
%		& +\frac{\mu}{2}\left\|\tilde{\boldsymbol{x}}_{i, k}^{\mathrm{a}}-\boldsymbol{z}_{i, k}^{\mathrm{a}}+\frac{\boldsymbol{y}_{i, k}}{\mu}\right\|_2^2 \\
%		\text { s.t. } & \quad \boldsymbol{a}_{i, k}\left(\tilde{\boldsymbol{x}}_{i, k}\right) \leq 0, \boldsymbol{d}_{i, k}^{\mathrm{a}}\left(\tilde{\boldsymbol{x}}_{i, k}^{\mathrm{a}}\right) \leq 0
%	\end{aligned}
%	\label{eq:stssub}
%\end{equation}
%and
%\begin{equation}
%	\begin{aligned}
%		\tilde{\boldsymbol{u}}_{i, k}^{n+1}=&\arg \min \frac{\tau}{2}\left\|\boldsymbol{u}_{i, k}-\tilde{\boldsymbol{u}}_{i, k}+\frac{\boldsymbol{\zeta}_{i, k}}{\tau}\right\|_2^2 \\
%		&\text { s.t. } \boldsymbol{b}_{i, k}\left(\tilde{\boldsymbol{u}}_{i, k}\right) \leq 0
%	\end{aligned}
%	\label{eq:ctrssub}
%\end{equation}
%which can be solved in parallel.
\begin{remark}
Optimization Step enables to decompose \eqref{eq:step2prb} into $M (2N+2)$ low-dimensional subproblems and hence the computation can performed in parallel for each agent as well as each time instant, which can greatly improve the computational efficiency of the algorithm. With reasonable convexification, the above subproblems \eqref{eq:stsub}, \eqref{eq:ctrsub}, \eqref{eq:tsub} can be transformed into Quadratic Programming (QP) problems, and thus solved by well-established QP optimization toolbox.
\end{remark}

{\it Optimization Step 3. Global Update}: After the local optimization step, the global variables $\boldsymbol{z}_{i,k}$, $\boldsymbol{s}_{i,N}$ are updated, i.e.,
\begin{equation}
	\left\{\boldsymbol{z}_{i,k}^{\mathrm{a}}, \boldsymbol{s}_{i,N}^{\mathrm{a}}\right\}^{n+1}=\arg \min \mathcal{L}\left(\boldsymbol{x}_{i,k}^{n+1}, \boldsymbol{u}_{i,k}^{n+1}, t_{i,N}^{n+1}, \tilde{\boldsymbol{x}}_{i,k}^{\mathrm{a},n+1}, \tilde{\boldsymbol{u}}_{i,k}^{n+1}, \tilde{\boldsymbol{t}}_{i,N}^{\mathrm{a},n+1}, \boldsymbol{z}_{i,k}^{\mathrm{a}}, \boldsymbol{s}_{i,N}^{\mathrm{a}}, \boldsymbol{\zeta}_{i,k}^n, \boldsymbol{\lambda}_{i,k}^n, {\col \nu_{i,N}^n}, \boldsymbol{y}_{i,k}^n, \boldsymbol{\eta}_{i,N}^n\right)
	\label{eq:step3prb}
\end{equation}

According to constraint \eqref{eq:consconstr}, we have
\begin{equation}
	\left\{\boldsymbol{z}_{i,k}, \boldsymbol{s}_{i,N}\right\}^{n+1}=\arg \min \sum_{j \in \mathcal{P}_i} \frac{\mu}{2}\left\|\tilde{\boldsymbol{x}}_{i,k}^j-\boldsymbol{z}_{i,k}+\frac{\boldsymbol{y}_{i,k}}{\mu}\right\|_2^2 +\frac{\gamma}{2}\left\|\tilde{\boldsymbol{t}}_{i,N}^j-\boldsymbol{s}_{i,N}+\frac{\boldsymbol{\eta}_{i,N}}{\gamma}\right\|_2^2
\end{equation}
which means that the global consensus variable at agent $i$ is the combined evaluation of the state {\colb from the perspective of $i$'s neighbor $j\in \mathcal{N}_i$, and thus the update rules of the global variables is given by}
%of agent $i$ by its neighbors $j \in \mathcal{N}_i$
\begin{subequations}
	\begin{align}
		\boldsymbol{z}_{i,k}^{n+1}&=\frac{1}{\left|\mathcal{P}_i\right|} \sum_{j \in \mathcal{P}_i} \left( \tilde{\boldsymbol{x}}_{i,k}^{j, n+1}+\frac{1}{\mu} \boldsymbol{y}_{i,k}^{j, n} \right)\\
		\boldsymbol{s}_{i,N}^{n+1}&=\frac{1}{\left|\mathcal{P}_i\right|} \sum_{j \in \mathcal{P}_i} \left( \tilde{t}_{i,N}^{j, n+1}+\frac{1}{\gamma} \boldsymbol{\eta}_{i,N}^{j, n} \right)
	\end{align}
	\label{eq:glbces}
\end{subequations}
where $\left\{\boldsymbol{z}_{j, k}\right\}_{j \in \mathcal{N}_i\backslash \{i\}}$ of $\boldsymbol{z}_{i,k}^{\mathrm{a}}$ is obtained from neighbors of agent $i$ through communication. Since the dual variables $\boldsymbol{y}_{i,k}$ and $\boldsymbol{\eta}_{i,N}$ will gradually become $0$ as the iterations $n$ increase, it can be seen from Eq. \eqref{eq:glbces} that the global update is gradually a local average of all copy variables of agent $i$. Finally, the dual updates are given as
\begin{subequations}
	\begin{align}
		\boldsymbol{\zeta}_{i,k}^{n+1}&=\boldsymbol{\zeta}_{i,k}^n+\tau\left(\boldsymbol{u}_{i,k}^{n+1}-\tilde{\boldsymbol{u}}_{i,k}^{n+1}\right)\\
		\boldsymbol{\lambda}_{i,k}^{n+1}&=\boldsymbol{\lambda}_{i,k}^n+\rho\left(\boldsymbol{x}_{i,k}^{n+1}-\tilde{\boldsymbol{x}}_{i,k}^{n+1}\right)\label{eq:dualupd:b}\\
		\boldsymbol{y}_{i,k}^{n+1}&=\boldsymbol{y}_{i,k}^n+\mu\left(\tilde{\boldsymbol{x}}_{i,k}^{\mathrm{a},n+1}-\boldsymbol{z}_{i,k}^{\mathrm{a},n+1}\right)\\
		\nu_{i,N}^{n+1}&=\nu_{i,N}^n+\sigma\left(t_{i,N}^{n+1}-\tilde{t}_{i,N}^{n+1}\right)\\
		\boldsymbol{\eta}_{i,N}^{n+1}&=\boldsymbol{\eta}_{i,N}^n+\gamma\left(\tilde{\boldsymbol{t}}_{i,N}^{\mathrm{a},n+1}-\boldsymbol{s}_{i,N}^{\mathrm{a},n+1}\right)
	\end{align}
	\label{eq:dualupd}
\end{subequations}

The stopping criterion for the update of the D-PDDP algorithm can be determined by checking whether the residuals of the primal and dual variables of the algorithm, named primal-residual and dual-residual, are below a certain threshold \cite{MAL-016}. The primal and dual residuals of the D-PDDP algorithm at the $n$-th iteration are defined as:
\begin{equation}
	\begin{aligned}
		\boldsymbol{r}_{\boldsymbol{U}}^{\mathrm{p},n} 			= \boldsymbol{U}^n - \tilde{\boldsymbol{U}}^n &\quad\quad 
		\boldsymbol{r}_{\boldsymbol{U}}^{\mathrm{d},n} 			= \tau (\tilde{\boldsymbol{U}}^n - \tilde{\boldsymbol{U}}^{n-1})\\
		\boldsymbol{r}_{\boldsymbol{X}}^{\mathrm{p},n} 			= \boldsymbol{X}^n - \tilde{\boldsymbol{X}}^n &\quad\quad 
		\boldsymbol{r}_{\boldsymbol{X}}^{\mathrm{d},n} 			= \rho (\tilde{\boldsymbol{X}}^n - \tilde{\boldsymbol{X}}^{n-1})\\
		\boldsymbol{r}_{\tilde{\boldsymbol{X}}}^{\mathrm{p},n} 	= \tilde{\boldsymbol{X}}^{\mathrm{a},n} - {\boldsymbol{Z}}^{\mathrm{a},n} &\quad\quad 
		\boldsymbol{r}_{\tilde{\boldsymbol{X}}}^{\mathrm{d},n} 	= \mu ({\boldsymbol{Z}}^{\mathrm{a},n} - {\boldsymbol{Z}}^{\mathrm{a},n-1})\\
		\boldsymbol{r}_{\boldsymbol{T}}^{\mathrm{p},n} 						= \boldsymbol{T}^n - \tilde{\boldsymbol{T}}^n &\quad\quad 
		\boldsymbol{r}_{\boldsymbol{T}}^{\mathrm{d},n} 						= \sigma (\tilde{\boldsymbol{T}}^n - \tilde{\boldsymbol{T}}^{n-1})\\
		\boldsymbol{r}_{\tilde{\boldsymbol{T}}}^{\mathrm{p},n} 	= \tilde{\boldsymbol{T}}^{\mathrm{a},n} - {\boldsymbol{S}}^{\mathrm{a},n} &\quad\quad 
		\boldsymbol{r}_{\tilde{\boldsymbol{T}}}^{\mathrm{d},n} 	= \gamma ({\boldsymbol{S}}^n - {\boldsymbol{S}}^{n-1})\\
	\end{aligned}
	\label{eq:res}
\end{equation}
where $\boldsymbol{r}_{(\cdot)}^{\mathrm{p},n}$ represents the primal residual and $\boldsymbol{r}_{(\cdot)}^{\mathrm{d},n}$ denotes the dual residual. $\boldsymbol{U} = \left[\boldsymbol{U}_1^{\rm{T}},\ldots,\boldsymbol{U}_i ^{\rm{T}},\ldots,\boldsymbol{U}_M^{\rm{T}} \right]^{\rm{T}}$ is the slack of the control variables from all agents and the entire time history. Similarly, we can define other vectors $\boldsymbol{X}$, $\boldsymbol{Z}$, $\boldsymbol{T}$, $\boldsymbol{S}$. The residuals here are a synthesized evaluation of the stacked vectors of all UAVs at all time instants. The primal and dual residuals can be used to monitor convergence of the optimization algorithm, i.e., the primal residual $\boldsymbol{r}_{(\cdot)}^{\mathrm{p},n}$ tends to zero when the algorithm update satisfies the consensus constraints in Problem \eqref{eq:prob2} and the dual residual $\boldsymbol{r}_{(\cdot)}^{\mathrm{d},n}$ tends to zero when the algorithm is updated close to the target minimum \cite{MAL-016}. Hence, the criterion for determining that the update satisfies the stopping condition is:
\begin{equation}
	\begin{aligned}
		\left\|\boldsymbol{r}_{\boldsymbol{U}}^{\mathrm{p},n}\right\|_2 \leq \sqrt{\mathcal{N}_{\boldsymbol{U}}}\epsilon_{\mathrm{abs}} + \epsilon^{\mathrm{rel}} \max \left\{\left\|\boldsymbol{U}^n\right\|_2, \left\|\tilde{\boldsymbol{U}}^n\right\|_2\right\} &\quad\quad 
		\left\|\boldsymbol{r}_{\boldsymbol{U}}^{\mathrm{d},n}\right\|_2 \leq \sqrt{\mathcal{N}_{\boldsymbol{U}}}\epsilon_{\mathrm{abs}} + \epsilon^{\mathrm{rel}} \left\|\boldsymbol{\Xi}^n\right\|_2 \\
		\left\|\boldsymbol{r}_{\boldsymbol{X}}^{\mathrm{p},n}\right\|_2 \leq \sqrt{\mathcal{N}_{\boldsymbol{X}}}\epsilon_{\mathrm{abs}} + \epsilon^{\mathrm{rel}} \max \left\{\left\|\boldsymbol{X}^n\right\|_2, \left\|\tilde{\boldsymbol{X}}^n\right\|_2\right\} &\quad\quad 
		\left\|\boldsymbol{r}_{\boldsymbol{X}}^{\mathrm{d},n}\right\|_2 \leq \sqrt{\mathcal{N}_{\boldsymbol{X}}}\epsilon_{\mathrm{abs}} + \epsilon^{\mathrm{rel}} \left\|\boldsymbol{\Lambda}^n\right\|_2 \\
		\left\|\boldsymbol{r}_{\boldsymbol{T}}^{\mathrm{p},n}\right\|_2 \leq \sqrt{\mathcal{N}_{\boldsymbol{T}}}\epsilon_{\mathrm{abs}} + \epsilon^{\mathrm{rel}} \max \left\{\left\|\boldsymbol{T}^n\right\|_2, \left\|\tilde{\boldsymbol{T}}^n\right\|_2\right\} &\quad\quad 
		\left\|\boldsymbol{r}_{\boldsymbol{T}}^{\mathrm{d},n}\right\|_2 \leq \sqrt{\mathcal{N}_{\boldsymbol{T}}}\epsilon_{\mathrm{abs}} + \epsilon^{\mathrm{rel}} \left\|\boldsymbol{\Pi}^n\right\|_2 \\
		\left\|\boldsymbol{r}_{\boldsymbol{Z}}^{\mathrm{p},n}\right\|_2 \leq \sqrt{\mathcal{N}_{\tilde{\boldsymbol{X}}^{\mathrm{a}}}}\epsilon_{\mathrm{abs}} + \epsilon^{\mathrm{rel}} \max \left\{\left\|\tilde{\boldsymbol{X}}^{\mathrm{a},n}\right\|_2, \left\|{\boldsymbol{Z}}^{\mathrm{a},n}\right\|_2\right\} &\quad\quad 
		\left\|\boldsymbol{r}_{\boldsymbol{Z}}^{\mathrm{d},n}\right\|_2 \leq \sqrt{\mathcal{N}_{\tilde{\boldsymbol{X}}^{\mathrm{a}}}}\epsilon_{\mathrm{abs}} + \epsilon^{\mathrm{rel}} \left\|\boldsymbol{Y}^n\right\|_2 \\
		\left\|\boldsymbol{r}_{\boldsymbol{S}}^{\mathrm{p},n}\right\|_2 \leq \sqrt{\mathcal{N}_{\tilde{\boldsymbol{T}}^{\mathrm{a}}}}\epsilon_{\mathrm{abs}} + \epsilon^{\mathrm{rel}} \max \left\{\left\|\tilde{\boldsymbol{T}}^{\mathrm{a},n}\right\|_2, \left\|{\boldsymbol{S}}^{\mathrm{a},n}\right\|_2\right\} &\quad\quad 
		\left\|\boldsymbol{r}_{\boldsymbol{S}}^{\mathrm{d},n}\right\|_2 \leq \sqrt{\mathcal{N}_{\tilde{\boldsymbol{T}}^{\mathrm{a}}}}\epsilon_{\mathrm{abs}} + \epsilon^{\mathrm{rel}} \left\|\boldsymbol{V}^n\right\|_2 \\
	\end{aligned}
	\label{eq:stop}
\end{equation}
where $\epsilon_{\mathrm{abs}} \geq 0$ is the absolute tolerance, $\epsilon_{\mathrm{rel}} \geq 0$ is the relative tolerance, and the coefficients $\mathcal{N}_{(\cdot)}$ denote the dimensions of the corresponding vector. The choice of the relative and absolute tolerances depends on the size of the typical variable values in different applications \cite{MAL-016}. 

%{\col In this paper, we take $\epsilon_{\mathrm{rel}} = 10^{-2}$ and $\epsilon_{\mathrm{abs}} = 10^{-3}$ if not specified.
%%\begin{remark}
%%	It should be specifically noted that this would require gathering information from all agents in the algorithm to make a judgment, thus destroying the fully distributed nature of the algorithm.
%%\end{remark}
%}

%we show the pseudo-code of the proposed D-PDDP in Algorithm~\ref{algo: D-PDDP}. the algorithm reaches the maximum number of iterations $iter_{\max}$ or 
By summarizing the previous derivation results, {\colb the pseudo-code in Algorithm~\ref{algo: D-PDDP} shows the detailed procedure of the proposed D-PDDP.} It is worth noting that a single agent problem is computed once at the beginning of the algorithm using the original DDP (considering only the dynamics constraints \eqref{eq:k1}) to warm-start the subsequent algorithms (lines 1-3). During the iterations of the ADMM algorithm, Optimization Steps 1, 2 and 3 are strictly followed (lines 5-12) {\colb until the computational results satisfy specific convergence criteria.} In particular, it should be noted that necessary information communication is required before updating the global and dual variables. {\colb All computational steps (lines 1, 5-8, 10, 12) of our proposed D-PDDP method could be determined by each agent $i$ in a parallel way, while each agent $i$ should share informations with their neighbors through communication (lines 2, 9, 11). Both computational and communication properties reflect the fully decentralized nature of D-PDDP.}
%\begin{remark}
%	
%\end{remark}
%\begin{remark}
%	The start trajectory for the D-PDDP algorithm is initialized with the trajectory generated by the constrained DDP (CDDP) method \cite{9582791} and CDDP is a method that integrates the augmented Lagrange multiplier method and control projection approach in the original DDP. The algorithm is capable of handling nonlinear state and control constraints such as path obstacles and control limits.
%\end{remark}
\begin{remark}
	{\colb The stopping criterion of the ADMM algorithm can be determined by checking whether the residuals calculated from the primal and dual variables, named the primal residual and the dual residual, are below a certain threshold \cite{MAL-016}. However, it is important to note that the computation of the residuals requires gathering information of all agents in the algorithm, which would destroy the distributed nature of the D-PDDP.}
	%The stopping criterion for an ADMM algorithm can be determined by checking whether the primal and dual residuals of the algorithm are below a certain threshold \cite{MAL-016}. However, it should be specifically noted that this would require gathering information from all agents in the algorithm to make a judgment, thus destroying the fully distributed nature of the algorithm.
\end{remark}

\begin{remark}
	{\colb 
For distributed trajectory optimization problem of UAV swarm, the number of the swarm is large, and the nonlinear constraints are coupled with each other. For the practical computation, the performance of the algorithm is indeed sensitive to the initial guess as the convergence of the algorithm is related to both the initial flight time and the interference of the trajectory with obstacles and other UAV trajectories. For this reason, it is reasonable to use fast single agent trajectory optimization techniques, e.g., DDP algorithm, to compute an initial trajectory that satisfies the dynamics and flight time for each UAV trajectory independently.}
\end{remark}

\begin{remark}
	{\colb 
Whether there is a minimum set of neighbors that must be satisfied to guarantee convergence of the algorithm and satisfaction of the constraints is a theoretical aspect that has not yet been derived and investigated in this paper. However, in practical algorithmic studies, the size of the neighbor set can theoretically be changed in the process of trajectory generation, but changing the size of the neighbor set in the process of optimization is likely to influence the convergence, e.g., convergence speed, of the algorithm. If the set of neighbors is too small, each UAV can communicate with fewer neighbors and obtain less information, which may lead to the swarm trajectories needing more iterations to converge.}
\end{remark}

\begin{algorithm}[htb]
	%\textsl{}\setstretch{1.8}
	\renewcommand{\algorithmicrequire}{\textbf{Input:}}
	\renewcommand{\algorithmicensure}{\textbf{Output:}}
	\caption{Distributed Parameterized Differential Dynamic Programming}
	\label{algo: D-PDDP}
	\begin{algorithmic}[1]
		\STATE Initial guesses $(\bar{\boldsymbol{x}}_{i,k}, \bar{\boldsymbol{u}}_{i,k}, \bar{t}_{i,N})$ generated by DDP for each agent $i\in \mathcal{M}$. %Solve single-agent problems in parallel $\forall i \in \mathcal{M}$ with DDP.
		\STATE { Each agent $i \in \mathcal{M}$ communicates with all its neighbors $j \in \mathcal{N}_i \backslash\{i\}$ and receives $(\bar{\boldsymbol{x}}_{j,k}, \bar{t}_{j,N})$.} %Each agent $i \in \mathcal{M}$ receives $(\bar{\boldsymbol{x}}_{j,k}, \bar{t}_{j,N})$ from all $j \in \mathcal{N}_i \backslash\{i\}$.
		\STATE Initialize: $\boldsymbol{x}_{i,k} \leftarrow \bar{\boldsymbol{x}}_{i,k}, \boldsymbol{u}_{i,k} \leftarrow \bar{\boldsymbol{u}}_{i,k}, t_{i,N} \leftarrow \bar{t}_{i,N}, \tilde{\boldsymbol{x}}_{i,k}^{\mathrm{a}} \leftarrow\left[\left\{\bar{\boldsymbol{x}}_j\right\}_{j \in \mathcal{N}_i}\right], \tilde{\boldsymbol{t}}_{i,N}^{\mathrm{a}} \leftarrow\left[\left\{\bar{t}_j\right\}_{j \in \mathcal{N}_i}\right]$, $\tilde{u}_{i,k} \leftarrow \bar{\boldsymbol{u}}_{i,k}, \boldsymbol{z}_{i,k}^{\mathrm{a}} \leftarrow \tilde{\boldsymbol{x}}_{i,k}^{\mathrm{a}}, \boldsymbol{s}_{i,N}^{\mathrm{a}} \leftarrow \tilde{\boldsymbol{t}}_{i,N}^{\mathrm{a}}, \boldsymbol{\zeta}_{i,k} \leftarrow 0, \boldsymbol{\lambda}_{i,k} \leftarrow 0, \boldsymbol{y}_{i,k} \leftarrow 0, {\col \nu_{i,N}} \leftarrow 0, \boldsymbol{\eta}_{i,N} \leftarrow 0$.
		\WHILE{$n \leq iter_{\max}$}
			\STATE $(\boldsymbol{x}_{i,k}, \boldsymbol{u}_{i,k}, t_{i,N}) \leftarrow$ Eq. \eqref{eq:step1} in parallel for each agent $i \in \mathcal{M}$. %Solve \eqref{eq:step1} in parallel $\forall i \in \mathcal{M}$.
			\STATE $\tilde{\boldsymbol{x}}_{i,k}^{\mathrm{a}} \leftarrow$ Eq. \eqref{eq:stsub} in parallel for each agent $i \in \mathcal{M}$. %Solve \eqref{eq:stsub} in parallel for $\forall i \in \mathcal{M}$.
			\STATE $\tilde{\boldsymbol{u}}_{i,k} \leftarrow$ Eq. \eqref{eq:ctrsub} in parallel for each agent $i \in \mathcal{M}$. %Solve \eqref{eq:ctrsub} in parallel $\forall i \in \mathcal{M}$.
			\STATE $\tilde{\boldsymbol{t}}_{i,N}^{\mathrm{a}} \leftarrow$ Eq. \eqref{eq:tsub} in parallel for each agent $i \in \mathcal{M}$. %Solve \eqref{eq:tsub} in parallel $\forall i \in \mathcal{M}$.
			\STATE { Each agent $i \in \mathcal{M}$ communicates with $j \in \mathcal{P}_i \backslash\{i\}$ and receives $(\tilde{\boldsymbol{x}}_{i,k}^j,\tilde{t}_{i,N}^j)$.} %Each agent $i \in \mathcal{M}$ receives $(\tilde{\boldsymbol{x}}_{i,k}^j,\tilde{t}_{i,N}^j)$ from all $j \in \mathcal{P}_i \backslash\{i\}$.
			\STATE $(\boldsymbol{z}_{i,k},\boldsymbol{s}_{i,N}) \leftarrow$ in parallel for each agent $i \in \mathcal{M}$ %Update with \eqref{eq:glbces} in parallel $\forall i \in \mathcal{M}$.
			\STATE { Each agent $i \in \mathcal{M}$ communicates with $j \in \mathcal{N}_i \backslash\{i\}$ and receives $(\boldsymbol{z}_{j,k},s_{j,N})$.} %Each agent $i \in \mathcal{M}$ receives $(\boldsymbol{z}_{j,k},s_{j,N})$ from all $j \in \mathcal{N}_i \backslash\{i\}$.
			\STATE $\boldsymbol{\zeta}_{i,k}, \boldsymbol{\lambda}_{i,k}, \boldsymbol{y}_{i,k}, \nu_{i,N}, \boldsymbol{\eta}_{i,N} \leftarrow$ Update Eq. \eqref{eq:dualupd} in parallel $\forall i \in \mathcal{M}$.
			\STATE Calculate residuals using Eq. \eqref{eq:res}.
			\IF{Eq. \eqref{eq:stop}} 
				\STATE End the algorithm.
			\ENDIF 
		\ENDWHILE 
	\end{algorithmic}
\end{algorithm}

\section{Adaptive Penalty Parameter Based Acceleration for D-PDDP} \label{sec: ACD-PDDP}

%The resulting algorithm is the Adaptive Consensus D-PDDP (ACD-PDDP), which significantly improves the performance and reliability of the ADMM algorithm. Further details on the algorithm are provided below. 

The setting of the penalty parameters in the original ADMM algorithm is important because {\colb it has a significant influence on the algorithm's convergence rate. As a result, a distributed adaptive acceleration scheme incorporated into D-PDDP updates to accelerate the convergence of the algorithm by introducing an `agent-specific' penalty parameter tuning method for each agent is proposed in this section.} It is known that the ADMM update is equivalent to performing a Douglas Rachford Splitting (DRS) on the dual of the original problem \cite{10288223}. Based on this idea, we propose an adaptive penalty parameter updating rule, that leverage different penalty parameters for the state, control and time variables to take into account the constraint satisfaction of different agents, for the MAOC problem.

%For the distributed optimization problem of multiagent systems, each agent faces different constraints and different situations, so using different penalty parameters for the state, control and time variables of each agent can adequately take into account the constraint satisfaction of different agents. 
%
%Adaptive Consensus ADMM (ACADMM) proposed by Xu et al. \cite{10288223} verified the effectiveness of this idea by performing the acceleration effect on the ADMM algorithm, while we go one step further by proposing an acceleration scheme for each agent with a different penalty parameter on the basis of the D-PDDP algorithm. Xu et al. proved that the ADMM update step is equivalent to performing a Douglas Rachford Splitting on the dual of the original problem. Based on this idea, the adaptive penalty parameter updating rule extended for multi-agent problems with non-convex constraints can be derived.

\subsection{Derivition of Spectral Stepsize}
The specific derivation of the adaptive penalty parameter rule will be presented below using the primal variable $\boldsymbol{x}_{i,k}$, its copy variable $\tilde{\boldsymbol{x}}_{i,k}$ and their dual variable $\boldsymbol{\lambda}_{i,k}$ as the example for the sake of brevity of the article. First, define
\begin{equation}
	\begin{aligned}
		H\left(\boldsymbol{x}_{i,k}\right) &= J_i\left(\boldsymbol{x}_{i,k}, \boldsymbol{u}_{i,k}\right)+\mathcal{I}_{\boldsymbol{f}_i}\left(\boldsymbol{x}_{i,k}, \boldsymbol{u}_{i,k}\right) \\
		G\left(\tilde{\boldsymbol{x}}_{i,k}\right) &= \mathcal{I}_{\boldsymbol{a}_{i,k}}\left(\tilde{\boldsymbol{x}}_{i,k},\tilde{\boldsymbol{u}}_{i,k}\right)
	\end{aligned}
\end{equation}

%Similar to the definition in \eqref{adprob}, the dual form of a single agent of Problem \eqref{eq:prob2} with respect to state $\boldsymbol{x}_{i,k}$ and its copy variable $\tilde{\boldsymbol{x}}_{i,k}$ can be abstracted as
Based on the equivalence of the ADMM and the DRS on the dual problem of the original problem, problem \eqref{eq:prob2} with respect to $\boldsymbol{x}_{i,k}$ and $\tilde{\boldsymbol{x}}_{i,k}$ can be reformulated as its dual form as
\begin{equation}
\begin{split}
	&\min _{\boldsymbol{\lambda}_{i,k}} H^*\left(\boldsymbol{\lambda}_{i,k}\right)+G^*\left(\boldsymbol{\lambda}_{i,k}\right)\\
	&\text{s.t.}\quad \boldsymbol{x}_{i,k}-\tilde{\boldsymbol{x}}_{i,k}=0
	\end{split}
	\label{adprobdual}
\end{equation}
where $H^*$, $G^*$ denotes the Fenchel conjugate of ${H}$, ${G}$, defined as $H^*(\boldsymbol{\lambda}_{i,k})=\sup _{\boldsymbol{x}_{i,k}}\langle \boldsymbol{x}_{i,k}, \boldsymbol{\lambda}\rangle-H(\boldsymbol{x}_{i,k})$ and $G^*(\boldsymbol{\lambda}_{i,k})=\sup _{-\tilde{\boldsymbol{x}}_{i,k}}\langle -\tilde{\boldsymbol{x}}_{i,k}, \boldsymbol{\lambda}\rangle-G(-\tilde{\boldsymbol{x}}_{i,k})$. An important property is that $\boldsymbol{\lambda}_{i,k} \in \partial H(\boldsymbol{x}_{i,k})$ if and only if $\boldsymbol{x}_{i,k} \in \partial H^*(\boldsymbol{\lambda}_{i,k})$, and $\boldsymbol{\lambda}_{i,k} \in \partial G(-\tilde{\boldsymbol{x}}_{i,k})$ if and only if $-\tilde{\boldsymbol{x}}_{i,k} \in \partial G^*(\boldsymbol{\lambda}_{i,k})$ \cite{Rockafellar2001}.

According to the DRS, dual problem \eqref{adprobdual} can be decomposed into two steps for solving, i.e.,
\begin{subequations}
	\begin{align}
		& 0 \in \frac{\hat{\boldsymbol{\lambda}}_{i,k}^{n+1}-\boldsymbol{\lambda}_{i,k}^{n}}{\rho_{i}}+\partial H^*\left(\hat{\boldsymbol{\lambda}}_{i,k}^{n+1}\right)+\partial G^*\left(\boldsymbol{\lambda}_{i,k}^{n}\right)\label{eq:dualvar:a}\\
		& 0 \in \frac{\boldsymbol{\lambda}_{i,k}^{n+1}-\boldsymbol{\lambda}_{i,k}^{n}}{\rho_{i}}+\partial H^*\left(\hat{\boldsymbol{\lambda}}_{i,k}^{n+1}\right)+\partial G^*\left(\boldsymbol{\lambda}_{i,k}^{n+1}\right)\label{eq:dualvar:b}
	\end{align}
	\label{eq:dualvar}
\end{subequations}
where $\partial H^*(\cdot)$ and $\partial G^*(\cdot)$ stands for the subgradient of $H^*(\cdot)$ and $G^*(\cdot)$. And $\rho_{i}$ is no longer a fixed value, but a variable that changes according to the different agents and is equal at each time instant $k$. The intermediate variable $\hat{\boldsymbol{\lambda}}_{i,k+1}$ is defined as
\begin{equation}
	\hat{\boldsymbol{\lambda}}^{n+1}_{i,k} =\boldsymbol{\lambda}^n_{i,k}+\rho^n_{i}\left(\boldsymbol{x}^{n+1}_{i,k}-\boldsymbol{\tilde{x}}^n_{i,k}\right)
\end{equation}
where $\boldsymbol{x}^{n+1}_{i,k}$ and $\boldsymbol{\tilde{x}}^n_{i,k}$ are variables of different iterations. According to the important property of $H(\cdot)$ and $H^*(\cdot)$, we can readily obtain
\begin{equation}
	\begin{aligned}
		\boldsymbol{x}^{n+1}_{i,k} &\in \partial H^*(\hat{\boldsymbol{\lambda}}^{n+1}_{i,k})\\
		-\tilde{\boldsymbol{x}}^{n+1}_{i,k} &\in \partial G^*({\boldsymbol{\lambda}}^{n+1}_{i,k})
	\end{aligned}
	\label{eqrelat}
\end{equation}

Due to the equivalence between the ADMM step and the DRS of the dual problem, we can obtain the following update formula
\begin{equation}
	\begin{aligned}
		\hat{\boldsymbol{\lambda}}^{n+1}_{i,k} &=\boldsymbol{\lambda}^n_{i,k}+\rho^n_{i}\left(\boldsymbol{x}^{n+1}_{i,k}-\boldsymbol{\tilde{x}}^n_{i,k}\right)\\
		&\in \boldsymbol{\lambda}^n_{i,k} + \rho^n_{i} \left(\partial H^*(\hat{\lambda}^{n+1}_{i,k}) + \partial G^*({\lambda}^{n}_{i,k})\right)\\
		{\boldsymbol{\lambda}}^{n+1}_{i,k} &=\boldsymbol{\lambda}^n_{i,k}+\rho^n_{i}\left(\boldsymbol{x}^{n+1}_{i,k}-\boldsymbol{\tilde{x}}^{n+1}_{i,k}\right)\\
		&\in \boldsymbol{\lambda}^n_{i,k} + \rho^n_{i} \left(\partial H^*(\hat{\lambda}^{n+1}_{i,k}) + \partial G^*({\lambda}^{n+1}_{i,k})\right)
	\end{aligned}
	\label{DRS}
\end{equation}

Inspired by the fact that the spectral gradient method mimics the hessian of the original function to find a quasi-Newton step \cite{10.1093/imanum/8.1.141,pmlr-v54-xu17a}, the gradients of the functions $H^*$ and $G^*$ can be approximated as linear functions as
\begin{equation}
	\begin{aligned}
		\partial H^*(\hat{\boldsymbol{\lambda}}_{i,k}^{n})&\approx\alpha_{i}^{n} \hat{\boldsymbol{\lambda}}_{i,k}^{n}+\boldsymbol{a}_{i}^{n}\\
		\partial G^*({\boldsymbol{\lambda}}_{i,k}^{n})&\approx\beta_{i}^{n} {\boldsymbol{\lambda}}_{i,k}^{n}+\boldsymbol{b}_{i}^{n}
	\end{aligned}
	\label{linearapprox}
\end{equation}
{\colb where $\alpha_{i}^{n} > 0, \beta_{i}^{n} > 0$ are the first-order linear approximations of the dual functions $\partial H^*(\hat{\boldsymbol{\lambda}}_{i,k}^{n})$ and $\partial G^*({\boldsymbol{\lambda}}_{i,k}^{n})$, and the constant vector $\boldsymbol{a}_{i}^{n}, \boldsymbol{b}_{i}^{n} \in \mathbb{R}^r$.}

Substituting Eq.~\eqref{linearapprox} into DRS step \eqref{eq:dualvar}, we can obtain the relationship between $\boldsymbol{\lambda}_{i,k}^{n+1}$ and $\hat{\boldsymbol{\lambda}}_{i,k}^{n+1}$ with respect to $\boldsymbol{\lambda}_{i,k}^n$ as
\begin{equation}
	\begin{aligned}
		\hat{\boldsymbol{\lambda}}_{i,k}^{n+1} &=\frac{1-\beta_{i}^{n} \rho_{i}^{n}}{1+\alpha_{i}^{n} \rho_{i}^{n}} \boldsymbol{\lambda}_{i,k}^{n}-\frac{\boldsymbol{a}_{i}^{n} \rho_{i}^{n}+\boldsymbol{b}_{i}^{n} \rho_{i}^{n}}{1+\alpha_{i}^{n} \rho_{i}^{n}}\\
		\boldsymbol{\lambda}_{i,k}^{n+1} &=\frac{\left(1+\alpha_{i}^{n} \beta_{i}^{n} {\rho_{i}^{n}}^2\right) \boldsymbol{\lambda}_{i,k}^{n}-(\boldsymbol{a}_{i}^{n}+\boldsymbol{b}_{i}^{n}) \rho_{i}^{n}}{(1+\alpha_{i}^{n} \rho_{i}^{n})(1+\beta_{i}^{n} \rho_{i}^{n})}
	\end{aligned}
\end{equation}

The residual at $\boldsymbol{\lambda}_{i,k}^{n+1}$ can be expressed as the norm of the subgradient $\partial H^*(\boldsymbol{\lambda}_{i,k})$ and $\partial G^*(\boldsymbol{\lambda}_{i,k})$, i.e.,
\begin{equation}
	\begin{aligned}
		\left\|\boldsymbol{r}_{\boldsymbol{X}}^{\rm{d},n}\right\|_2 & =\left\|(\alpha_{i}^{n}+\beta_{i}^{n}) \boldsymbol{\Lambda}^{n+1}+(\boldsymbol{a}_{i}^{n}+\boldsymbol{b}_{i}^{n})\right\|_2 \\
		& =\frac{1+\alpha_{i}^{n} \beta_{i}^{n} {(\rho_{i}^{n})}^2}{(1+\alpha_{i}^{n} \rho_{i}^{n})(1+\beta_{i}^{n} \rho_{i}^{n})}\left\|(\alpha_{i}^{n}+\beta_{i}^{n}) \boldsymbol{\Lambda}^{n}+(\boldsymbol{a}_{i}^{n}+\boldsymbol{b}_{i}^{n})\right\|_2
	\end{aligned}
\end{equation}
where $\boldsymbol{\Lambda} $ is the slack of $\boldsymbol{\lambda}_{i,k}$ from all agents and all time instants.

The optimal penalty parameter for the $n$-th iteration is obtained by minimizing the residual $\left\|\boldsymbol{r}_{\boldsymbol{X}}^{\rm{d},n}\right\|_2$ as
\begin{equation}
	\begin{aligned}
		\rho_i^{n} & =\arg \min _{\rho_i} \left\|\boldsymbol{r}_{\boldsymbol{X}}^{\rm{d},n}\right\|_2 \\
		&=\arg \max _{\rho_i} \frac{(1+\alpha_{i}^{n} \rho_i)(1+\beta_{i}^{n} \rho_i)}{1+\alpha_{i}^{n} \beta_{i}^{n} \rho_i^2} \\
		& =\arg \max _{\rho_i} \frac{(\alpha_{i}^{n}+\beta_{i}^{n}) \rho_i}{1+\alpha_{i}^{n} \beta_{i}^{n} \rho_i^2}=1 / \sqrt{\alpha_{i}^{n} \beta_{i}^{n}}
	\end{aligned}
\end{equation}

\subsection{Local Curvature Estimation}
After the expression of the optimal step size is derived, the curvature information $\alpha_{i}^{n}$, $\beta_{i}^{n}$ needs to be determined. With \eqref{eqrelat} and \eqref{linearapprox}, it can be obtained that
\begin{equation}
	\begin{aligned}
		\boldsymbol{x}_{i,k}^{n} &\approx\alpha_i^{n} \hat{\boldsymbol{\lambda}}_{i,k}^{n}+\boldsymbol{a}_i^{n}\\
		-\tilde{\boldsymbol{x}}_{i,k}^{n} &\approx\beta_i^{n} {\boldsymbol{\lambda}}_{i,k}^{n}+\boldsymbol{b}_i^{n}
		\label{eqrelat2}
	\end{aligned}
\end{equation}

%These curvature parameters are estimated based on the results from iteration $n$ and earlier iterations $n_l < n$
{\colb The estimates for these curvature parameters are derived from the outcomes of iteration $n$ as well as previous iterations $n_l < n$} ($\alpha_i^n$, $\beta_i^n$ can be computed between a few iterations rather than at each iteration). Taking the estimation of $\alpha_i^n$ as an example, the following definitions are given as
\begin{equation}
	\begin{aligned}
		& \Delta \boldsymbol{x}_{i,k}^n =\boldsymbol{x}_{i,k}^n - \boldsymbol{x}_{i,k}^{n_l}\\
		& \Delta \hat{\boldsymbol{\lambda}}_{i,k}^n \coloneqq \hat{\boldsymbol{\lambda}}_{i,k}^n-\hat{\boldsymbol{\lambda}}_{i,k}^{n_l} \\
		& \Delta H_{i,k}^* \coloneqq \partial H^*(\hat{\boldsymbol{\lambda}}_{i,k}^n)-\partial H^*(\hat{\boldsymbol{\lambda}}_{i,k}^{n_l})
	\end{aligned}
\end{equation}

Considering the property of $\partial H^*(\hat{\boldsymbol{\lambda}}_{i,k}^n)$ in Eq. \eqref{eqrelat} and the linear approximating for $\partial H^*(\hat{\boldsymbol{\lambda}}_{i,k}^n)$ in Eq. \eqref{linearapprox}, we can get $\Delta \boldsymbol{x}_{i,k}^n \in \Delta H_{i,k}^*$ and $\Delta H_{i,k}^* \approx \alpha_i^n  \Delta \hat{\boldsymbol{\lambda}}_{i,k}^n + \boldsymbol{a}_i^n$. Then the value of curvature $\alpha_i^n$ can be estimated by the least square problem as \cite{Zhou2006}
\begin{equation}
	\min _{\alpha_i^n}\left\|\Delta \boldsymbol{x}_{i,k}^n-\alpha_i^n \Delta \hat{\boldsymbol{\lambda}}_{i,k}^n\right\|_2^2 \text { or } \min _{\alpha_i^n}\left\|{(\alpha_i^n)}^{-1} \Delta \boldsymbol{x}_{i,k}^n-\Delta \hat{\boldsymbol{\lambda}}_{i,k}^n\right\|_2^2 
\end{equation}
%With \eqref{eqrelat}, we have
%\begin{equation}
%	\Delta H_{i,k}^* = \Delta \boldsymbol{x}_{i,k}^n =\boldsymbol{x}_{i,k}^n - \boldsymbol{x}_i^{n_l}
%\end{equation}

Defining $\hat{\alpha}_i^n = 1/\alpha_i^n$ for notational convenience, the closed-form solutions of the curvature estimates $\hat{\alpha}_i^n$ corresponding to the preceding two least square problems are
\begin{equation}
	\hat{\alpha}_i^{n,\mathrm{SD}}=\frac{\left\langle\Delta \hat{\boldsymbol{\lambda}}_{i,k}^n, \Delta \hat{\boldsymbol{\lambda}}_{i,k}^n\right\rangle}{\left\langle\Delta \boldsymbol{x}_{i,k}^n, \Delta \hat{\boldsymbol{\lambda}}_{i,k}^n\right\rangle} \text { and } \hat{\alpha}_i^{n,\mathrm{MG}}=\frac{\left\langle\Delta \boldsymbol{x}_{i,k}^n, \Delta \hat{\boldsymbol{\lambda}}_{i,k}^n\right\rangle}{\left\langle\Delta \boldsymbol{x}_{i,k}^n, \Delta \boldsymbol{x}_{i,k}^n\right\rangle}
\end{equation}%where SD stands for steepest descent and MG for minimum gradient,
{\colb in which SD denotes the steepest descent while MG denotes the minimum gradient,} and $\hat{\alpha}_i^{n, \mathrm{SD}} \geq  \hat{\alpha}_i^{n, \mathrm{MG}}$ resulting from Cauchy-Schwarz inequality \cite{Zhou2006}. A hybrid stepsize calculation rule is utilized for better convergence as
\begin{equation}
	\hat{\alpha}_i^n = \begin{cases}\hat{\alpha}_i^{n,\mathrm{MG}} & \text { if } 2 \hat{\alpha}_i^{n,\mathrm{MG}}>\hat{\alpha}_i^{n,\mathrm{SD}} \\ \hat{\alpha}_i^{n,\mathrm{SD}}-\hat{\alpha}_i^{n,\mathrm{MG}} / 2 & \text { otherwise. }\end{cases}
	\label{alphacal}
\end{equation}

Similarly, the dual functions $G_i^*(\boldsymbol{\lambda}_i^n)$ has the same property, thus we can get the spectral stepsize $\hat{\beta}^n = 1/\beta^n$ with the same derivation process as
\begin{equation}
	\hat{\beta}_i^n= \begin{cases}\hat{\beta}_i^{n,\mathrm{MG}} & \text { if } 2 \hat{\beta}_i^{n,\mathrm{MG}}>\hat{\beta}_i^{n,\mathrm{SD}} \\ \hat{\beta}_i^{n,\mathrm{SD}}-\hat{\beta}_i^{n,\mathrm{MG}} / 2 & \text { otherwise. }\end{cases}
	\label{betacal}
\end{equation}
where $\hat{\beta}_i^{n,\mathrm{SD}} = \left\langle\Delta \boldsymbol{\lambda}_{i,k}^n, \Delta \boldsymbol{\lambda}_{i,k}^n\right\rangle /\left\langle\Delta \tilde{\boldsymbol{x}}_{i,k}^n, \Delta \boldsymbol{\lambda}_{i,k}^n\right\rangle$, $\hat{\beta}_i^{n,\mathrm{MG}} = \left\langle\Delta \tilde{\boldsymbol{x}}_{i,k}^n, \Delta \boldsymbol{\lambda}_{i,k}^n\right\rangle /\left\langle\Delta \tilde{\boldsymbol{x}}_{i,k}^n, \Delta \tilde{\boldsymbol{x}}_{i,k}^n\right\rangle$, with $\Delta G_i^* = \Delta \tilde{\boldsymbol{x}}_{i,k}^n = -\tilde{\boldsymbol{x}}_{i,k}^n+\tilde{\boldsymbol{x}}_i^{n_l}$, and $\Delta \boldsymbol{\lambda}_{i,k}^n = \boldsymbol{\lambda}_{i,k}^n - \boldsymbol{\lambda}_{i,k}^{n_l}$. It is worthy noting that $\hat{\alpha}_i^n$ and $\hat{\beta}_i^n$ are computed from intermediate quantities of different agents in different D-PDDP iterations, indicating that this update rule is especially useful for different agents in a distributed problem.

\subsection{Adaptive Penalty Parameter Update}
It should be pointed out that linear approximation of the gradients $\partial H_i^*$, $\partial G_i^*$ might be inaccurate, leading to invalid step sizes. {\colb Thus, it is necessary to evaluate the accuracy of the curvature estimations in order to guarantee the safety of the updated step size. To test the quality, the cosine of the angle between the variation of the dual sub-gradient as well as the variation of the dual variables are computed as}
%Therefore, it should be ensured that the updated step size is safe by assessing the quality of the curvature estimates. 
\begin{equation}
	\begin{aligned}
		\alpha_i^{n,\text {cor }}=\frac{\left\langle\Delta \boldsymbol{x}_{i,k}^n, \Delta \hat{\boldsymbol{\lambda}}_{i,k}^n\right\rangle}{\left\|\Delta \boldsymbol{x}_{i,k}^n\right\|\left\|\Delta \hat{\boldsymbol{\lambda}}_{i,k}^n\right\|}\\
		\beta_i^{n,\text {cor }}=\frac{\left\langle\Delta \tilde{\boldsymbol{x}}_{i,k}^n, \Delta {\boldsymbol{\lambda}}_{i,k}^n\right\rangle}{\left\|\Delta \tilde{\boldsymbol{x}}_{i,k}^n\right\|\left\|\Delta {\boldsymbol{\lambda}}_{i,k}^n\right\|}
	\end{aligned}
	\label{saferule}
\end{equation}

The spectral stepsizes are only allowed to be updated if the above assessment is lower than a pre-defined threshold. The specific update rule can be expressed as
\begin{equation}
	\rho_i^{n+1}= \begin{cases}\sqrt{\hat{\alpha}_i^n \hat{\beta}_i^n} & \text { if } \alpha_i^{n,\text {cor }}>\epsilon^{\text {cor }} \text { and } \beta_i^{n,\text {cor }}>\epsilon^{\text {cor }} \\ \hat{\alpha}_i^n & \text { if } \alpha_i^{n,\text {cor }}>\epsilon^{\text {cor }} \text { and } \beta_i^{n,\text {cor }} \leq \epsilon^{\text {cor }} \\ \hat{\beta}_i^n & \text { if } \alpha_i^{n,\text {cor }} \leq \epsilon^{\text {cor }} \text { and } \beta_i^{n,\text {cor }}>\epsilon^{\text {cor }} \\ \rho_i^n & \text { otherwise, }\end{cases}
	\label{updaterho}
\end{equation}
{\colb in which $\epsilon^{\text {cor}}$ is curvature estimations' accuracy threshold. Note that $\rho_i^{n+1}$ remains equal to $\rho_i^{n}$ from the previous computation when both curvature estimations are determined to be invalid.} In addition, it is necessary to limit the above parameter update to ensure the numerical stability of the acceleration algorithm, where the specific limiting rules are as follows \cite{pmlr-v70-xu17c}:
%Note that when both curvature estimates are considered inaccurate, $\rho_i^{n+1}$ keeps the same as $\rho_i^{n}$ obtained from the previous computation. 
\begin{equation}
	\rho_i^{n+1}=\max \left\{\min \left\{{\rho}_i^{n+1},\left(1+\frac{C_{\mathrm{cg}}}{n^2}\right) \rho_i^n\right\}, \frac{\rho_i^n}{1+C_{\mathrm{cg} / n^2}}\right\}
\end{equation}
where $C_{\mathrm{cg}}$ is a large convergence constant.

We summarize the pseudocode of the necessary steps for adaptive penalty parameters in Algorithm~\ref{algo: adapt}. Although only the derivation with respect to $(\boldsymbol{x}_{i,k}^n,\tilde{\boldsymbol{x}}_{i,k}^n,\boldsymbol{\lambda}_{i,k}^n)$ is presented above, there are similar derivation procedures for $(\boldsymbol{u}_{i,k}^n,\tilde{\boldsymbol{u}}_{i,k}^n,\boldsymbol{\zeta}_{i,k}^n), (\tilde{\boldsymbol{x}}_{i,k}^{\rm{a},n},\boldsymbol{z}_{i,k}^n,\boldsymbol{y}_{i,k}^n), (t_{i,N}^n,\tilde{t}_{i,N}^n,\nu_{i,N}^n), (\tilde{\boldsymbol{t}}_{i,N}^{\rm{a},n},\boldsymbol{s}_{i,N}^n,\boldsymbol{\eta}_{i,N}^n)$, of which the procedure for calculating the adaptive penalty parameters are also listed in Algorithm~\ref{algo: adapt}. This algorithm adds the content of the adaptive penalty parameter calculation without affecting Algorithm~\ref{algo: D-PDDP}. Algorithm~\ref{algo: adapt} shows that the results obtained in the intermediate steps of the D-PDDP algorithm can be used to obtain the adaptive penalty parameters through the spectral gradient method, which helps to reduce the number of iterations of the algorithm and accelerates the algorithm's efficiency. In order to increase the reliability of the algorithm, it is recommended to execute Algorithm~\ref{algo: adapt} every $C_{\it{Freq}}$ iteration steps instead of executing it at every step.

%\begin{remark}
%	Regarding the convergence analysis of the above acceleration algorithms, it has been shown that ACADMM is guaranteed to converge in the case of a bounded increase (decrease) \cite{pmlr-v70-xu17c}. The simulations in this paper also show that the adaptive tuning criterion converges stably in all the simulation cases.
%\end{remark}

\begin{algorithm}[htb]
	%\textsl{}\setstretch{1.8}
	\renewcommand{\algorithmicrequire}{\textbf{Input:}}
	\renewcommand{\algorithmicensure}{\textbf{Output:}}
	\caption{Adaptive Penalty Parameter Scheme for D-PDDP}
	\label{algo: adapt}
	\begin{algorithmic}[1]
		\STATE Initialize: $\tau_i^0, \rho_i^0, \gamma_i^0, \mu_i^0, \sigma_i^0, n_l=0$.
		\WHILE{Not Converge}
			\STATE Update D-PDDP using lines 5-11 in Algorithm~\ref{algo: D-PDDP}.
			\STATE Update dual variables using lines 12 in Algorithm~\ref{algo: D-PDDP} with adaptive penalty parameters $\tau_i^n, \rho_i^n, \mu_i^n, \gamma_i^n, \sigma_i^n$.
			\IF{$\mod(n, C_{\it{Freq}}) == 1$}
				\STATE Locally update intermediate variable $\hat{\boldsymbol{\zeta}}^{n+1}_{i,k} =\boldsymbol{\zeta}^n_{i,k}+\tau^n_i\left(\boldsymbol{u}^{n+1}_{i,k}-\tilde{\boldsymbol{u}}^n_{i,k}\right)$
				\STATE Locally update intermediate variable $\hat{\boldsymbol{\lambda}}^{n+1}_{i,k} =\boldsymbol{\lambda}^n_{i,k}+\rho^n_i\left(\boldsymbol{x}^{n+1}_{i,k}-\tilde{\boldsymbol{x}}^n_{i,k}\right)$
				\STATE Locally update intermediate variable $\hat{\boldsymbol{y}}^{n+1}_{i,k} =\boldsymbol{y}^n_{i,k}+\mu^n_i\left(\boldsymbol{x}^{\mathrm{a},n+1}_{i,k}-{\boldsymbol{z}}^{\mathrm{a},n}_{i,k}\right)$
				\STATE Locally update intermediate variable $\hat{\nu}^{n+1}_{i,N} =\nu^n_{i,N}+\sigma^n_i\left(t^{n+1}_{i,N}-\tilde{t}^n_{i,N}\right)$
				\STATE Locally update intermediate variable $\hat{\boldsymbol{\eta}}^{n+1}_{i,N} =\boldsymbol{\eta}^n_{i,N}+\gamma^n_i\left(t^{\mathrm{a},n+1}_{i,N}-s^{\mathrm{a},n}_{i,N}\right)$
				\STATE Locally compute spectral stepsizes and safely update $\rho_i^n$ with \eqref{alphacal}\eqref{betacal}\eqref{saferule}\eqref{updaterho}
				\STATE Locally compute spectral stepsizes and safely update $\tau_i^n,\mu_i^n,\gamma_i^n,\sigma_i^n$ with the same calculation steps as \eqref{alphacal}\eqref{betacal}\eqref{saferule}\eqref{updaterho}
			\ELSE 
				\STATE $\tau_i^{n+1} = \tau_i^{n}$
				\STATE $\rho_i^{n+1} = \rho_i^{n}$
				\STATE $\mu_i^{n+1} = \mu_i^{n}$
				\STATE $\gamma_i^{n+1} = \gamma_i^{n}$
				\STATE $\sigma_i^{n+1} = \sigma_i^{n}$
			\ENDIF
		\ENDWHILE 
	\end{algorithmic}
\end{algorithm}

%%%%%%%%%%%%%%%%%%%%%%%%%%%%%%%%%%%%%%%%%%

\section{Application Example: UAV Swarm Distributed Optimization}  \label{sec: application}
In this section, the effectiveness of the proposed methodology is validated through several numerical simulation studies using a fleet of UAV swarm. We first verify the effectiveness of the D-PDDP algorithm for distributed spatio-temporal trajectory optimization non-convex dynamics, obstacle constraints, inter-aircraft constraints, and terminal time constraints in four different scenarios. Then, we demonstrate the efficiency of the proposed adaptive penalty parameter updating approach by comparing it with other adaptive penalty rules. 

%All simulations are conducted in a workstation environment with i9-10900K and 64 GB RAM.

\subsection{Mission Scenario Setting}  \label{ssec:simset}
%We consider the problem of a swarm of UAVs in the horizontal plane. 
{\colb It is assumed} that each UAV in the swarm is moving with a constant speed in a two-dimensional plan. The dynamics of the UAV is an ideal mass model, without considering its attitude dynamics. Under this assumption, the kinematics model of each UAV can be described by
\begin{equation}
	\begin{aligned}
		\dot{x}_i &= V \cos \theta_i\\
		\dot{y}_i &= V \sin \theta_i\\
		\dot{\theta}_i &= {\col \omega_i}
	\end{aligned}
\end{equation}
where $\left[x_i,y_i,\theta_i\right]^T$ is the state variable of a single UAV, $\left[x_i,y_i\right]^T$ denotes the $i$-th UAV's coordinates and $\theta_i$ is the flight heading angle. The flight velocity of the UAV in the horizontal plane is set to a fixed value $V$. The UAV is controlled by the turning angular rate that adjusts the heading angle. Since the turning angular rate of the UAV is limited by physical constraints, we set the range of the turning rate $\omega_i$ to be $\left\lvert \omega_i\right\rvert \leq \omega_{\max}$, where $\omega_{\max}$ is a positive constant. {\colb Additionally, we establish $t_{\min} \leq t_N \leq t_{\max}$ as the limitations for each UAV's flying time.}
%Also, we set the lower and upper bounds of the flight time for each UAV as $t_{\min} \leq t_N \leq t_{\max}$.

We consider the mission that the swarm should approach the desired terminal states while respecting different constraints. To this end, a quadratic cost function $J_i(\boldsymbol{X}_i,\boldsymbol{U}_i)$ for all UAVs in the swarm is utilized as
\begin{equation}
	\begin{aligned}
		J_i(\boldsymbol{X}_i,\boldsymbol{U}_i) & = \frac{1}{2}{\left( {{\boldsymbol{x}_{i,N}} - {\boldsymbol{x}_{i,N}^{\mathrm{d}}}} \right)^{\rm{T}}}\mathbf{W}_i^{N} \left( {{\boldsymbol{x}_{i,N}} - {\boldsymbol{x}_{i,N}^{\mathrm{d}}}} \right)  \\
		&  + \sum\limits_{k = 0}^{N-1} { \left[ \frac{1}{2}{\boldsymbol{u}}_{i,k}^{\rm{T}} \mathbf{R}_i {{\boldsymbol{u}}_{i,k}} + \frac{1}{2}{{({{\boldsymbol{x}}_{i,k}} - {{\boldsymbol{x}}_i^{d}})}^{\rm{T}}} \mathbf{W}_i^s({{\boldsymbol{x}}_{i,k}} - {{\boldsymbol{x}}_i^{d}}) \right] }
	\end{aligned}
	\label{eq:cost1}
\end{equation}
where $\boldsymbol{x}_{i,N}^{d} = \left[x_i^{d}, y_i^{d}, \theta_i^{d}\right]^{T}$ denotes the desired terminal state vector of the $i$-th UAV . $\mathbf{W}_i^N$, $\mathbf{R}_i$ and $\mathbf{W}_i^s$ are weighting diagonal matrices of the cost function. 

%{\col The magnitude of the original cost weighting matrices $\mathbf{W}_i^N$, $\mathbf{R}_i$, $\mathbf{W}_i^s$ and the penalty parameter $\tau$, $\rho$, $\mu$, $\sigma$, $\gamma$ in \textit{optimization steps 1} and \textit{optimization steps 2} is a tradeoff between local optimization and inter-agent consensus, which means that a larger original cost implies that the algorithm is updated with more emphasis on the results of local optimization while a larger penalty parameter implies that more emphasis is placed on consensus.}

(1) {\it Path constraints}: The obstacles present in the environment during flight are considered as a circular no-fly zone, i.e., 
\begin{equation}
	(x_{i,k} - x_o)^2 + (y_{i,k}-y_o)^2 \geq (r_o + d_o)^2
	\label{obscons}
\end{equation}
where $\left[x_o,y_o\right]^T$ denotes the center coordinates of the obstacle region, $r_o$ is the radius, {\colb the smallest safe distance that every agent needs to keep with the obstacle $o \in \mathcal{O}$ is denoted as $d_o$, and the collection of obstacles in the environment are represented as $\mathcal{O}$.}
%$d_o$ denotes the minimum safe distance from obstacle $o \in \mathcal{O}$, and $\mathcal{O}$ represents the collection of all obstacles.

(2) {\it Inter-agent constraints}: Each UAV within the swarm needs to maintain a safe distance from its neighbors to avoid collisions, and it also should keep within a certain distance from its neighbors to maintain the functionality of the communication link due to physical limitations of the communication devices. Hence, these constraints can be described as
\begin{subequations}
	\begin{align}
		(x_{i,k} - x_{j,k})^2 + (y_{i,k}-y_{j,k})^2 &\geq d_{\mathrm{col}}^2\label{collision}\\
		(x_{i,k} - x_{j,k})^2 + (y_{i,k}-y_{j,k})^2 &\leq d_{\mathrm{con}}^2\label{connection}
	\end{align}
	\label{eq:intercons}
\end{subequations}
where $d_{\mathrm{col}}$ denotes the safe distance and $d_{\mathrm{con}}$ denotes maximum effective communication distance between two UAVs. 

%Constraints \eqref{obscons} and \eqref{eq:intercons} force the UAV to be in a ring-shaped domain to its neighbors and both constraints need to be convexified using \eqref{convexification} when solved. These constraints can be modeled as QP problems for trajectory points at each time instant after convexification. Since inter-agent constraints with neighbors need to be determined at each update step of the algorithm, trajectory information about neighbors needs to be collected through communication to update the constraints \eqref{eq:intercons} after the completion of \textit{Optimization Step 1}. Therefore an additional communication step is required between \textit{Optimization Steps 1} and \textit{Steps 2}, where UAV $i$ sends trajectory $\boldsymbol{X}_i^n$ to its neighbors $j \in \mathcal{P}_i\setminus \{i\}$ for linearization.

(3) {\it Terminal time sequence constraints}: We consider the UAVs arrive at the targets sequentially with a specified time interval. This constraint can be mathematically formulated as
	\begin{equation}
		\mathbf{A}_i \tilde{\boldsymbol{t}}_i^{\mathbf{a}}= \boldsymbol{t}_{\delta_i}
		\label{timeeq}
	\end{equation}
where $\boldsymbol{t}_{\delta_i}$ is the time interval vector and the matrix $\mathbf{A}_i$ represents the time series relationship between agent $i$ and its neighbors as
\begin{equation}
	\mathbf{A}_i =  \left[\begin{array}{llll} 
							1 & -1 & \cdots & 0\\
							1 & \vdots & \ddots & \vdots\\
							\vdots & 0 & \cdots & -1\\
							1 & 0 & \cdots & 0
					\end{array}\right]
\end{equation}
where the $j$-th row of $\mathbf{A}_i$ represents the temporal relationship between UAV $i$ and its neighbor $j$, corresponding to the $j$-th row of $\boldsymbol{t}_{\delta_i}$ denoting the pre-defined time interval between $i$ and $j$. Notice that when $\boldsymbol{t}_{\delta_i}$ is a zero vector, it means that the time interval is zero and all UAVs arrive at the targets simultaneously. When $\boldsymbol{t}_{\delta_i}$ is a non-zero vector, each UAV then follows the time interval of the elements in $\boldsymbol{t}_{\delta_i}$. In the implementation of the D-PDDP algorithm, we use the relaxation form of constraint \eqref{timeeq} for better convergence as
\begin{equation}
	\vert \mathbf{A}_i \tilde{\boldsymbol{t}}_{i,N}^{\mathbf{a}} - \boldsymbol{t}_{\delta_i} \vert \leq \boldsymbol{\epsilon}_{t_\delta}
	\label{timeeq2}
\end{equation}
where $\boldsymbol{\epsilon}_{t_\delta}$ is a constant relaxation parameter vector.

% In this paper, we take $\epsilon_{\mathrm{rel}} = 6\times 10^{-2}$ and $\epsilon_{\mathrm{abs}} = 10^{-3}$.

The conditions on the parameters of the algorithmic cost function are summarized in Table~\ref{tab:condit1}.
\begin{table}[htb]
	\centering
	\caption{Initial conditions and aerodynamics.}
	\label{tab:condit1}
	\begin{tabular}{cc}
		\hline \hline
		Parameters & Value \\ \hline
		Terminal weighting matrix $\mathbf{W}_i^N$ & $\text{diag}(25{\mathbf I_3})$ \\
		Control weighting matrix $\mathbf{R}_i$ & $1$ \\
		State weighting matrix $\mathbf{W}_i^s$ & $\boldsymbol{0}$ \\
%		{\col Relaxation parameter $\boldsymbol{\epsilon}_{t_\delta}$} & \\
		Penalty $\tau^{0}$ of $\boldsymbol{U}_i$ & 0.2 \\
		Penalty $\rho^{0}$ of $\boldsymbol{X}_i$ & 2 \\
		Penalty $\mu^{0}$ of $\tilde{\boldsymbol{X}}_i^{\mathrm{a}}$ & 1 \\
		Penalty $\sigma^{0}$ of $\boldsymbol{T}_i$ & 2 \\
		Penalty $\gamma^{0}$ of $\tilde{\boldsymbol{T}}_i^{\mathrm{a}}$ & 1 \\
%		Number of discrete Nodes of Trajectory $N$ & 149 \\
		Damping coefficient of PDDP $\alpha_l$ & 0.4 \\
		Stopping threshold $\epsilon_{\mathrm{rel}}$, $\epsilon_{\mathrm{abs}}$ & $6\times 10^{-2}$, $10^{-3}$\\
		\hline \hline
	\end{tabular}
\end{table}
%& Communication maintenance distance $d_{\mathrm{con}}$ & 300m
%Table~\ref{tab:condit1} summarizes the relevant parametric conditions considered in the above description, including the cost function, dynamic model and spatial-temporal constraints, and UAV-related parameters 

\begin{remark}
	Nonlinear state constraints \eqref{obscons} and \eqref{eq:intercons} can be mathematically expressed in a vector form $\left\|\boldsymbol{p}_{i, k}-\boldsymbol{p}_o\right\|_2 \leq r_o$, where $\boldsymbol{p}_{i, k}= \left[x_i^{d}, y_i^{d}\right]^{T}$ represents the position of the $i$th UAV and $\boldsymbol{p}_o (o \in \mathcal{O})$ denotes the center coordinate of the region. This constraint is non-convex and needs to be solved with the necessary convexification before using the optimization toolbox. {\colb The constraints in this paper are convexified using a linear approximation around the nominal trajectory for the sake of simplicity}
	%For simplicity, we convexify the constraint around the nominal trajectory using a first-order Taylor expansion as
	\begin{equation}
		\begin{aligned}
			& \left\|\bar{\boldsymbol{p}}_{i,k}-\boldsymbol{p}_{o}\right\|_2+ \vartheta^{\rm{T}}\left[\left(\boldsymbol{p}_{i,k}-\boldsymbol{p}_{o}\right)-\left(\bar{\boldsymbol{p}}_{i,k}-\boldsymbol{p}_{o}\right)\right] \geq r_o
		\end{aligned}
		\label{convexification}
	\end{equation}
	where
	\begin{equation}
		\vartheta=\frac{\bar{\boldsymbol{p}}_{i,k}-\boldsymbol{p}_{o}}{\left\|\bar{\boldsymbol{p}}_{i,k}-\boldsymbol{p}_{o}\right\|_2}
	\end{equation}
	where $\bar{\boldsymbol{p}}_{i,k}$ is the position of vehicle $i$ corresponding to the nominal trajectory.
\end{remark}
%\begin{remark}
%	Constraints \eqref{obscons} and \eqref{eq:intercons} can be modeled as QP problems for trajectory points at each time instant after convexification. Since inter-agent constraints with neighbors need to be determined at each update step of the algorithm, trajectory information about neighbors needs to be collected through communication to update the constraints \eqref{eq:intercons} after the completion of \textit{Optimization Step 1}. Therefore an additional communication step is required between \textit{Optimization Steps 1} and \textit{Steps 2}, where UAV $i$ sends trajectory $\boldsymbol{X}_i^n$ to its neighbors $j \in \mathcal{P}_i\setminus \{i\}$.
%\end{remark}

\subsection{Characteristics of Proposed Algorithm} \label{ssec:simulations}
In this subsection, we evaluate the distributed spatial-temporal optimization capability of the proposed D-PDDP algorithm for UAV swarms in four different scenarios through numerical simulations. In all simulations, we use the normal version of D-PDDP, which has a constant penalty parameter for all agents without any parameter adaptation. {\colm The initial guess of the D-PDDP algorithm uses the trajectory optimized by the DDP algorithm, which not considering obstacles, neighbors and free terminal time, and only considering each UAV itself. The constraints considered in the four scenarios include path constraints with obstacles, and inter-intelligence constraints with collision avoidance and communication maintenance, as well as terminal time-series constraints that satisfy certain time intervals. In order to quickly distinguish difference between the scenarios, the constraints in the 4 different scenarios are summarized in Table \ref{table:comparison}. The Scenario 2 and 3 mainly aim to illustrate that the proposed D-PDDP algorithm is able to compute the UAV swarm trajectory optimization problem that considering temporal constraints with time interval through different scenario settings. The Scenario 4 are mainly used to illustrate that the proposed D-PDDP algorithm is able to solve trajectory optimization problems in larger scale (tens of them) UAV swarm problems.}
\begin{table}[htb]
	\centering
	\caption{\col The constraints for the 4 scenarios.}
	\label{table:comparison}
	\setlength{\tabcolsep}{1.5mm}{
	\begin{tabular}{ccccc}
		\hline \hline
		Constraints & Scenario 1 & Scenario 2 & Scenario 3 & Scenario 4 \\ \hline
		Obstacles & $\surd$ & $\surd$ & $\surd$ & $\surd$ \\
		Collision avoidance & $\surd$ & $\surd$ & $\surd$ & $\surd$ \\
		Communication maintenance & $\surd$ & $\surd$ & $\surd$ & $\surd$ \\
		Terminal time-series & $\times$ & $\surd$ & $\surd$ & $\times$ \\ \hline\hline
	\end{tabular}}
\end{table}
%$149$ nodes are used to discretize the dynamic system as well as the trajectory. 

\textit{Scenario 1}: We consider four UAVs in Scenario 1, where each two of them flies towards the other two UAVs. There is a circular obstacle in the environment from which all UAVs need to maintain a safe distance from it. In addition, the flight times of all UAVs are free and not subject to constraint \eqref{timeeq}, which means that each UAV can optimize its own flight time without satisfying specific time sequence requirements with its neighbors (although each UAV still needs to share time variable with its neighbors and reach consensus). {\col The four swarm UAVs in the scenario are neighbors of each other, which means that the constraint \eqref{collision} is naturally satisfied in this scenario, i.e., each UAV needs to communicate with all other UAVs. Due to the need for communication, the UAV need to stay within communication distance from the other UAVs. The communication distance is set slightly larger than the distance between the UAV and its farthest neighbor. Table~\ref{tab:condit2} summarizes the specific state of each UAV, the positions and radius of the obstacle, the inter-agent distance, and the initialized values and bounds of the flight time. Notice that flight time here is the total flight duration of the vehicle, and its value changes constantly with iterations of trajectory optimization. In contrast, the initial guess of flight time is the empirical flight time value before starting trajectory optimization, which is used only once.}
\begin{table}[htb]
	\centering
	\caption{Initial conditions of Scenario 1.}
	\label{tab:condit2}
	\begin{tabular}{cc}
		\hline \hline
		Parameters & Value \\ \hline
		Initial state of UAV 1 $\left[x_{1_0},y_{1_0},\theta_{1_0}\right]^{\rm{T}}$ & $\left[15.0m, 110.0m, 0^{\circ}\right]^{\rm{T}}$\\
		Initial state of UAV 2 $\left[x_{2_0},y_{2_0},\theta_{2_0}\right]^{\rm{T}}$ & $\left[15.0m, 140.0m, 0^{\circ}\right]^{\rm{T}}$\\
		Initial state of UAV 3 $\left[x_{3_0},y_{3_0},\theta_{3_0}\right]^{\rm{T}}$ & $\left[285.0m, 110.0m, 180^{\circ}\right]^{\rm{T}}$\\
		Initial state of UAV 4 $\left[x_{4_0},y_{4_0},\theta_{4_0}\right]^{\rm{T}}$ & $\left[285.0m, 140.0m, 180^{\circ}\right]^{\rm{T}}$\\
		Target state of UAV 1 $\left[x_{1_f},y_{1_f},\theta_{1_f}\right]^{\rm{T}}$ & $\left[285.0m, 110.0m, 0^{\circ}\right]^{\rm{T}}$\\
		Target state of UAV 2 $\left[x_{2_f},y_{2_f},\theta_{2_f}\right]^{\rm{T}}$ & $\left[285.0m, 140.0m, 0^{\circ}\right]^{\rm{T}}$\\
		Target state of UAV 3 $\left[x_{3_f},y_{3_f},\theta_{3_f}\right]^{\rm{T}}$ & $\left[15.0m, 110.0m, 180^{\circ}\right]^{\rm{T}}$\\
		Target state of UAV 4 $\left[x_{4_f},y_{4_f},\theta_{4_f}\right]^{\rm{T}}$ & $\left[15.0m, 140.0m, 180^{\circ}\right]^{\rm{T}}$\\
		Velocity of each UAV $V_i$ & $30m/s$\\
		UAV turning maneuverability $\omega_{\max}$ & $0.5768 \textit{rad}/s$ \\
		Initial guess of flight time of each UAV $t_{i,N}$ & $9.3s$\\
		Lower and upper bound of flight time $\left[t_{\min}, t_{\max}\right]$ & $\left[0.1s, 20s\right]$ \\
		Safe distance to obstacle $d_{o}$ and neighbors $d_{\mathrm{col}}$ & $10m, 10m$\\
		{\col Communication maintenance distance with neighbors} $d_{\mathrm{col}}$ & $300m$\\
		Center and Radius of circular obstacle $p_{o},r_o$ & $\left[150.0m,125.0m\right]^{\rm{T}},20.0m$\\
		\hline \hline
	\end{tabular}
\end{table}
%Communication maintenance distance $d_{\mathrm{con}}$ & 300m\\

\begin{figure*}[htb]
	\centering
	\subfigure[\label{fig:f1:1}The initial guess trajectory (transparent line) and optimized flight trajectory (solid line) of each UAV.] %
		{\includegraphics[width=0.30\textwidth]{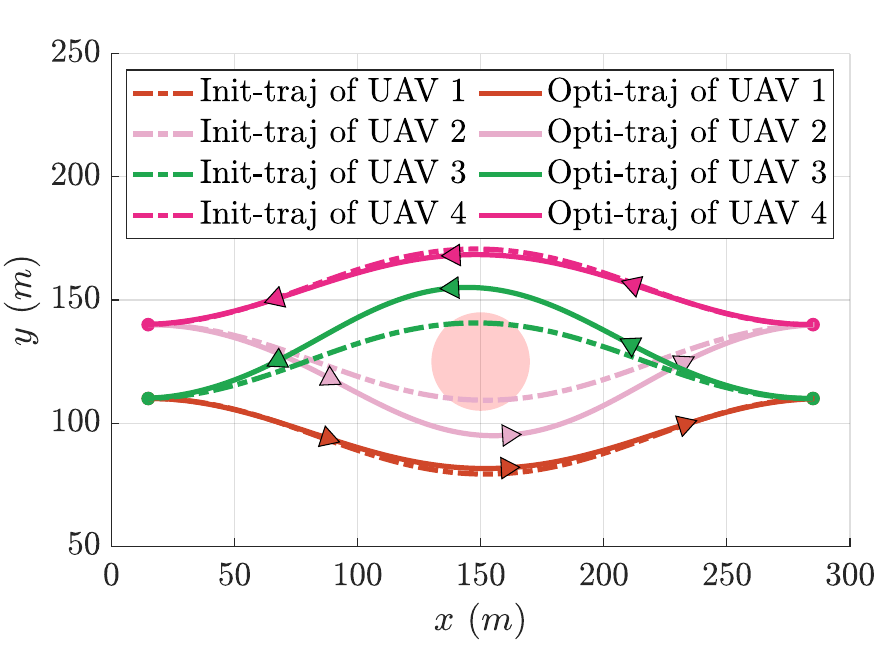}}
		\quad
	\subfigure[\label{fig:f1:2}Control angular velocity variation curve of each UAV.] %
		{\includegraphics[width=0.30\textwidth]{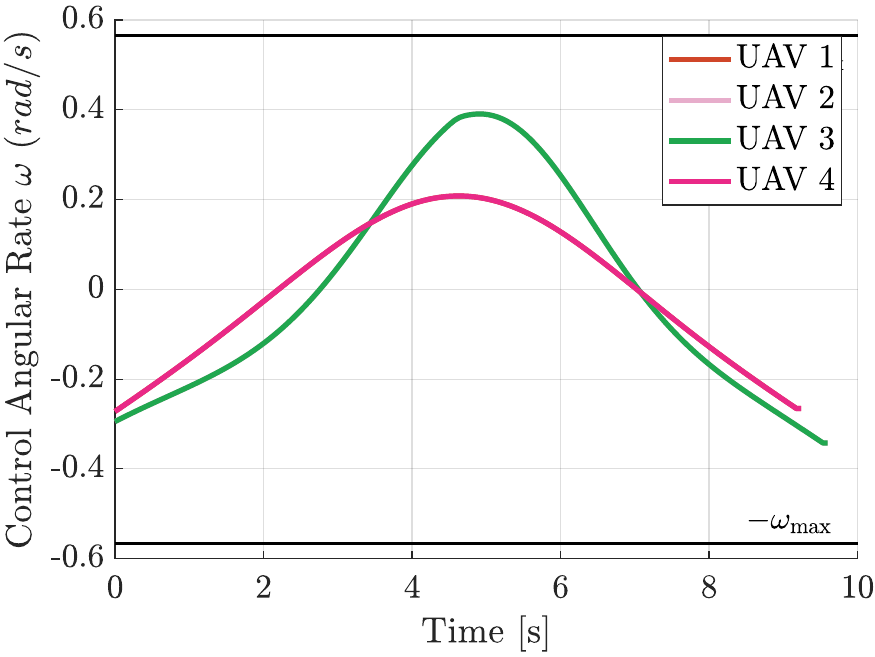}}
		\quad
	\subfigure[\label{fig:f1:3}The variation curve of the flight time for each UAV with the number of iterations.] %
		{\includegraphics[width=0.30\textwidth]{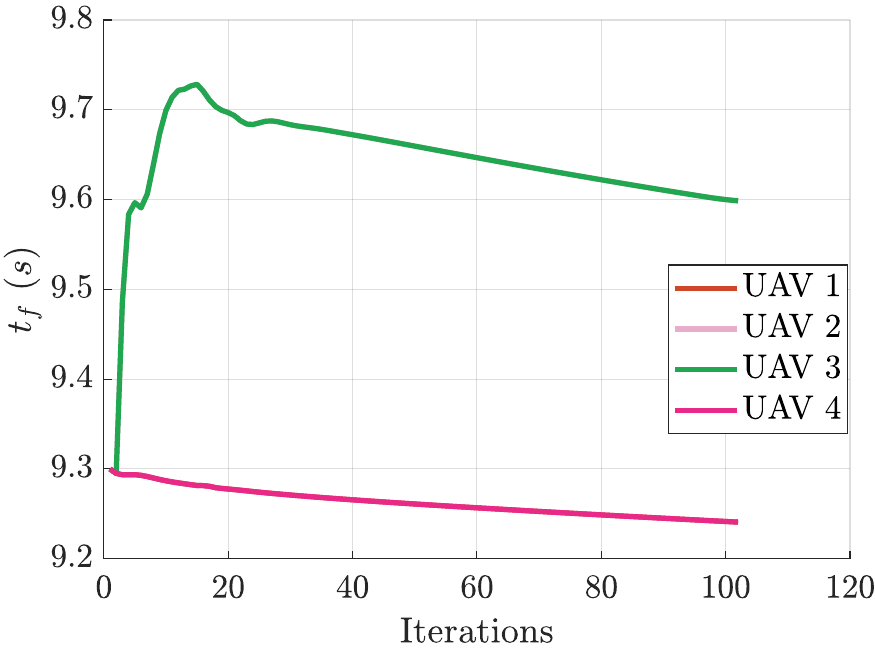}} \\
	\subfigure[\label{fig:f1:4}The snapshot of the scene at the initial moment of flight $t_0$.] %
		{\includegraphics[width=0.30\textwidth]{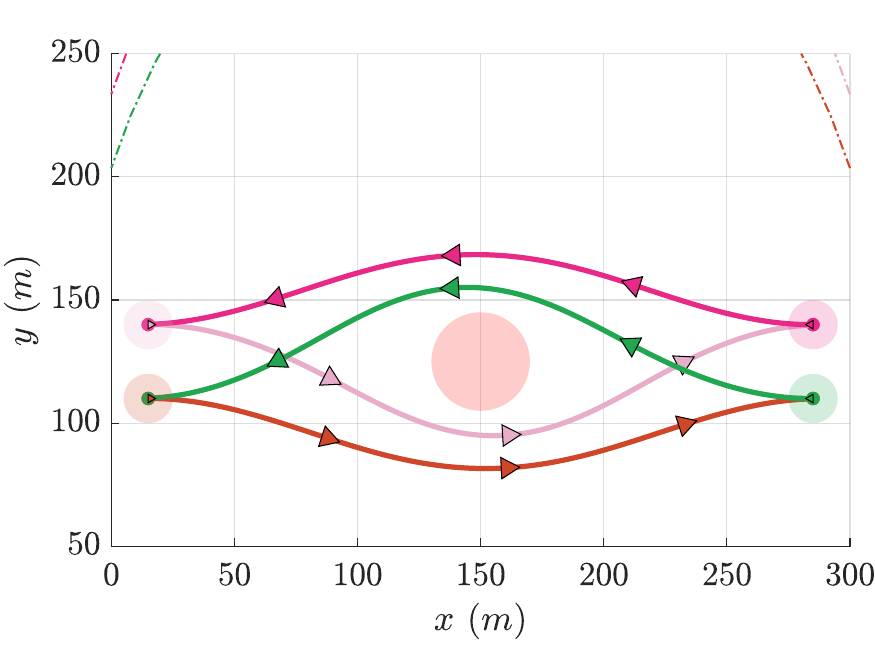}}
		\quad
	\subfigure[\label{fig:f1:5}The snapshot of the scene at the moment of minimum distance between the vehicles.] %
		{\includegraphics[width=0.30\textwidth]{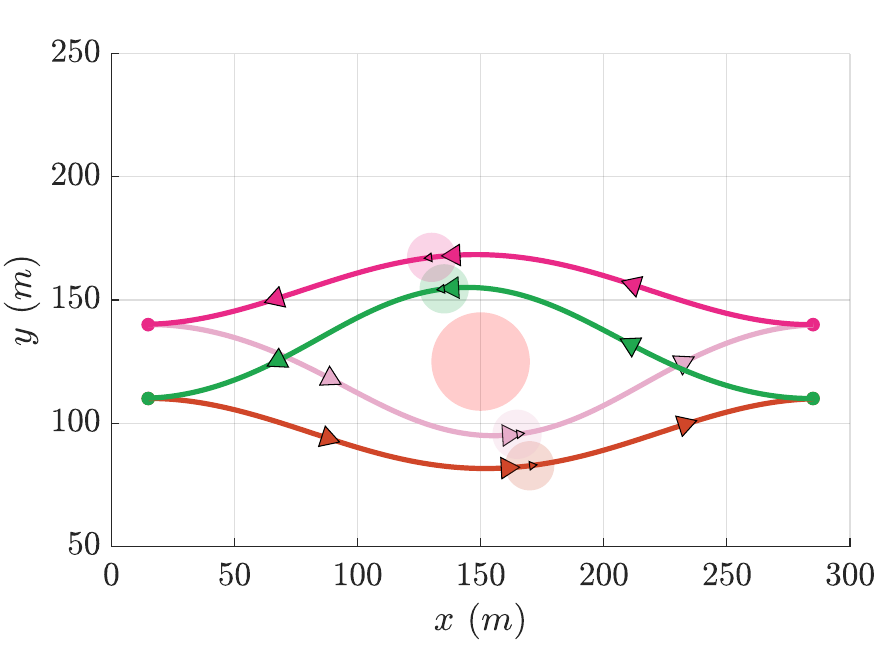}}
		\quad
	\subfigure[\label{fig:f1:6}The snapshot of the scene at the terminal moment of flight $t_f$.] %
		{\includegraphics[width=0.30\textwidth]{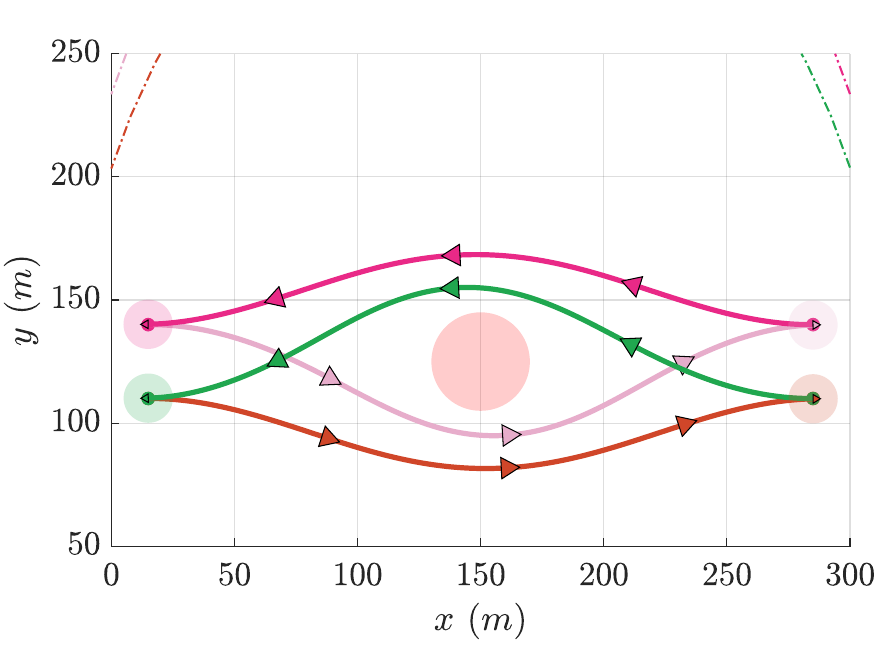}}
	\caption{ \label{fig:f1:plain}
	Distributed spatial-temporal joint optimization results for Scenario 1.}
\end{figure*}

The optimization results of the four UAVs in Scenario 1 are shown in Fig.~\ref{fig:f1:plain}. In Fig.~\ref{fig:f1:1}, the dashed lines are the initial guesses of the flight trajectories and the solid lines are the optimized results, from which we can see that the proposed D-PDDP method can optimize from the initialized trajectory that violates the obstacle constraints to obstacle avoidance. The trajectory results show that the optimized trajectories are able to avoid obstacles in the environment beyond the safe distance while the trajectory of each UAV maintains a safe distance from its neighbors. The control history of the four UAVs is shown in Fig.~\ref{fig:f1:2}. Since the constraint \eqref{timeeq} is not required to be satisfied in this scenario, the four UAVs optimize their own flight times from $9.3s$ to $9.24s$ and $9.60s$, as can be seen from the change in trajectories from the initial guess to the optimized results in Fig.~\ref{fig:f1:1} and the convergence of the UAV flight times with the iterations of the algorithm in Fig.~\ref{fig:f1:3}. Fig.~\ref{fig:f1:4} - Fig.~\ref{fig:f1:6} show the snapshots of the UAV flying according to the optimized trajectory after the distributed optimization. The circles above and below the obstacle in the figure that match the color of the trajectory represent the safe distance range of the UAVs, where Fig.~\ref{fig:f1:5} is drawn at the closest point ($13.77m$) of the flight paths of the two UAVs. {\col The dashed circle indicates the range of the communication distance of the UAV. Since the four UAVs are able to communicate with each other, all four UAVs are within each other's maximum communication distance circle, as shown in the figures.} In summary, the results demonstrate that the proposed D-PDDP provides the ability to optimize the flight trajectories of multiple UAVs to avoid obstacles and satisfy inter-aircraft avoidance constraints while optimizing their own flight time.

\textit{Scenario 2}: A five-UAV consecutive interception scenario is considered to validate the optimization capability of the D-PDDP algorithm for time-series tasks. The UAV swarm intercepts different targets in the same region from different orientations with the requirement of position and flight path angle constraints. {\col The 5 swarm UAVs need to communicate with their neighbors and maintain a safe distance while flying to the target point. The neighborhood size of each UAV is set to $|\mathcal{N}_i| = 3$.} In addition, the UAVs need to intercept the targets at certain time intervals, i.e., they satisfy the time-series constraints \eqref{timeeq2}. The initial time for all the UAVs is set to $9.0s$, the impact times of all UAVs satisfies equal interval $\boldsymbol{t}_{\delta_i}=0.1s$ constraints, and the relaxation variable in Eq. \eqref{timeeq2} is set to $0.01s$ for all UAV. To ensure higher accuracy of the time sequence constraints, we tune the stopping condition as $\epsilon_{\mathrm{abs}} = 5\times 10^{-4}$ in this scenario. In addition, the communication distance is set based on the initial position of the UAV using the closest distance criterion. {\col Table~\ref{tab:condit3}} summarizes the relevant settings for Scenario 2, where the same parameters as in Scenario 1 are not repeated.
\begin{table}[htb]
	\centering
	\caption{Initial conditions of Scenario 2.}
	\label{tab:condit3}
	\begin{tabular}{cc}
		\hline \hline
		Parameters & Value \\ \hline
		Initial state of UAV 1 $\left[x_{1_0},y_{1_0},\theta_{1_0}\right]^{\rm{T}}$ & $\left[328.75m, 26.5m, 60^{\circ}\right]^{\rm{T}}$\\
		Initial state of UAV 2 $\left[x_{2_0},y_{2_0},\theta_{2_0}\right]^{\rm{T}}$ & $\left[484.0m, 95.6m, 120^{\circ}\right]^{\rm{T}}$\\
		Initial state of UAV 3 $\left[x_{3_0},y_{3_0},\theta_{3_0}\right]^{\rm{T}}$ & $\left[587.0m, 242.8m, 120^{\circ}\right]^{\rm{T}}$\\
		Initial state of UAV 4 $\left[x_{4_0},y_{4_0},\theta_{4_0}\right]^{\rm{T}}$ & $\left[551.2m, 411.85m, 180^{\circ}\right]^{\rm{T}}$\\
		Initial state of UAV 5 $\left[x_{4_0},y_{4_0},\theta_{4_0}\right]^{\rm{T}}$ & $\left[437.5m, 538.16m, 240^{\circ}\right]^{\rm{T}}$\\
		Target state of UAV 1 $\left[x_{1_f},y_{1_f},\theta_{1_f}\right]^{\rm{T}}$ & $\left[302.6m, 275.14m, 120^{\circ}\right]^{\rm{T}}$\\
		Target state of UAV 2 $\left[x_{2_f},y_{2_f},\theta_{2_f}\right]^{\rm{T}}$ & $\left[324.45m, 281.42m, 180^{\circ}\right]^{\rm{T}}$\\
		Target state of UAV 3 $\left[x_{3_f},y_{3_f},\theta_{3_f}\right]^{\rm{T}}$ & $\left[342.45m, 294.8m, 210^{\circ}\right]^{\rm{T}}$\\
		Target state of UAV 4 $\left[x_{4_f},y_{4_f},\theta_{4_f}\right]^{\rm{T}}$ & $\left[322.84m, 310.17m, 240^{\circ}\right]^{\rm{T}}$\\
		Target state of UAV 5 $\left[x_{4_f},y_{4_f},\theta_{4_f}\right]^{\rm{T}}$ & $\left[312.5m, 321.65m, 270^{\circ}\right]^{\rm{T}}$\\
		{\col Initial guess of flight time of each UAV} $t_{i,N}$ & $9.0s$\\
		{\col Communication maintenance distance with neighbors} $d_{\mathrm{col}}$ & $380m$\\
		Center of circular obstacle $\boldsymbol{p}_{o_1}$ & $\left[400.0m,200.0m\right]^{\rm{T}}$\\
		Center of circular obstacle $\boldsymbol{p}_{o_2}$ & $\left[450.0m,380.0m\right]^{\rm{T}}$\\
		Radius of circular obstacle $r_o$ & $40.0m$\\
		\hline \hline
	\end{tabular}
\end{table}

\begin{figure*}[htb]
	\centering
	\subfigure[\label{fig:f2:1}The initial guess trajectory (transparent line) and optimized flight trajectory (solid line) of each UAV.] %
		{\includegraphics[width=0.30\textwidth]{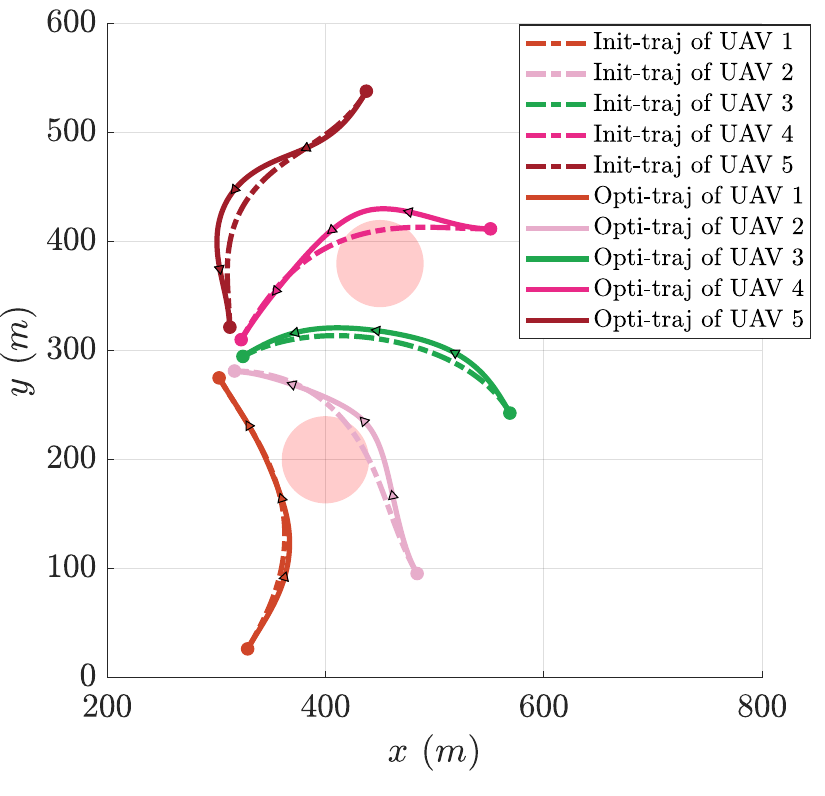}}
		\quad
	\subfigure[\label{fig:f2:2}Control angular velocity variation curve of each UAV.] %
		{\includegraphics[width=0.30\textwidth]{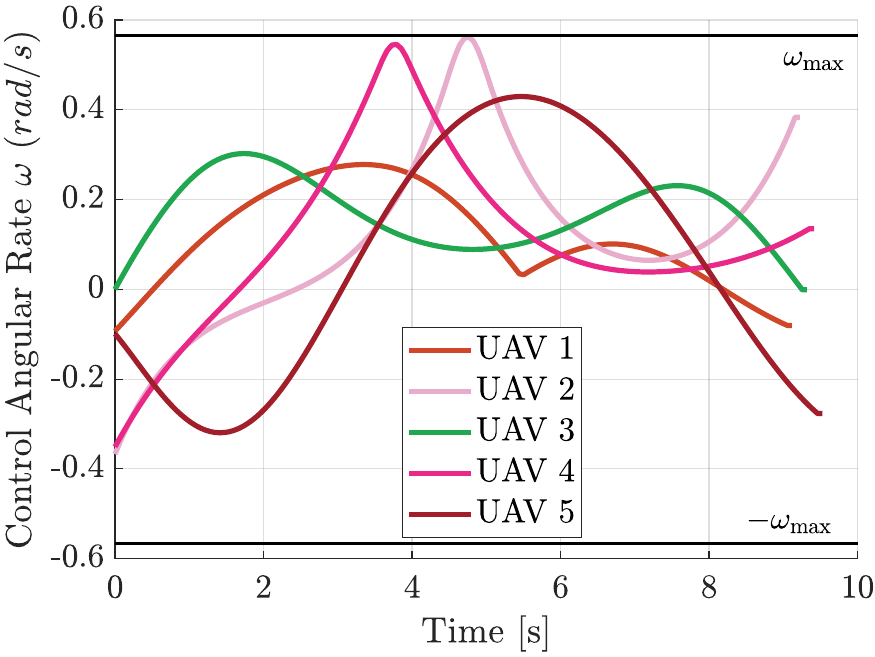}}
		\quad
	\subfigure[\label{fig:f2:3}The variation curve of the flight time for each UAV with the number of iterations.] %
		{\includegraphics[width=0.30\textwidth]{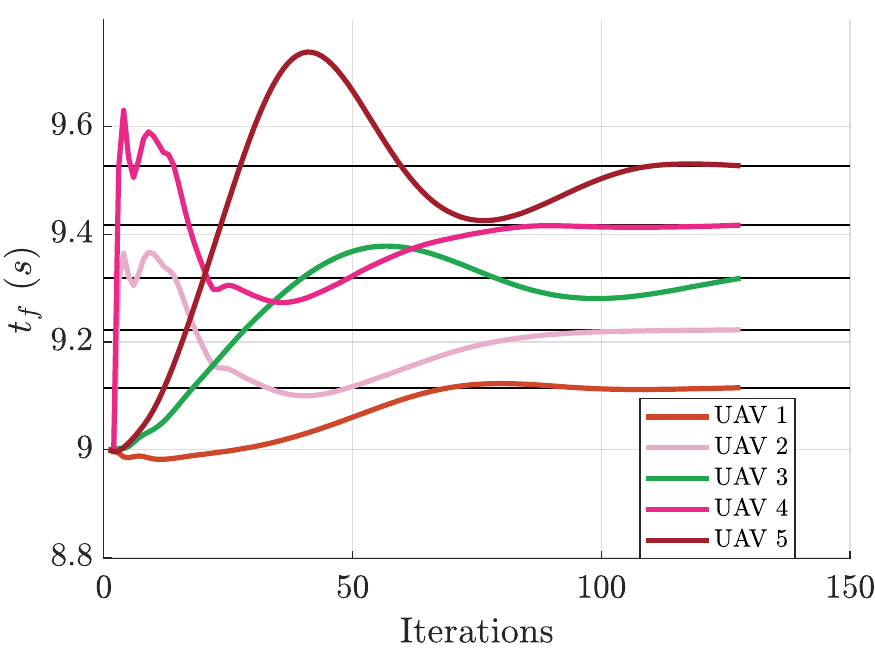}}% \\
		\quad
	\subfigure[\label{fig:f2:4}The snapshot of the scene at the initial moment of flight $t_0$.] %
		{\includegraphics[width=0.30\textwidth]{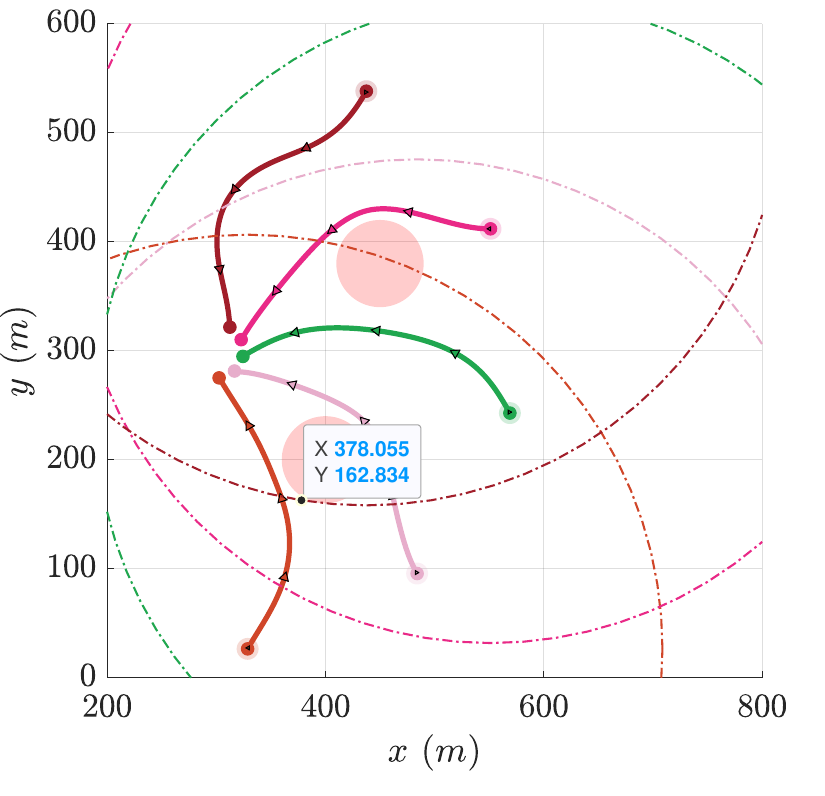}}
		\quad
	\subfigure[\label{fig:f2:5}The snapshot of the scene at the moment of minimum distance between the vehicles.] %
		{\includegraphics[width=0.30\textwidth]{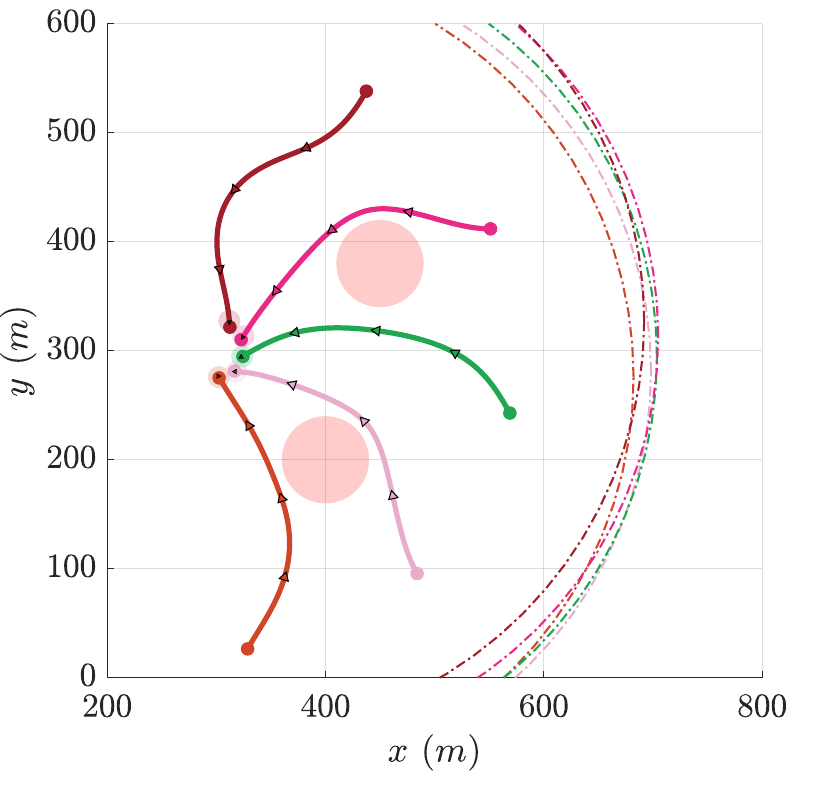}}
		\quad
	\subfigure[\label{fig:f2:6}The snapshot of the scene at the terminal moment of flight $t_f$.] %
		{\includegraphics[width=0.30\textwidth]{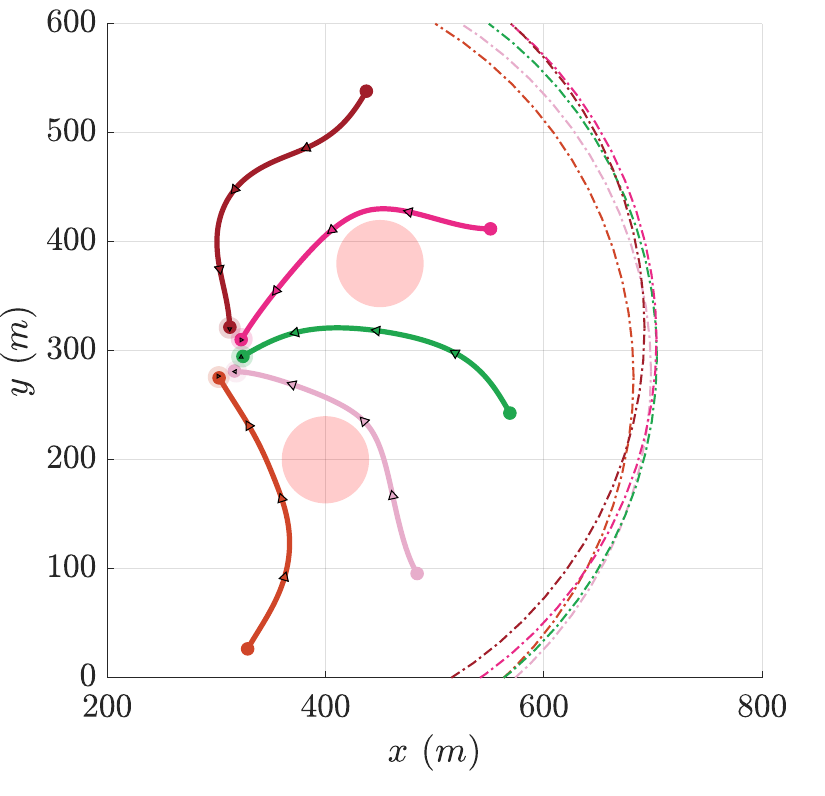}}
	\caption{ \label{fig:f2:plain}
	Distributed spatial-temporal joint optimization results for Scenario 2.}
\end{figure*}

{\col The optimization results of the UAV swarm in Scenario 2 are shown in Fig.~\ref{fig:f2:plain}, which contains the optimized trajectories, control history, time changes, and snapshots of the flights of the UAV swarm.} From Fig.~\ref{fig:f2:1}, it can be observed that the trajectories of UAV $2$ and UAV $4$ change from constraint-violating trajectories to obstacle-avoiding trajectories, and the trajectory of UAV $5$ is more curved compared to the initial guess to satisfy the time sequence constraints. {\col Similarly, it can be seen in Fig.~\ref{fig:f2:4} through Fig.~\ref{fig:f2:6} that all four UAVs are within each other's maximum communication distance circle and collision free, which means the UAVs satisfy the collision avoidance constraints and the communication maintenance constraints.} On the other hand, it can be seen from Fig.~\ref{fig:f2:3} that the flight times of the five UAVs are optimized from the same initial guess $9.0s$ to $9.12s$, $9.22s$, $9.32s$, $9.42s$, $9.53s$, respectively, which satisfy the consecutive interception constraint with time interval being $0.1s$ between two UAVs. The results illustrate that the D-PDDP algorithm is capable of accomplishing the UAV swarm trajectory optimization task with time sequence constraints through distributed spatial-temporal joint optimization.

\textit{Scenario 3}: We consider 16 UAVs evenly distributed in a circular formation to cross an obstacle at the center of the formation while flying to another point symmetric to the center of the circle in this scenario. The 16 swarm UAVs need to communicate with their neighbors and maintain a safe distance while flying to the target point. The neighborhood size of each UAV is set to $|\mathcal{N}_i| = 5$. In addition, all UAVs are required to complete flight missions simultaneously, i.e., they need to satisfy the constraint \eqref{timeeq}, where the time interval vector is $\boldsymbol{0}$. This means that each UAV needs to optimize its own flight time while keeping the same time as its neighbors. {\col Table~\ref{tab:condit4}} summarizes the relevant settings for Scenario 3.
\begin{table}[htb]
	\centering
	\caption{Initial conditions of Scenario 3.}
	\label{tab:condit4}
	\begin{tabular}{cc}
		\hline \hline
		Parameters & Value \\ \hline
		Center and radius of circular formation $\boldsymbol{p}_{\rm{form}},r_{\rm{form}}$ & $\left[0.0m, 0.0m\right]^{\rm{T}},270.0m$\\
		{\col Initial guess of flight time of each UAV} $t_{i,N}$ & $9.2s$\\
		Communication maintenance distance with neighbors $d_{\mathrm{con}}$ & $120m$\\
		Center and radius of circular obstacle $\boldsymbol{p}_{o},r_o$ & $\left[0.0m,0.0m\right]^{\rm{T}},20.0m$\\
		\hline \hline
	\end{tabular}
\end{table}

\begin{figure*}[htb]
	\centering
	\subfigure[\label{fig:f3:1}The initial guess trajectory (transparent line) and optimized flight trajectory (solid line) of each UAV.] %
		{\includegraphics[width=0.30\textwidth]{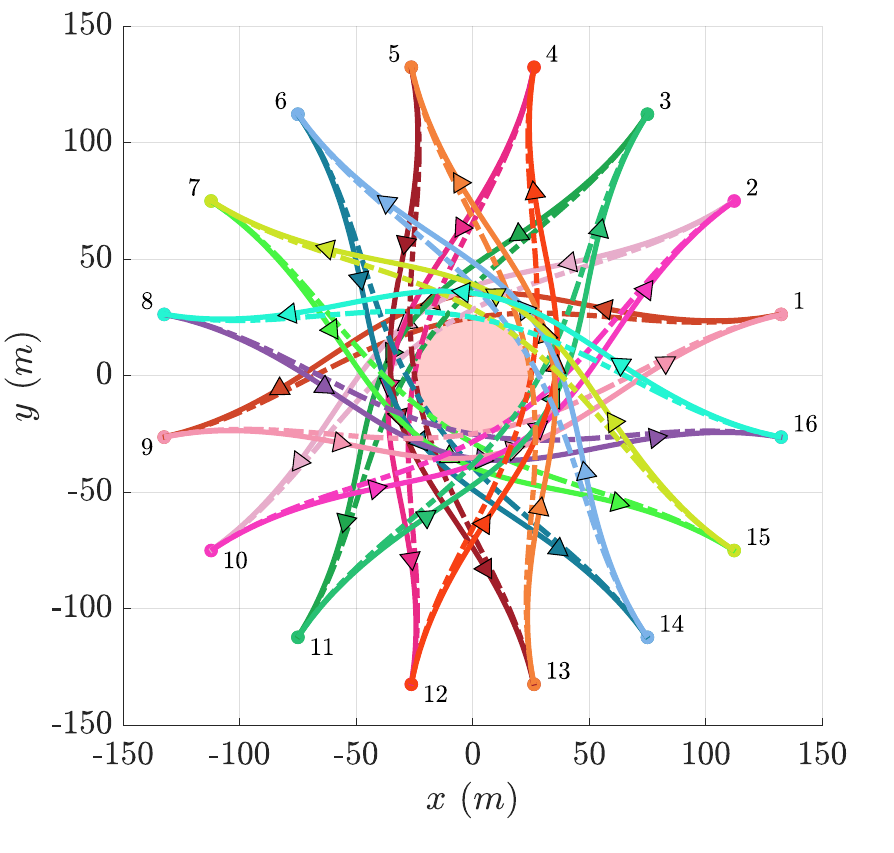}}
		\quad
	\subfigure[\label{fig:f3:2}Control angular velocity variation curve of each UAV.] %
		{\includegraphics[width=0.30\textwidth]{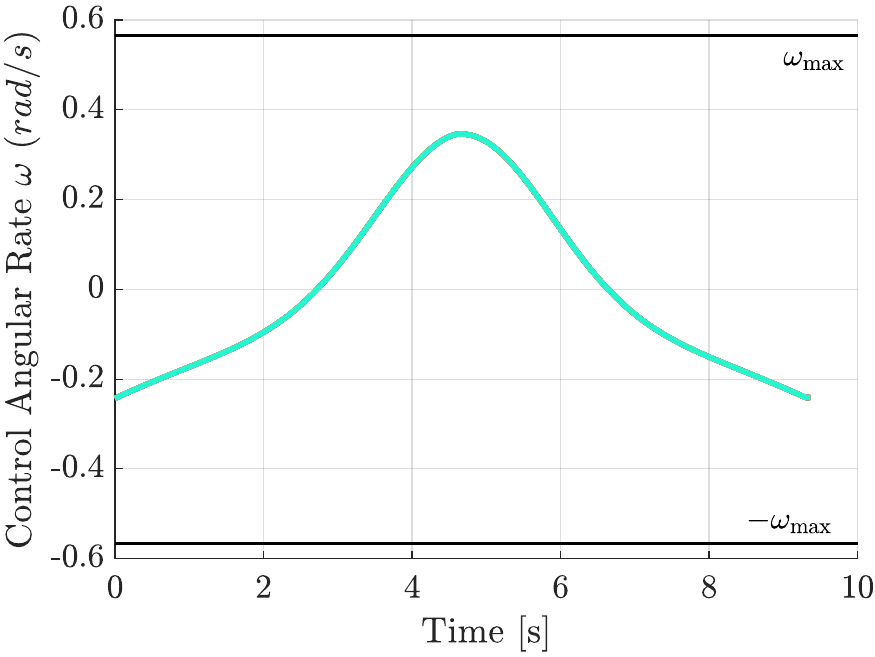}}
		\quad
	\subfigure[\label{fig:f3:3}The variation curve of the flight time for each UAV with the number of iterations.] %
		{\includegraphics[width=0.30\textwidth]{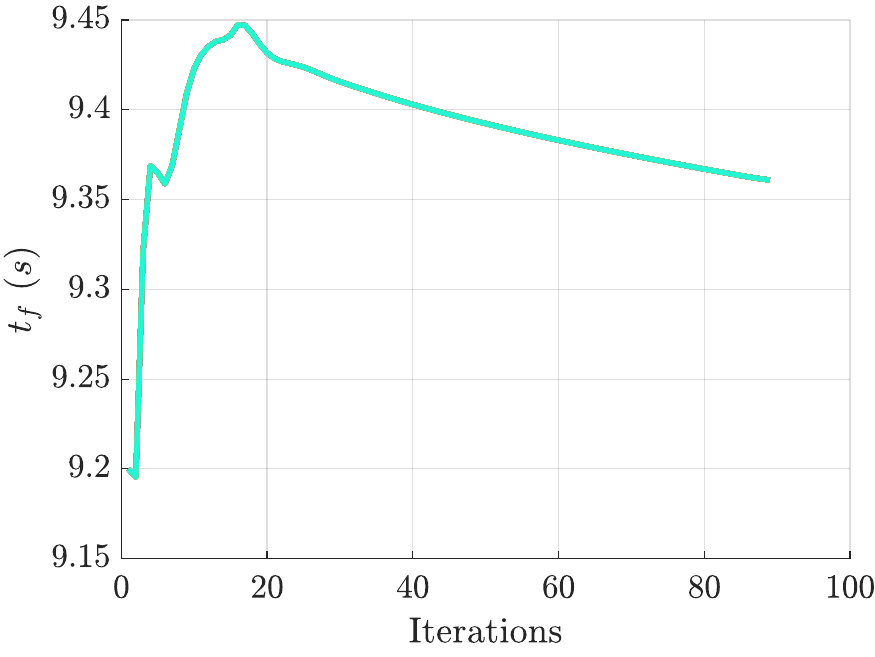}} \\
	\subfigure[\label{fig:f3:4}The snapshot of the scene at the initial moment of flight $t_0$.] %
		{\includegraphics[width=0.30\textwidth]{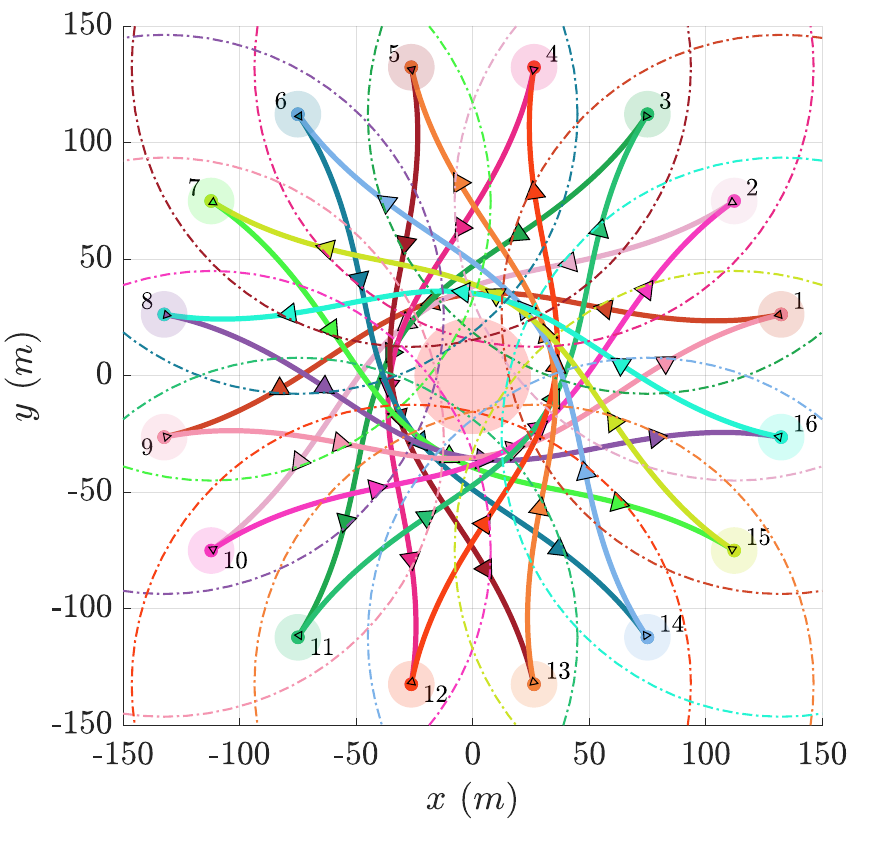}}
		\quad
	\subfigure[\label{fig:f3:5}The snapshot of the scene at the moment of minimum distance between the vehicles.] %
		{\includegraphics[width=0.30\textwidth]{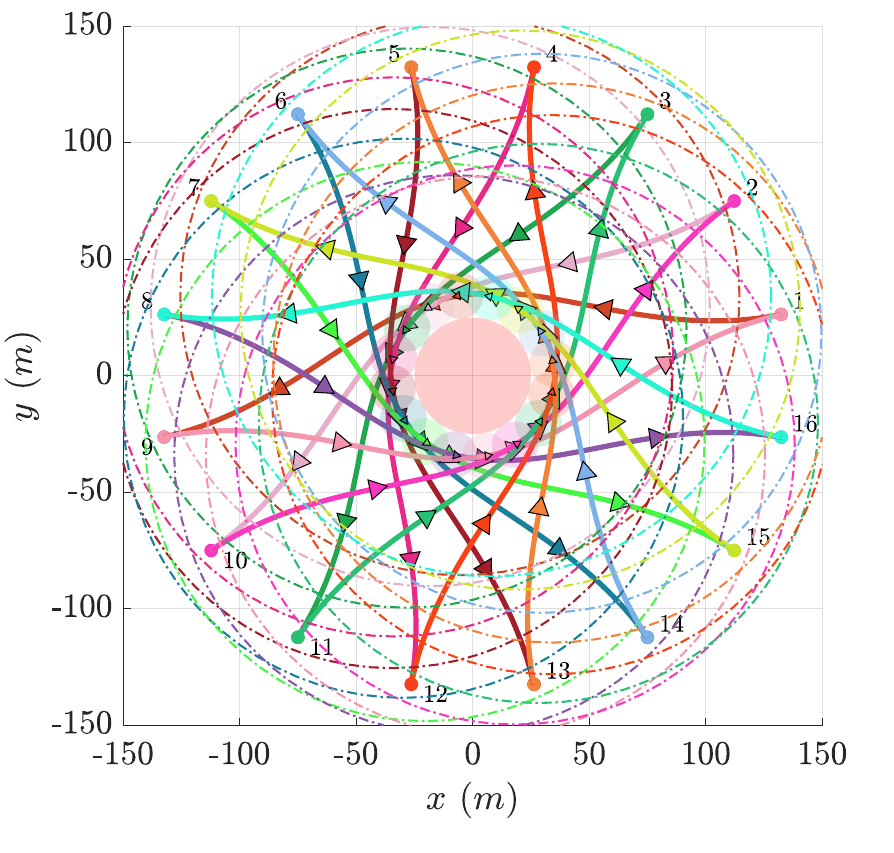}}
		\quad
	\subfigure[\label{fig:f3:6}The snapshot of the scene at the terminal moment of flight $t_f$.] %
		{\includegraphics[width=0.30\textwidth]{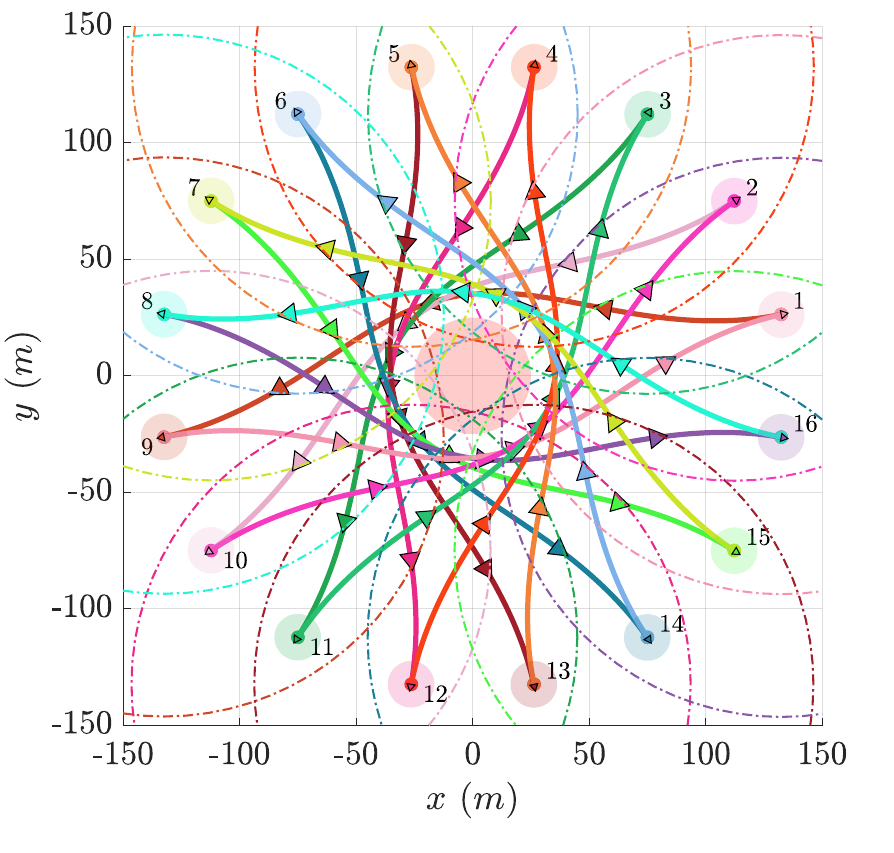}}
	\caption{ \label{fig:f3:plain}
	Distributed spatial-temporal joint optimization results for Scenario 3.}
\end{figure*}

The optimization results of the UAV swarm in Scenario 3 are shown in Fig.~\ref{fig:f3:plain}, which contains the optimized trajectories, control histories, time changes, and snapshots of the flights of the swarm UAVs. The control history and flight time optimization of the swarm UAVs have the same result due to the axisymmetric nature of each UAV. As shown in Fig.~\ref{fig:f3:plain}, {\colb the swarm UAVs can reach their respective target locations safely while maintaining communication and not conflicting with their neighbors as well as obstacles. Specifically, the closest distance between the UAVs are $13.8m$, and the communication maintenance distance $120m$ is not violated in this scenario.} The flight time of the swarm UAVs is optimized from the initial $9.2s$ to $9.36s$, showing that the algorithm can meet the distributed time optimization requirements of the scenario. {\col Notice that all UAVs in this scenario are placed evenly with distributed in a circular formation to cross an obstacle at the center of the formation while flying to another point symmetric to the center of the circle, the relative geometry between each UAV and its destination is the same across all UAVs. This, thus, generates the same command and mission time history, as shown in Figs. \ref{fig:f3:2} and \ref{fig:f3:3}.}

\textit{Scenario 4}: We consider 20 UAVs sequentially and uniformly distributed, flying through several random obstacles to the other side of the scene. During the flight to the target point, each swarm UAV needs to communicate with the UAVs with a neighborhood size of $|\mathcal{N}_i| = 5$ and maintain a safe distance. The length of each UAV's flight path may be different due to the complex environment, and hence each UAV optimizes its own flight time individually without requiring to satisfy the constraint \eqref{timeeq2}. {\colm Table~\ref{tab:condit4} summarizes the specific state of each UAV, the positions and radius of the obstacle, the inter-agent distance, and the initialized values and bounds of the flight time.}
\begin{table}[htb]
	\centering
	\caption{Initial conditions of Scenario 4.}
	\label{tab:condit4}
	\begin{tabular}{cc}
		\hline \hline
		Parameters & Value \\ \hline
		Initial state of UAV $i$ $\left[x_{i_0},y_{i_0},\theta_{i_0}\right]^{\rm{T}}$ & $\left[(20+10\times (-1)^i)m, (100+(i-1)\times 30)m, 0^{\circ}\right]^{\rm{T}}$\\
		Target state of UAV $i$ $\left[x_{i_f},y_{i_f},\theta_{i_f}\right]^{\rm{T}}$ & $\left[(290+10\times (-1)^i)m, (100+(i-1)\times 30)m, 0^{\circ}\right]^{\rm{T}}$\\
		{\col Initial guess of flight time of each UAV} $t_{i,N}$ & $9.2s$\\
		Center and Radius of circular obstacle $\boldsymbol{p}_{o_1},r_{o_1}$ & $\left[200.0m,200.0m\right]^{\rm{T}},15.0m$\\
		Center and Radius of circular obstacle $\boldsymbol{p}_{o_1},r_{o_2}$ & $\left[150.0m,120.0m\right]^{\rm{T}},20.0m$\\
		Center and Radius of circular obstacle $\boldsymbol{p}_{o_1},r_{o_3}$ & $\left[180.0m,300.0m\right]^{\rm{T}},20.0m$\\
		Center and Radius of circular obstacle $\boldsymbol{p}_{o_1},r_{o_4}$ & $\left[150.0m,390.0m\right]^{\rm{T}},20.0m$\\
		Center and Radius of circular obstacle $\boldsymbol{p}_{o_1},r_{o_5}$ & $\left[210.0m,500.0m\right]^{\rm{T}},20.0m$\\
		Center and Radius of circular obstacle $\boldsymbol{p}_{o_1},r_{o_6}$ & $\left[150.0m,580.0m\right]^{\rm{T}},15.0m$\\
		Center and Radius of circular obstacle $\boldsymbol{p}_{o_1},r_{o_7}$ & $\left[180.0m,680.0m\right]^{\rm{T}},20.0m$\\
		{\col Communication maintenance distance with neighbors} $d_{\mathrm{con}}$ & $170m$\\
		Safe distance to obstacle $d_{o}$ and neighbors $d_{\mathrm{col}}$ & $10m, 10m$\\
		\hline \hline
	\end{tabular}
\end{table}

\begin{figure*}[!p]
	\centering
	\begin{minipage}{0.49\linewidth}
		\centering
		\subfigure[\label{fig:f4:1}The initial guess trajectory (transparent line) and optimized flight trajectory (solid line) of each UAV.] %
			{\includegraphics[width=1.0\linewidth]{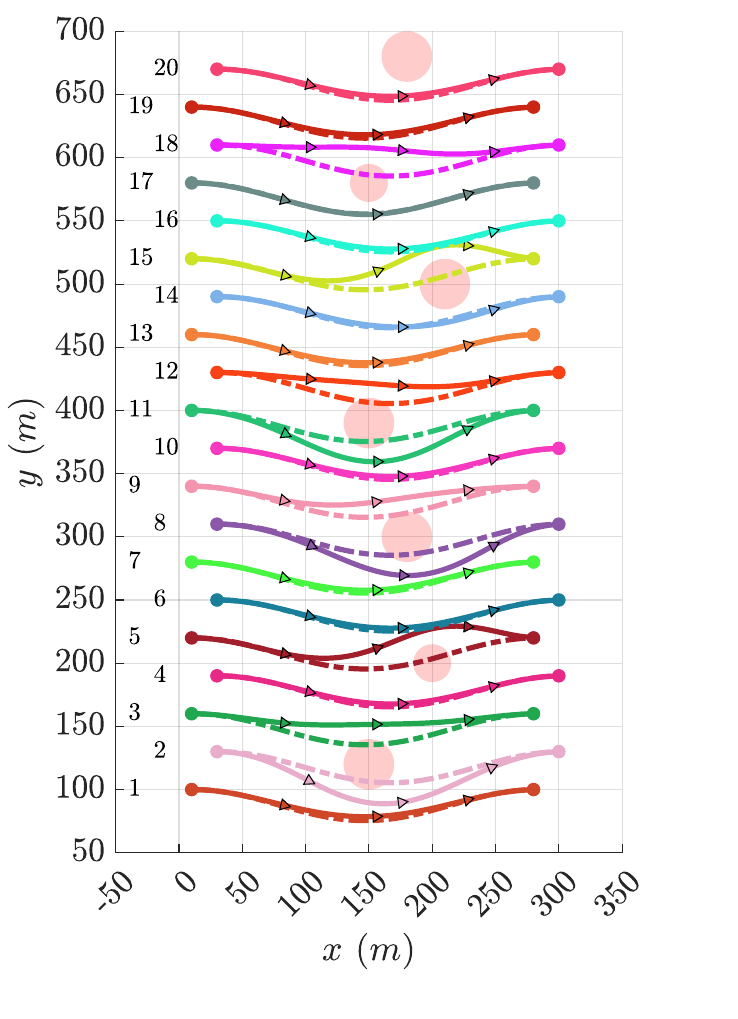}}
	\end{minipage}
	\begin{minipage}{0.49\linewidth}
		\centering
		\begin{minipage}{0.9\linewidth}
			\centering
			\subfigure[\label{fig:f4:2}Control angular velocity variation curve of each UAV.] %
				{\includegraphics[width=0.95\linewidth]{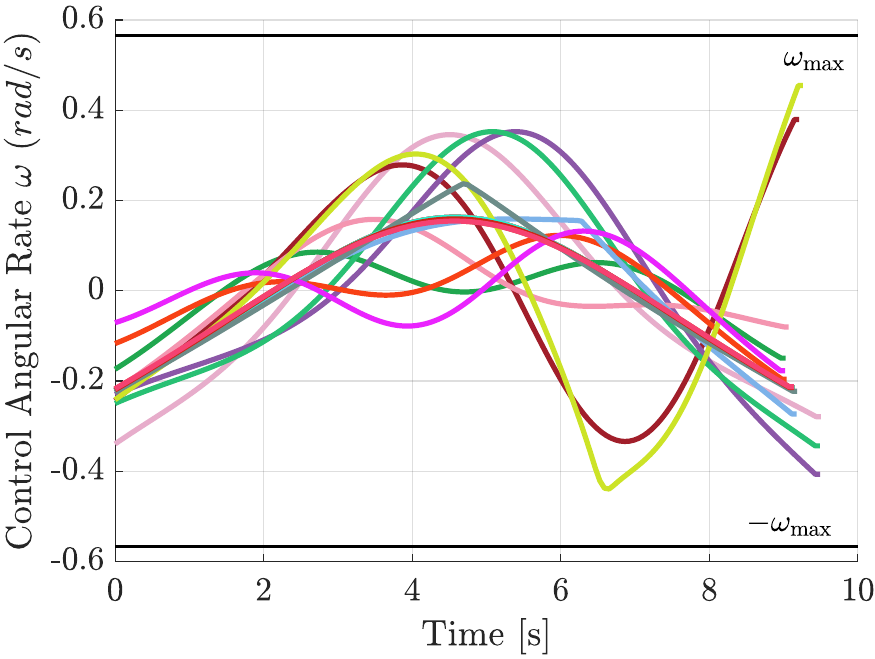}}
		\end{minipage}
		% 空格是必须的, the blank line is needed for the formatting
		\begin{minipage}{0.9\linewidth}
			\centering
			\subfigure[\label{fig:f4:3}The variation curve of the flight time for each UAV with the number of iterations.] %
				{\includegraphics[width=0.95\linewidth]{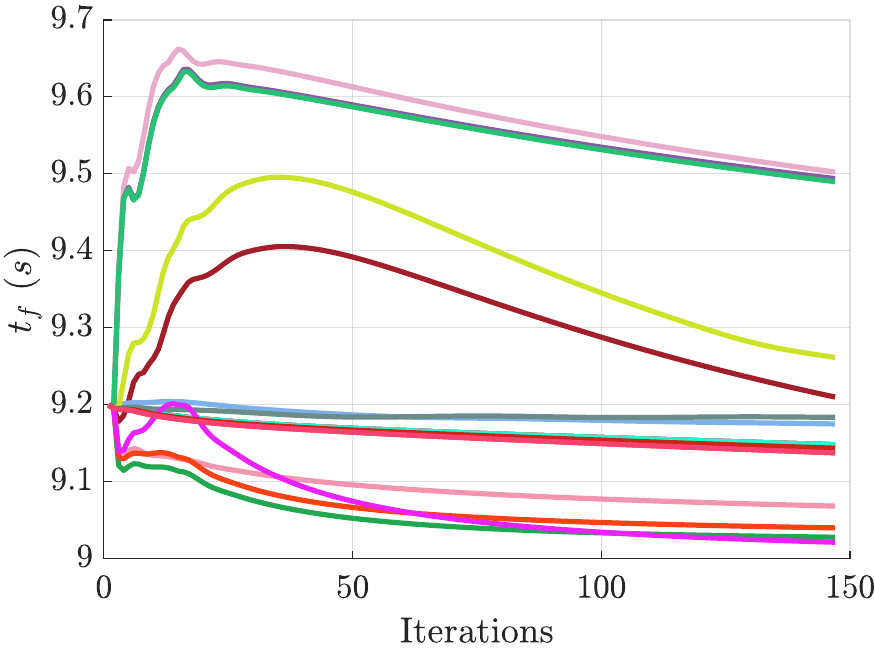}} 
		\end{minipage}
	\end{minipage} \\
	\subfigure[\label{fig:f4:4}The snapshot of the scene at the initial moment of flight $t_0$.] %
		{\includegraphics[width=0.30\textwidth]{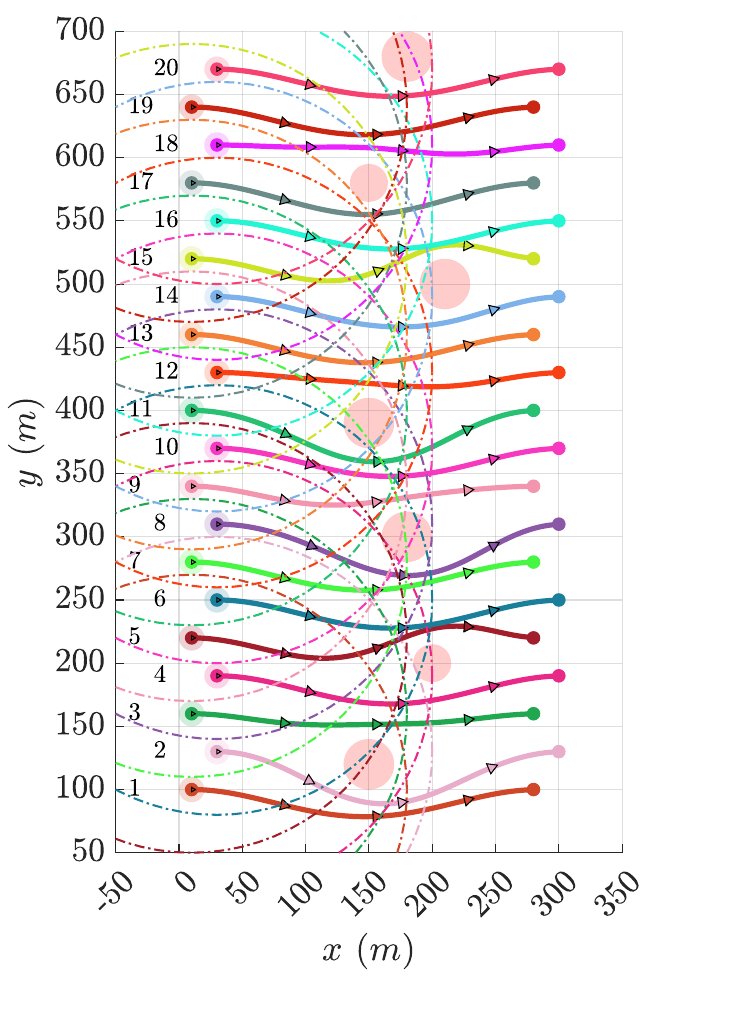}}
		\quad
	\subfigure[\label{fig:f4:5}The snapshot of the scene at the moment of minimum distance between the vehicles.] %
		{\includegraphics[width=0.30\textwidth]{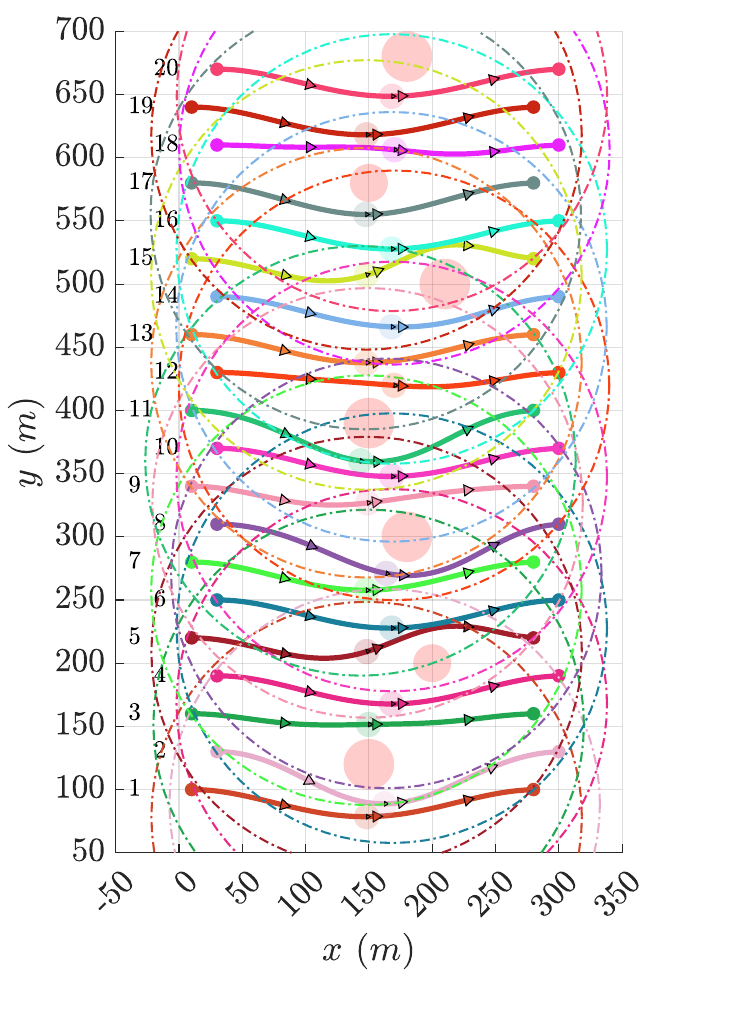}}
		\quad
	\subfigure[\label{fig:f4:6}The snapshot of the scene at the terminal moment of flight $t_f$.] %
		{\includegraphics[width=0.30\textwidth]{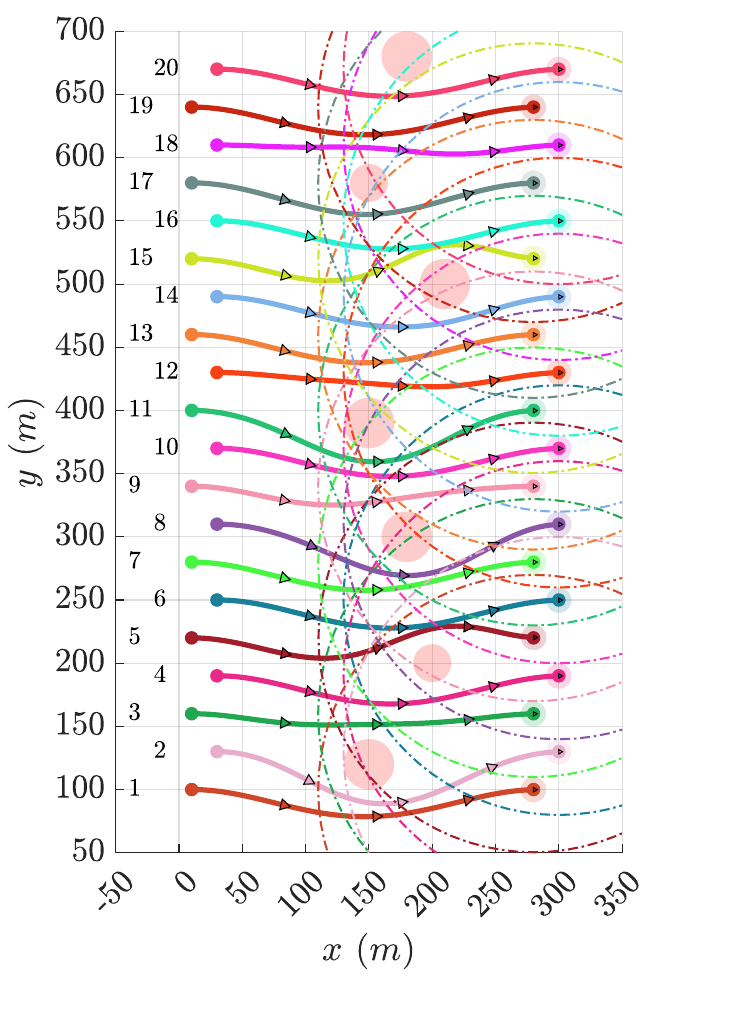}}
	\caption{ \label{fig:f4:plain}
	Distributed spatial-temporal joint optimization results for Scenario 4.}
\end{figure*}

{\col The optimization results of the UAV swarm in Scenario 4 are shown in Fig.~\ref{fig:f4:plain}, which contains the optimized trajectories, control histories, time changes, and snapshots of the flights of the UAV swarm. As shown in Fig.~\ref{fig:f4:4} to Fig.~\ref{fig:f4:6}, we can see that all $20$ UAVs can reach a consensus with their neighbors and complete the trajectory optimization of their own trajectories against obstacles and other UAVs' avoidance, as well as keeps each other within the communication distance.} From Fig.~\ref{fig:f2:3} we can also see that each UAV is able to adjust its flight time according to its flight path length.

\subsection{Comparison with Differential Acceleration Schemes} \label{ssec:comparations}
In this section, the fixed penalty parameters in the previous section will be regarded as tuning parameters and different acceleration schemes are compared in order to verify the superior acceleration performance of the proposed spectral gradient adaptive parameter (AP) scheme for the D-PDDP algorithm. The proposed acceleration scheme is compared with the original fixed penalty parameter scheme, the residual balancing (RB) scheme \cite{10288223} and the Nesterov acceleration (NA) scheme \cite{doi:10.1137/120896219} in the four aforementioned scenarios to demonstrate superior performance in terms of acceleration. {\colb To ensure that the comparisons are fair, all
%To ensure the fairness of the comparison, all the 
scenarios} and simulation-related settings are the same as in Sec.~\ref{ssec:simulations}, except that the penalty parameter is shifted from fixed to varying. {\col In addition, the implementation of the RB scheme is summarized in Eq.~\eqref{eq:RB}, same as in \cite{10288223}. The penalty parameter is dynamically adjusted based on the residuals ratio to enhance convergence efficiency.}
\begin{equation}
\label{eq:RB}
	\rho_i^{n+1} = 
	\begin{cases}
		\rho_i^{n} / 2 & \text { if } \left\|\boldsymbol{r}_{(\cdot)}^{\mathrm{p},n}\right\|_2 \geq 10\left\|\boldsymbol{r}_{(\cdot)}^{\mathrm{d},n}\right\|_2 \\ 
		2\rho_i^{n} & \text { if } \left\|\boldsymbol{r}_{(\cdot)}^{\mathrm{d},n}\right\|_2 \geq 10\left\|\boldsymbol{r}_{(\cdot)}^{\mathrm{p},n}\right\|_2
	\end{cases}
%	\label{alphacal}
\end{equation}

The NA parameter $\alpha_{\it{Nes}_1}$ is set to $1$ and the tuning parameter $\eta_{\it{Nes}_1}$ is set to $0.9$. The relevant parameters for the adaptive parameter penalty are chosen as the safeguarding threshold $\epsilon_{\mathrm{cor}} = 0.5$ and the convergence constant $C_{\mathrm{cg}} = 5\times 10^2$. The penalty parameter is updated every 10 iteration steps in all acceleration schemes, i.e., $C_{\it{Freq}}$ in Algorithm~\ref{algo: adapt} is set to $10$. The comparison data of the three different schemes in the four scenarios are summarized in Table~\ref{tab:compare}.

{\colm The different schemes are all able to generate spatial-temporal optimized trajectories that satisfy the individual constraints as well as the inter-UAV constraints in all four scenarios, for which the specific results are not repeated in this section. As can be seen from Table~\ref{tab:compare}, the proposed AP and the RB schemes reduce the number of iterations and the computational time of the original D-PDDP algorithm in all four scenarios, while the NA scheme increases the iterations in Scenario 2 instead. In all simulations, the proposed approach reduces the number of iterations more substantially than the other two schemes. And in Scenario 1, 2 and 4, the proposed method significantly reduces the computational time over the other two schemes. In summary, the proposed acceleration scheme proves to be an effective acceleration scheme.}
\begin{table}[htb]
	\centering
	\caption{\colm Iterations of different acceleration schemes.}
	\label{tab:compare}
	\begin{tabular}{c|cc|ccc|ccc|ccc}
		\hline \hline
		Schemes & Vanilla & Time & RB & Iter & Time & NA & Iter & Time & AP & Iter & Time \\ \hline
		1 & $102$ & $90.45s$	& $64$ & $37.25\%$ 	 & $54.56s$  & $70$ & $31.37\%$	& $66.74s$		& $48$  & $\textbf{52.94\%}$ & $\textbf{40.39s}$\\
		2 & $128$ & $132.51s$	& $100$ & $21.875\%$  & $103.59s$ & $129$ & $-0.78\%$ & $137.89s$		& $95$ & $\textbf{25.78\%}$ & $\textbf{100.22s}$\\
		3 & $89$ & $219.60s$	& $73$ & $17.98\%$ 	 & $179.69s$  & $69$ & $22.47\%$	& $\textbf{173.93s}$		& $68$ & $\textbf{23.60\%}$ & $175.50s$\\
		4 & $147$ & $406.20s$	& $71$ & $51.70\%$ 	 & $196.77s$  & $130$ & $11.56\%$ & $369.47s$		& $68$& $\textbf{53.74\%}$ & $\textbf{189.27s}$\\
		\hline \hline
	\end{tabular}
\end{table}

\section{Conclusions} \label{sec: conclusion}
In this paper, a fully distributed spatial-temporal trajectory optimization method is proposed for large-scale UAV swarm missions. Besides, an adaptive penalty parameter based on the spectral gradient method is proposed to efficiently reduce the number of iterations of the proposed algorithm. Our algorithm is capable of generating safe (both inter and outer) trajectories that satisfy different terminal time constraints (including time sequential constraints, simultaneous arrival and optimal free terminal time). The results with different scenarios demonstrate the flexibility, convergence and optimization capability of the proposed approach.

{\colm Due to the distributed computing nature of the D-PDDP algorithm, when migrating the algorithm to a real UAV swarm, it is only necessary to deploy the algorithm executed by each UAV in this paper to run independently on each real UAV, while guaranteeing the optimality of the global task. However, since this paper does not consider the communication delays of UAV swarm in real world, the algorithms need to be further investigated before deploying them to real swarm.}

%The proposed D-PDDP algorithm is validated in four scenarios for its distributed optimization capability for spatial-temporal coupled constrained trajectories of up to dozens of UAVs. The proposed acceleration scheme is compared with the Residual Balancing scheme and the Nesterov Acceleration Scheme, and the comparison results illustrate the proposed adaptive penalty parameter scheme for the D-PDDP is a promising acceleration scheme.

%Of course, the proposed algorithm still has shortcomings. Due to the large number of iterative steps in D-PDDP and the need to reach consensus through communication within the UAV swarm, the computational efficiency of the current algorithm is still insufficient to support online applications, and future research will focus on accelerating the algorithm more sufficiently to support online applications. Moreover, the D-PDDP algorithm based on distributed architecture operates under the assumption of the existence of an ideal communication situation, but communication delays and data loss are bound to exist in the actual UAV swarm environment. The presence of these factors will place new demands on the algorithm, so future research will consider communication-related failure factors.
\bibliography{sample}

\end{document}